\documentclass[manuscript,screen]{acmart}

\usepackage{sidecap}
\sidecaptionvpos{table}{c}
\sidecaptionvpos{figure}{c}

\usepackage{amsmath}
\usepackage{booktabs}
\usepackage{mathtools}
\usepackage{xspace}
\usepackage{hyperref}
\usepackage{graphicx}
\usepackage{caption}
\usepackage{subcaption}
\usepackage{multirow}
\usepackage{color, colortbl}
\usepackage{fancyvrb}
\usepackage{blindtext}
\usepackage{appendix}
\usepackage{natbib}
\usepackage{bm}
\usepackage{algorithm}
\usepackage{etoolbox}
\usepackage{lipsum}
\usepackage{array}
\newcolumntype{M}{>{$}c<{$}}

\usepackage[noend]{algpseudocode}

\definecolor{Gray}{gray}{0.9}
\definecolor{pink}{rgb}{1.0, 0.13, 0.32}

\newcommand{\ie}{\emph{i.e.,}\xspace}

\AtBeginDocument{%
  \providecommand\BibTeX{{%
    \normalfont B\kern-0.5em{\scshape i\kern-0.25em b}\kern-0.8em\TeX}}}

\begin{document}

\title{Improving Visual Question Answering Models through Robustness Analysis and In-Context Learning with a Chain of Basic Questions}


\author{Jia-Hong Huang}
\affiliation{%
  \institution{University of Amsterdam}
  \city{Amsterdam}
  \country{The Netherlands}}
\email{j.huang@uva.nl}

\author{Modar Alfadly}
\affiliation{%
  \institution{King Abdullah University of Science and Technology}
  \city{Makkah}
  \country{KSA}}
\email{modar.alfadly@kaust.edu.sa}

\author{Bernard Ghanem}
\affiliation{%
  \institution{King Abdullah University of Science and Technology}
  \city{Makkah}
  \country{KSA}}
\email{bernard.ghanem@kaust.edu.sa}

\author{Marcel Worring}
\affiliation{%
  \institution{University of Amsterdam}
  \city{Amsterdam}
  \country{The Netherlands}}
\email{m.worring@uva.nl}



\begin{abstract}
\section*{Abstract}
Deep neural networks have been critical in the task of Visual Question Answering (VQA), with research traditionally focused on improving model accuracy. Recently, however, there has been a trend towards evaluating the robustness of these models against adversarial attacks. This involves assessing the accuracy of VQA models under increasing levels of noise in the input, which can target either the image or the proposed query question, dubbed the main question. However, there is currently a lack of proper analysis of this aspect of VQA. This work proposes a new method that utilizes semantically related questions, referred to as basic questions, acting as noise to evaluate the robustness of VQA models. It is hypothesized that as the similarity of a basic question to the main question decreases, the level of noise increases. To generate a reasonable noise level for a given main question, a pool of basic questions is ranked based on their similarity to the main question, and this ranking problem is cast as a $LASSO$ optimization problem. Additionally, this work proposes a novel robustness measure, $R_{score}$, and two basic question datasets to standardize the analysis of VQA model robustness. The experimental results demonstrate that the proposed evaluation method effectively analyzes the robustness of VQA models.  Moreover, the experiments show that in-context learning with a chain of basic questions can enhance model accuracy.

\end{abstract}




\maketitle

\section{Introduction}
Visual Question Answering (VQA) is a complex computer vision task that involves providing an algorithm with a natural language question relating to an image and requiring it to produce a natural language answer for that particular question-image pair. In recent times, numerous VQA models \cite{4,5,9,31,37,41,57,58,59,77,82,agrawal2018don,vedantam2019probabilistic,chen2020counterfactual,sheng2021human,kolling2022efficient} have been proposed to address this challenge. The primary performance metric used to evaluate these models is accuracy.


The research community has begun to acknowledge that accuracy alone is not a sufficient metric to assess model performance \cite{kafle2017visual,kafle2017analysis}. In addition to accuracy, models should also be robust, meaning their output should not be significantly affected by minor \emph{perturbations} or \emph{noise} added to the input. This includes replacing words with similar words, phrases, or sentences in input questions, or slightly altering pixel values in the image. The analysis of model robustness and training of robust models is a rapidly growing research topic for deep learning models applied to images \cite{61,62,63}. However, to the best of our knowledge, an acceptable and standardized method for measuring robustness in VQA models does not currently exist.

\begin{SCfigure}
  \includegraphics[width=0.42 \linewidth]{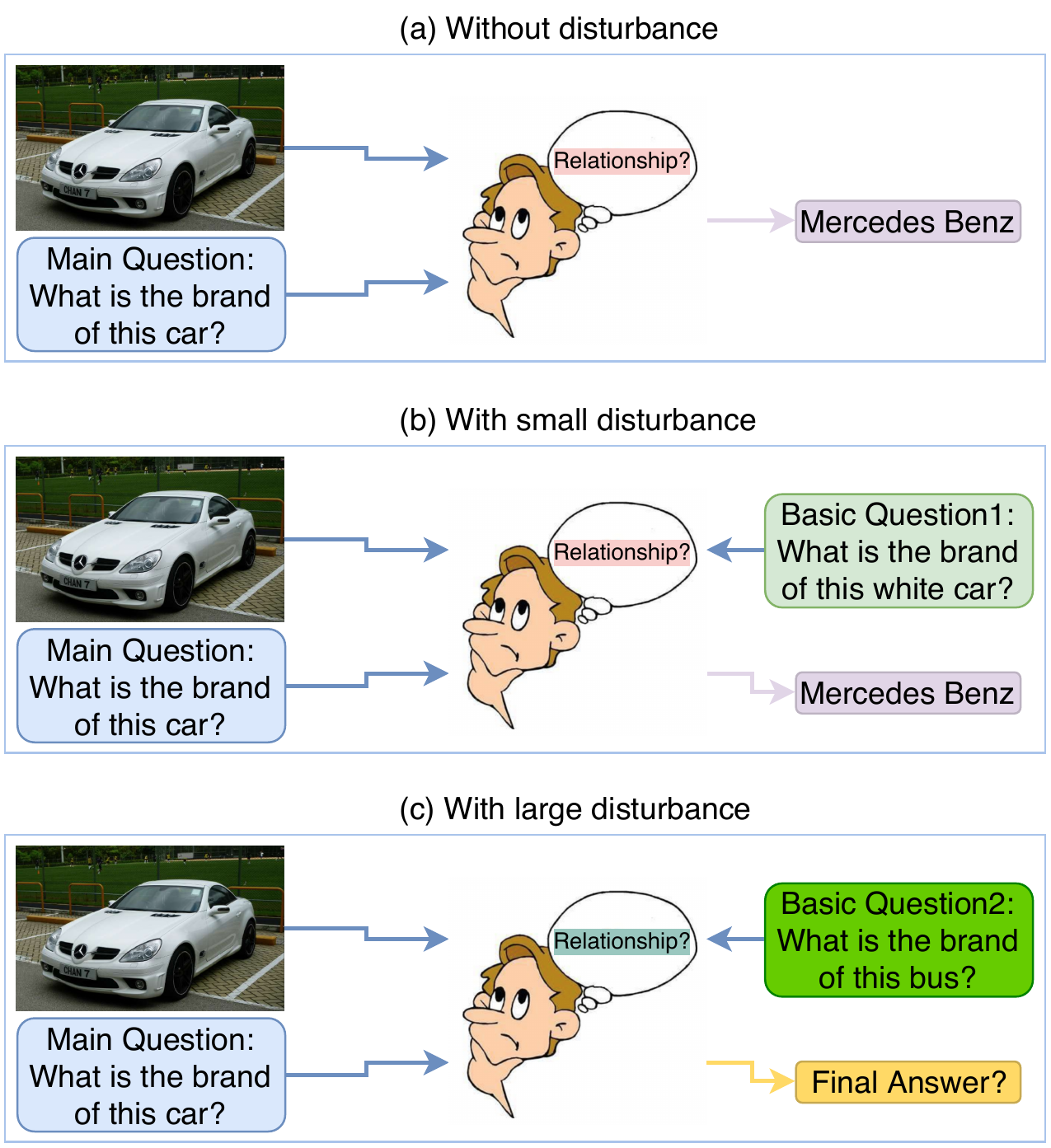}
  \caption{This figure (Figure \ref{fig:figure1}) is inspired by Deductive Reasoning in Human Thinking \cite{74} and illustrates how humans behave when subjected to multiple questions about a specific topic. In cases (a) and (b), the person might have the same answer ``Mercedes Benz'' in mind for both cases. However, in case (c), the person would start to consider the relationships among the provided questions and candidate answers to form the final answer, which may differ from the final answer in cases (a) and (b). When given more basic questions, the person would need to consider all the possible relationships among the questions and answer candidates. These relationships can be very complex, especially when the additional basic questions have low similarity scores to the main question, which can mislead the person. In such cases, the extra basic questions act as large disturbances. It is important to note that the relationships and final answer in cases (a) and (b) could be the same but different from the case (c). To help clarify these relationships, we have used different colors in the figure.}
\label{fig:figure1}
\end{SCfigure}
\begin{SCfigure}
\scalebox{0.9}{
  \includegraphics[width=0.42 \linewidth]{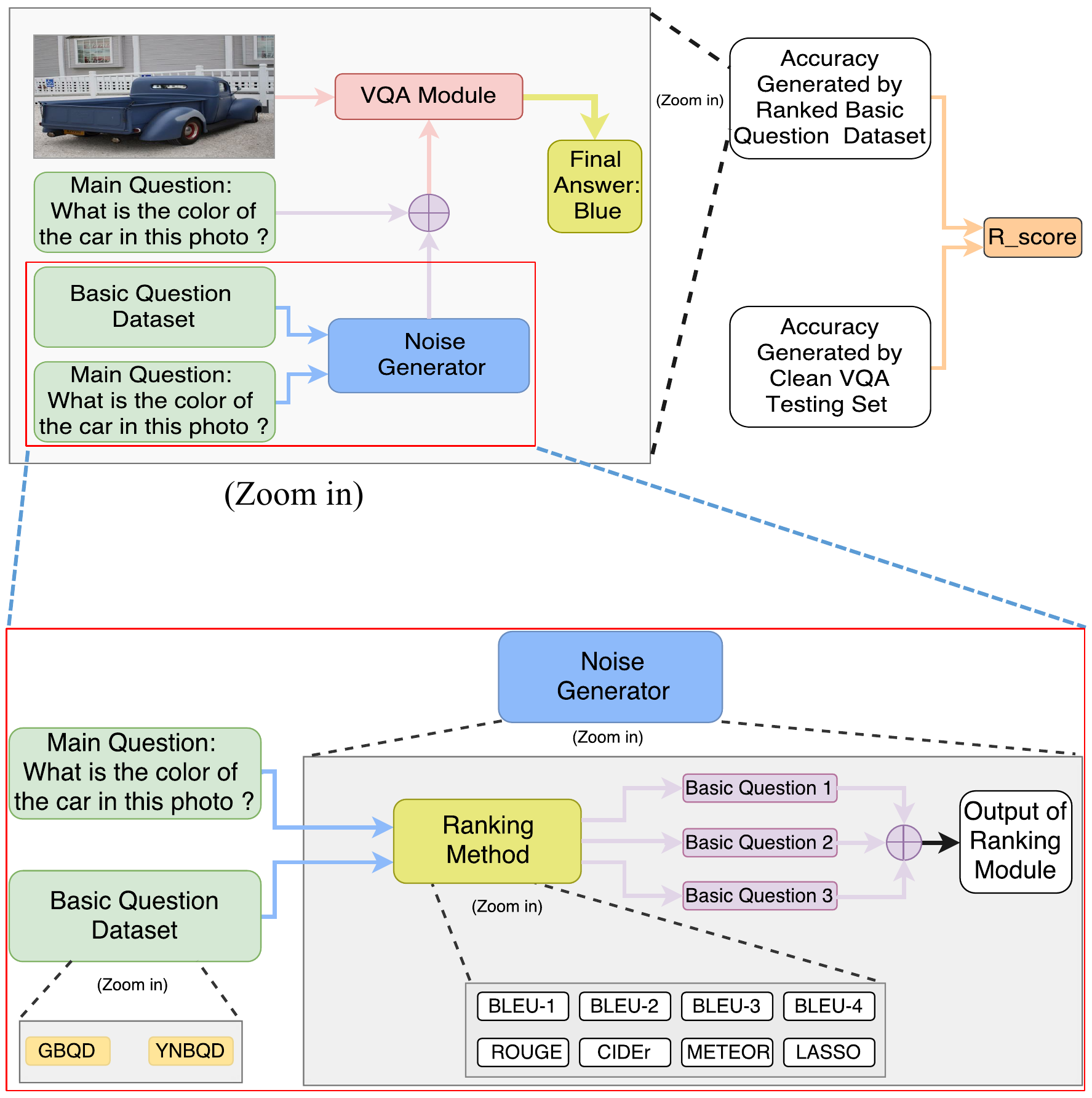}}
  \caption{The proposed method for measuring the robustness of VQA models. Our proposed method is based on the \textit{``Accuracy Generated by Ranked Basic Question Dataset''} and \textit{``Accuracy Generated by Clean VQA Testing Set''}, which are combined to generate the $R_{score}$, our proposed measure of robustness. The upper part of the figure depicts the VQA Module and Noise Generator, which are the two main components of our method. The lower part of the figure shows a detailed view of the Noise Generator. Our method allows for the use of two different Basic Question Datasets, GBQD and YNBQD, and eight different question ranking methods. If new datasets or ranking methods become available in the future, they can be incorporated into our method. The output of the Noise Generator is the concatenation of three ranked basic questions. ``$\oplus$'' denotes the direct concatenation of basic questions.}
\label{fig:figure100}
\end{SCfigure}

To establish a measure of robustness, we note that the ultimate goal for VQA models is to perform comparably to humans. When presented with a question or a highly similar question, humans typically provide the same or a very similar answer. This phenomenon has been reported in psychology research \cite{74}. In this work, we designate the input question as the main question and define a basic question as a question that is semantically similar to the main question. If we add or replace some words or phrases in the main question with semantically similar entities, the VQA model should output the same or a very similar answer. This is illustrated in Figure \ref{fig:figure1}, and we consider these added entities as small perturbations or noise to the input. The model is considered robust if it produces the same answer. As studying robustness necessitates the analysis of VQA model accuracy under varying noise levels, we require a method for quantifying the level of noise for a given question. We posit that a basic question with a higher similarity score to the main question introduces less noise when added to the main question, and vice versa. Inspired by this idea, we present a novel method for measuring the robustness of VQA models, as illustrated in Figure \ref{fig:figure100}. The method comprises two modules: a VQA model and a Noise Generator. The Noise Generator accepts a plain text main question (MQ) and a plain text basic question dataset (BQD) as input. It begins by ranking the basic questions in BQD based on their similarity to MQ using a text similarity ranking method. We measure the robustness of the VQA model by comparing its accuracy with and without generated noise at different noise levels. We propose a robustness measure $R_{score}$ to evaluate performance.

When considering the similarity between a main question and a basic question, there are various measures that can be used to produce a score. These scores then determine the ranking of the basic questions. Text similarity metrics like BLEU (BiLingual Evaluation Understudy) \cite{49} are commonly used to compute the overlap between two texts, but they cannot effectively capture the semantic meaning of the text. As a result, rankings based on these metrics may not accurately reflect the similarity between questions.
To enhance the quality of question ranking, we introduce a new method formulated using $LASSO$ optimization and compare it with commonly used textual similarity measures. We evaluate the effectiveness of our method by ranking our proposed Basic Question Datasets (BQDs), including the General Basic Question Dataset (GBQD) and Yes/No Basic Question Dataset (YNBQD). We also examine the robustness of six pre-trained state-of-the-art VQA models \cite{4,41,57,59} and compare the results obtained from our proposed $LASSO$ ranking method with other metrics in BQD ranking through extensive experiments. The experimental results indicate the effectiveness of our proposed method and demonstrate that in-context learning with a chain of basic questions improves the model's accuracy.


It is crucial to emphasize that commonly used textual similarity measures are not effective in controlling the noise level in basic question (BQ) rankings. Consequently, conducting a robustness analysis becomes extremely challenging. However, our proposed $LASSO$ basic question ranking method is highly efficient in quantifying and controlling the intensity of the injected noise level. This approach empowers us to evaluate the robustness of VQA models under various noise levels and explore their performance accurately. This paper presents several contributions to the field of Visual Question Answering (VQA):


\begin{itemize}
    \item[$\bullet$]  We introduce two datasets of basic questions, which can be used to evaluate the robustness of VQA models. These datasets are made publicly available.
    \item[$\bullet$]  We propose a novel method for measuring the robustness of VQA models and apply it to six state-of-the-art models. Our method can generate noise levels of varying strength and quantifies the impact of this noise on model performance.
    \item[$\bullet$]  We introduce a new text similarity ranking method based on $LASSO$ optimization and demonstrate its superiority over seven popular similarity metrics. This method can effectively rank basic questions according to their similarity to a main question.
   \item[$\bullet$]  We adopt an in-context learning perspective to explore how basic questions, \ie chain-of-question, can enhance the performance of VQA models.
\end{itemize}

The rest of this paper is structured as follows. In Section 2, we provide an overview of the related works. In Section 3, we describe the details of our proposed method and demonstrate how to use it to measure the robustness of VQA models. Furthermore, in Sections 4 and 5, we present various analyses on our proposed General Basic Question Dataset (GBQD) and Yes/No Basic Question Dataset (YNBQD) \cite{huang2019novel}. Finally, in Section 6, we compare the robustness and accuracy performance of state-of-the-art VQA models.


\noindent
\vspace{+0.1cm}\textbf{Relations to our previous work}

This paper builds upon our previous work, which was presented as an oral paper at the Thirty-Third AAAI Conference on Artificial Intelligence (AAAI-2019) \cite{huang2019novel}, and presents several improvements. Firstly, we propose a framework, depicted in Figure \ref{fig:figure8}, and a threshold-based criterion, outlined in Algorithm \ref{algorithm1}, to leverage basic questions (BQs) for analyzing the robustness of the HieCoAtt VQA model \cite{41}. Secondly, we adopt an in-context learning perspective to demonstrate how BQs can enhance the performance of the HieCoAtt VQA model. Thirdly, we  emphasize the necessity of preprocessing question sentences for our proposed LASSO ranking method, which ensures its correct functioning. Finally, we present an extended experiment on YNBQD. The current paper is a fully restructured and rewritten version of our previous work and incorporates these new contributions.


\section{Related Work}
In recent years, VQA has emerged as a captivating and challenging task, attracting significant attention from researchers in various fields, such as natural language processing (NLP), computer vision, and machine learning. A wide range of approaches has been proposed to tackle this task, as evidenced by a growing body of literature \cite{2,5,6,9,16,26,34,39,41,42,54,64,agrawal2018don,vedantam2019probabilistic,chen2020counterfactual,sheng2021human,kolling2022efficient,ravi2023vlc}. In this paper, we review related works from different perspectives, such as sentence evaluation metrics, models' accuracy and robustness, and datasets.

\noindent
\vspace{+0.1cm}\textbf{Sentence Evaluation Metrics}

Various sentence evaluation metrics have been widely adopted in different tasks, such as video/image captioning \cite{yu2016video,huang2021deepopht,huang2022non,huang2021contextualized,huang2021deep,huang2021longer} and text summarization \cite{barzilay1999using}. In this paper, we leverage commonly used metrics to measure the similarity between BQ and MQ. BLEU (BiLingual Evaluation Understudy) \cite{49} is a widely used metric for machine translation, based on precision. However, its effectiveness has been questioned by some studies \cite{70,71}. METEOR \cite{69}, on the other hand, is based on the harmonic mean of unigram precision and recall, and it can handle stemming and synonym matching. It has been proposed as a solution to some of the problems found with BLEU and produces a better correlation with translations by human experts. While METEOR evaluates the correlation at the sentence and segment level, BLEU looks for correlations at the corpus level. ROUGE (Recall Oriented Understudy of Gisting Evaluation) \cite{68} is another recall-based metric that is popular in the text summarization community. It tends to reward longer sentences with higher recall. CIDEr \cite{67}, a consensus-based metric, rewards a sentence for being similar to the majority of descriptions written by human experts and is often used in the image captioning community. It extends existing metrics with \emph{tf-idf} weights of \emph{n}-grams between a candidate sentence and a reference sentence. However, CIDEr can be inefficient for natural language sentence evaluation, as it may weigh unnecessary parts of the sentence and lead to ineffective scores. In our experiments, we use all of the above metrics along with our proposed $LASSO$ ranking approach to rank BQs and compare their performance.

\noindent
\vspace{+0.1cm}\textbf{Evaluating Image Captioning}

Several techniques commonly used in image captioning tasks have also been applied to the VQA task \cite{1,27,28,48}. For instance, in \cite{48}, the authors utilize a language model to combine a set of possible words detected in multiple regions of the input image and generate a corresponding description. In \cite{28}, a convolutional neural network model is used to extract high-level image features, which are then given to an LSTM unit as the first input. In \cite{1}, an algorithm is proposed to generate a word at each time step by focusing on local image regions related to the predicted word at the current time step. The authors of \cite{27} suggest a deep neural network model to learn how to embed language and visual information into a common multimodal space. Furthermore, while BLEU is a commonly used metric to evaluate image captioning results, it may not be the most appropriate metric to assess the quality of the captions due to its inherent limitations.

\noindent
\vspace{+0.1cm}\textbf{Evaluating Visual Question Answering}

VQA is a multimodal task, involving two types of inputs with different modalities: the question sentence and the image. Researchers have focused on modeling the interactions between the two different embedding spaces in several ways. For instance, bilinear interaction between two embedding spaces has been shown to be successful in deep learning for fine-grained classification and multimodal language modeling in previous works such as \cite{26,60}. Other methods proposed to compute the outer product between visual and textual features, such as Multimodal Compact Bilinear (MCB) pooling \cite{58}, or parameterize the full bilinear interactions between image and question sentence embedding spaces, as in Multimodal Low-rank Bilinear (MLB) pooling \cite{59}. An alternative method proposed in \cite{57} efficiently parameterizes the bilinear interactions between textual and visual representations, and shows that MCB and MLB are special cases of their proposed method. Some researchers exploit Recurrent Neural Networks (RNN) and Convolutional Neural Networks (CNN) to build a question generation algorithm in \cite{32}, and RNN to combine the word and image features for the VQA task in \cite{9,33,64}. The authors of \cite{30} have tried to exploit convolutions to group the neighboring features of word and image, while the authors of \cite{6} use Gated Recurrent Unit (GRU) \cite{10} to encode an input question and introduce a dynamic parameter layer in their CNN model, where the weights of the model are adaptively predicted by the embedded question features. However, to the best of our knowledge, no existing VQA method has been evaluated by a robustness-based dataset, since such a dataset does not exist.

\noindent
\vspace{+0.1cm}\textbf{Robustness of Neural Network Models}

Several recent works (e.g., \cite{61,62,63,kafle2017visual,kafle2017analysis,huang2019novel,huang2017vqabq,huang2017robustness,huang2017robustnessMS,hu2019silco,huck2018auto,liu2018synthesizing,yang2018novel,di2021dawn,huang2022causal,wu2023expert}) have explored the issue of deep learning model robustness from an image or text perspective. In \cite{61,62}, the authors analyze model robustness by adding noise or perturbations to images and observing their impact on predicted results. The authors of \cite{moosavi2018robustness} provide theoretical evidence for a strong relationship between small curvature and large robustness, proposing an efficient regularizer that encourages small curvatures and leads to significant boosts in neural network robustness. While most existing works focus on adding noise to the image input, our work instead focuses on adding noise to the text input \cite{huang2019novel}. Specifically, we consider the semantically related BQs of a given MQ as a type of noise for the MQ, using these BQs to evaluate the robustness of VQA models.



\noindent
\vspace{+0.1cm}\textbf{Datasets for Visual Question Answering}



Recently, several VQA datasets focused on accuracy have been proposed. The first dataset is DAQUAR (DAtaset for QUestion Answering on Real-world images) \cite{2}, containing around $12.5k$ manually annotated question-answer pairs for approximately $1449$ indoor scenes \cite{84}. The original DAQUAR dataset provides only one ground truth answer per question, but additional answers are collected by the authors of \cite{64}. Three other VQA datasets based on MS-COCO \cite{51} are subsequently proposed: \cite{85,4,33}. In \cite{85}, existing image caption generation annotations are transformed into question-answer pairs using a syntactic parser \cite{8} and hand-designed rules. VQA \cite{4}, another popular dataset, includes approximately $614k$ questions about the visual content of $205k$ real-world images, along with $150k$ questions based on $50k$ abstract scenes. The VQA dataset provides $10$ answers for each image, and the test set answers have not been released due to the VQA challenge workshop. In \cite{33}, approximately $158k$ images are annotated with $316k$ Chinese question-answer pairs and their English translations. Visual Madlibs \cite{90} is introduced to simplify VQA model performance evaluation by introducing a multiple-choice question-answering task. In this task, the VQA model chooses one of four provided answers based on a given image and prompt, eliminating ambiguity in answer candidates. The performance of different VQA models is measured using a simple accuracy metric. However, the holistic reasoning required by VQA models based on the given images in this task remains challenging for machines, despite the simple evaluation. Automatic and simple performance evaluation metrics have been incorporated into building the VQA dataset \cite{2,77,87}. The Visual7W dataset, developed by the authors \cite{34}, contains over $330k$ natural language question-answer pairs based on the Visual Genome dataset \cite{91}. Unlike other datasets such as VQA and DAQUAR, the Visual Genome dataset focuses on answering the six Ws (\textit{what, where, when, who, why,} and \textit{how}) with a text-based sentence. Visual7W builds upon this foundation by including extra correspondences between questions and answers, as well as requiring answers that locate objects. Multiple-choice answers, similar to those in Visual Madlibs \cite{90}, are also included. Additionally, the authors of \cite{92} have proposed Xplore-M-Ego, a dataset of images with natural language queries, a media retrieval system, and collective memories. Xplore-M-Ego focuses on a dynamic, user-centric scenario where answers are conditioned not only on the question, but also on the geographical position of the questioner. Another related task is video question answering, which requires understanding long-term relations in videos. The authors of \cite{89} have proposed a task that involves filling in blanks in captions associated with videos, requiring inference of the past, present, and future across a diverse range of video descriptions data from movies \cite{89,94,95}, cooking videos \cite{93}, and web videos \cite{ji2019query}. However, these datasets are accuracy-based and cannot evaluate the robustness of VQA models. In this study, we propose robustness-based datasets GBQD and YNBQD to address this issue.

\section{Methodology}



This section presents our proposed method, which aims to analyze the robustness of pre-trained VQA models using a set of BQs generated with different metrics. First, we discuss how we embed questions and use various ranking methods, including BLEU-1, BLEU-2, BLEU-3, BLEU-4, ROUGE, CIDEr, METEOR, and our proposed $LASSO$ ranking method, to create BQs. Next, we explain how we evaluate the robustness of six state-of-the-art VQA models using these BQs. The method consists of two main components: the VQA module, which includes the model under analysis, and the Noise Generator, which generates noise for a given main question using the ranking methods. Our hypothesis, as introduced in the previous section, is that an accurately ranked set of BQs should lead to decreasing accuracy of the VQA model. To facilitate the discussion, we introduce some basic notations for our method. The overall approach is illustrated in Figure \ref{fig:figure100}.


\noindent
\vspace{+0.1cm}\textbf{Question Encoding}

The first step in our method is the embedding of the question sentences. Let $w_{i}^{1},...,w_{i}^{N}$ be the words in question $q_{i}$, with $w_{i}^{t}$ denoting the $t$-th word for $q_{i}$ and $\mathbf{x}_{i}^t$ denoting the $t$-th word embedding for $q_{i}$. Various text encoders such as Word2Vec \cite{47}, GloVe \cite{11}, and Skip-thoughts \cite{43} are commonly used in natural language processing \cite{huang2020query,huang2021gpt2mvs}. Since we aim to generate BQs that are semantically similar to the given MQ, we need an encoder that can accurately capture the meaning of a sentence. Among these options, Skip-thoughts is particularly suited for this task because it focuses on capturing the semantic relationships between words within a sentence. Therefore, we use Skip-thoughts to embed the questions in this paper. The Skip-thoughts model utilizes an RNN encoder with GRU activations to map an English sentence, denoted by $q_{i}$, to a feature vector $\mathbf{v} \in \mathbf{R}^{4800}$. We encode all the training and validation questions from the VQA dataset \cite{4} into a matrix $\mathbf{A}$, where each column represents a Skip-thoughts embedded basic question candidate. In our approach, we use $\mathbf{b}$ to represent the Skip-thoughts encoded main question.

At each time step, the question encoder generates a hidden state $\mathbf{h}_{i}^{t}$. This state can be viewed as the representation of the sequence \{$w_{i}^{1},..., w_{i}^{t}$\}. As such, the final hidden state $\mathbf{h}_{i}^{N}$ represents the entire sequence $\{ w_{i}^{1},...,w_{i}^{t},...,w_{i}^{N}\}$, which corresponds to a question sentence in our case. To simplify the presentation, we omit the index $i$ and use the following sequential equations to encode a question:
\begin{equation}
    ~~~~~~~~~~~~~~~~~\mathbf{r}^{t}~=~\sigma (\mathbf{U}_{r}\mathbf{h}^{t-1}+\mathbf{W}_{r}\mathbf{x}^{t})
\end{equation}
\begin{equation}
    ~~~~~~~~~~~~~~~~~\mathbf{z}^{t}~=~\sigma(\mathbf{U}_{z}\mathbf{h}^{t-1}+\mathbf{W}_{z}\mathbf{x}^{t})
\end{equation} 
\begin{equation}
    ~~~~~~~~~~~~\bar{\mathbf{h}}^{t}~=~\mathrm{tanh}(\mathbf{U}(\mathbf{r}^{t}\odot \mathbf{h}^{t-1})+\mathbf{Wx}^{t})
\end{equation} 
\begin{equation}
    ~~~~~~~~~~~~~~\mathbf{h}^{t}~=~\mathbf{z}^{t}\odot  \bar{\mathbf{h}}^{t}+(1-\mathbf{z}^{t})\odot \mathbf{h}^{t-1}, 
\end{equation}

\noindent
where the matrices of weight parameters are denoted by $\mathbf{U}_{r}$, $\mathbf{U}_{z}$, $\mathbf{W}_{r}$, $\mathbf{W}_{z}$, $\mathbf{U}$ and $\mathbf{W}$, respectively. At the time step $t$, $\bar{\mathbf{h}}^{t}$ represents the state update, $\mathbf{r}^{t}$ is the reset gate, and $\mathbf{z}^{t}$ is the update gate. The symbol $\odot$ denotes an element-wise product, and the activation function is denoted by $\sigma$. Note that ${\mathbf{h}}^{t}=0$ for $t=0$.

\noindent
\vspace{+0.1cm}\textbf{Level-controllable Noise Generator}

According to the assumption mentioned in the \textit{Introduction}, generating level-controllable noise, {\em i.e.}, BQ, will involve similarity-based ranking. However, existing textual similarity measures such as BLEU, CIDEr, METEOR, and ROUGE are not effective in capturing semantic similarity. To address this issue, we propose a new optimization-based ranking method in this work. We cast the problem of generating BQs that are similar to an MQ as a $LASSO$ optimization problem. By embedding all the main questions and the basic question candidates using Skip-thoughts, $LASSO$ modeling enables us to determine a sparse number of basic questions that are suitable to represent the given main question. The $LASSO$ model can be expressed as follows:
\vspace{-0.15cm}
\begin{equation}
    ~~~~~~~~~~~~~~~~~\min_{\mathbf{x}}~\frac{1}{2}\left \| \mathbf{A}\mathbf{x}-\mathbf{b} \right \|_{2}^{2}+\lambda \left \| \mathbf{x} \right \|_{1}, 
\label{eq:lasso}
\end{equation}
where $\lambda$ denotes a tradeoff parameter that controls the quality of BQs.


To create our basic question dataset (BQD), we combine the unique questions from the training and validation datasets of the popular VQA dataset \cite{4}, and use the testing dataset as our main question candidates. However, to ensure effective $LASSO$ modeling, we must preprocess the question sentences by ensuring that none of the main questions are already present in our basic question dataset. If any main questions are already in the BQD, it will result in an unhelpful ranking. Since we are encouraging sparsity, all other questions will be neglected with a similarity score of zero.

\noindent
\vspace{+0.1cm}\textbf{BQ Generation by $\mathbf{LASSO}$-based Ranking Method}

In this subsection, we outline the procedure for using the $LASSO$-based ranking method to generate basic questions corresponding to a given main question, as illustrated in Figure \ref{fig:figure100}. To obtain the sparse solution $\mathbf{x}$, we solve the $LASSO$ optimization problem, with the elements of $\mathbf{x}$ representing the similarity scores between the main question $\mathbf{b}$ and each corresponding BQ in $\mathbf{A}$. The BQ candidates are embedded using Skip-thoughts, and the top-$k$ BQs for a given MQ are selected based on the ranking of scores in $\mathbf{x}$. Higher similarity scores indicate greater similarity between the BQ and the MQ, and vice versa. Moreover, we note that VQA models tend to perform best on yes/no questions, which are comparatively simple. Consequently, we also generate a Yes/No Basic Question dataset using the aforementioned basic question generation approach for further experiments.

\noindent
\vspace{+0.1cm}\textbf{Details of the Proposed Basic Question Dataset for Robustness Analysis and In-context Learning}

We recognize that the size of the basic question dataset plays a crucial role in the effectiveness of the noise generation method. Generally, having a larger dataset increases the likelihood of finding similar questions to any given main question. With this in mind, we propose two large-scale basic question datasets, the General Basic Question Dataset and the Yes/No Basic Question Dataset, using the ${LASSO}$-based ranking method. We set $k=21$ to limit the number of top-ranked BQs to avoid having similarity scores that are too low. As a result, we obtain the ranked BQs of $244,302$ testing question candidates.

To analyze the robustness and enable in-context learning \cite{dai2022can,zhang2023multimodal,hu2022context,irie2022dual} for VQA models, we utilize the proposed General and Yes/No BQ datasets, which are structured as $\{\text{Image},~MQ,~21~(BQ + \text{corresponding~ similarity~score})\}$. These datasets comprise $81,434$ images from the testing images of MS COCO dataset \cite{51} and $244,302$ main questions from the testing questions of VQA dataset (open-ended task) \cite{4}, respectively. We generate the corresponding similarity scores of General and Yes/No BQ by our $LASSO$ ranking approach. Our General and Yes/No basic questions are extracted from the validation and training questions of the VQA dataset (open-ended task). In total, our GBQD and YNBQD contain $5,130,342$ (General BQ $+$ corresponding similarity score) tuples and $5,130,342$ (Yes/No BQ $+$ corresponding similarity score) tuples.

\noindent\vspace{+0.1cm}\textbf{Analyzing Robustness Using General and Yes/No Basic Questions with $R_{score}$}

To evaluate the robustness of a VQA model, it is important to measure how its accuracy is affected when its input is corrupted with noise. This noise can take various forms, such as random, structured, or semantically related to the final task. In VQA, the input consists of an MQ-image pair, and the noise can be introduced into both components. When injecting noise into the question, it is important to maintain some contextual semantics to ensure the measure is informative, rather than introducing misspellings or randomly changing or dropping words. In this study, we propose a novel measure of robustness for VQA by introducing semantically relevant noise to the questions, with the ability to control the level of noise. 

The VQA dataset \cite{4} provides both open-ended and multiple-choice tasks for evaluation, with the latter requiring the selection of an answer from $18$ candidates. For the former, the answer can be any phrase or word. In both cases, accuracy is used as the evaluation metric, as it is considered to reflect human consensus. We adopt the accuracy measure as defined in \cite{4}:




\begin{equation}
    \label{eq:acc-vqa}
    \text{Accuracy}_{VQA}=\frac{1}{N}\sum_{i=1}^{N}\min\left \{ \frac{\sum_{t\in T_{i}}\mathbb{I}[a_{i}=t]}{3},1 \right \},
\end{equation}
where $\mathbb{I}[\cdot]$ is an indicator function, while $N$ is the total number of examples. $a_{i}$ is the predicted answer, and $T_{i}$ is the answer set of the $i^{th}$ image-question pair. For a predicted answer to be considered correct, it must have the agreement of at least three annotators. When the predicted answer is incorrect, the score depends on the total number of agreements.


\begin{equation}
    \label{eq:absolute-difference}
    ~~~~~~~~~~~~~~~~~Acc_{di} = \left|Acc_{vqa}-Acc_{bqd}\right|,
\end{equation}
where $Acc_{vqa}$ and $Acc_{bqd}$ are calculated based on Equation (\ref{eq:acc-vqa}).

\begin{SCfigure}
  \scalebox{0.9}{
  \includegraphics[width=0.46 \linewidth]{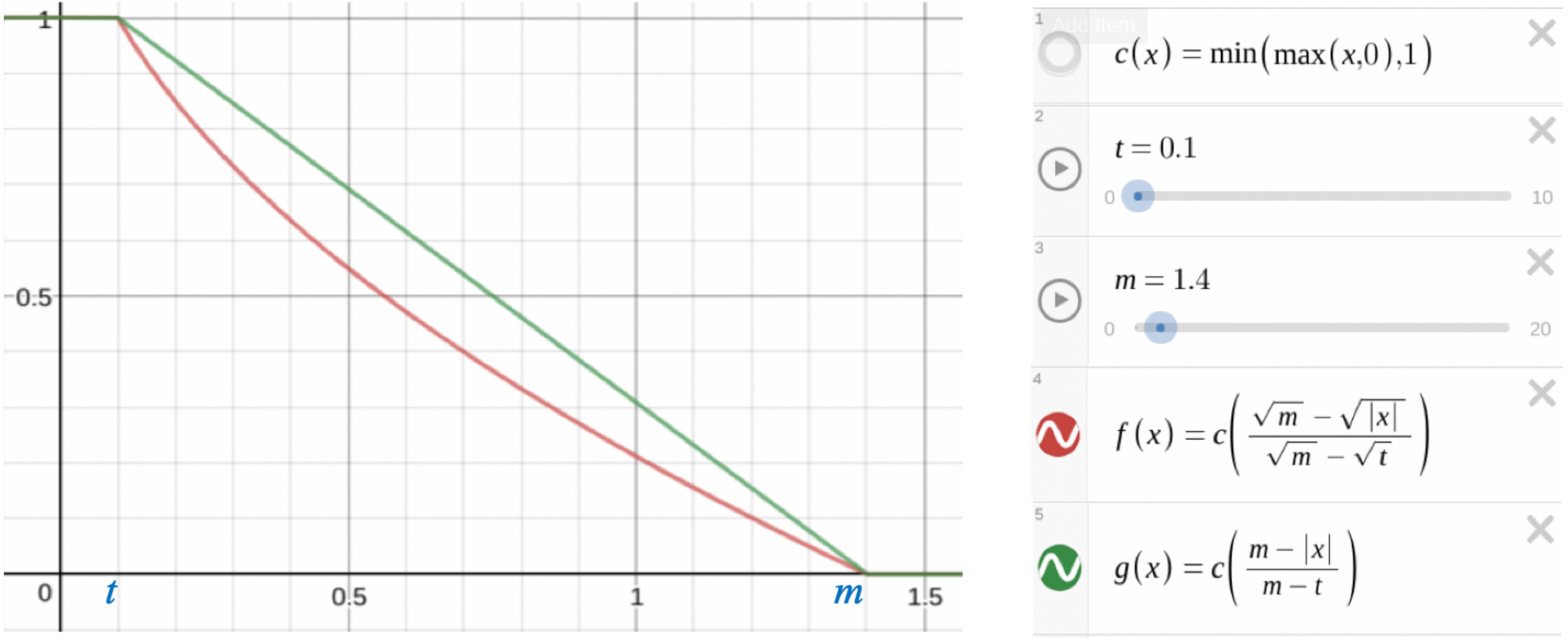}}
  \caption{A visualization of the $R_{score}$ function is denoted in red. The right-hand part of the $f(x)$ and $g(x)$ is plotted for convenience. The tolerance and maximum robustness limit are represented by $t$ and $m$, respectively. For further explanation, please see the section \textit{Analyzing Robustness Using General and Yes/No Basic Questions with $R_{score}$}.
  }
\label{fig:figure71}
\end{SCfigure}

To assess the robustness of a VQA model, we begin by computing its accuracy on the clean VQA dataset \cite{4}, denoted by $Acc_{vqa}$. We then introduce noise into each question-answer pair by appending the top-ranked $k$ BQs to the original question MQ, and re-evaluate the model's accuracy on this noisy input, denoted by $Acc_{bqd}$. Next, we compute the absolute difference between $Acc_{vqa}$ and $Acc_{bqd}$ using Equation (\ref{eq:absolute-difference}) to obtain $Acc_{di}$, which we use to compute the robustness score $R_{score}$. The parameters $t$ and $m$ in Equation (\ref{eq:robustness-cleaner}) represent the tolerance and maximum robustness limit, respectively. We aim to make the score sensitive to small differences in $Acc_{di}$, but only above $t$, and less sensitive for larger differences, but only below $m$. Therefore, $R_{score}$ is designed to smoothly decrease from $1$ to $0$ as $Acc_{di}$ varies from $t$ to $m$, with the rate of change transitioning from exponential to sublinear within the range $[t, m]$.


\begin{equation}
    \label{eq:robustness-cleaner}
    ~~~~~~~~~~~~R_{score} = clamp_{0}^{1}\left(\frac{\sqrt{m}-\sqrt{{Acc_{di}}}}{\sqrt{m}-\sqrt{t}}\right)
\end{equation}

\begin{equation}
    \label{eq:robustness-clamp}
    ~~~~~~~~~~~~~~~~clamp_{a}^{b}(x) = \max\left(a,{\min\left(b,x\right)}\right),
\end{equation}
where $0 \leq t < m \leq 100$. To provide a better understanding, we present a visualization of the $R_{score}$ function in Figure \ref{fig:figure71}.

\section{Experiments}
This section presents the implementation details and experiments performed to validate and analyze the proposed method.


\begin{table*}
\centering 
\begin{subtable}{0.75\linewidth}
\resizebox{\columnwidth}{110.0pt}{
\scalebox{1}{
\begin{tabular}{|c|c|c|}
\hline
BQ ID & \multicolumn{1}{l|}{Similarity Score} & BQ \\ \hline
\begin{tabular}[c]{@{}c@{}}01\\
02\\
03\end{tabular}  & \begin{tabular}[c]{@{}c@{}}0.295\\
0.240\\
0.142\end{tabular} & \begin{tabular}[c]{@{}c@{}}How old is the truck? \\    
How old is this car? \\   
How old is the vehicle?\end{tabular} \\
\hline
\begin{tabular}[c]{@{}c@{}}04\\
05\\ 
06\end{tabular} & \begin{tabular}[c]{@{}c@{}}0.120\\
0.093\\ 
0.063\end{tabular} & \begin{tabular}[c]{@{}c@{}}What number is the car?\\ 
What color is the car?\\ 
How old is the bedroom?\end{tabular}\\
\hline
\begin{tabular}[c]{@{}c@{}}07\\
08\\ 
09\end{tabular} & \begin{tabular}[c]{@{}c@{}}0.063\\ 
0.037\\ 
0.033\end{tabular} & \begin{tabular}[c]{@{}c@{}}What year is the car?\\
Where is the old car?\\
How old is the seat?\end{tabular}\\ 
\hline
\begin{tabular}[c]{@{}c@{}}10\\
11\\ 
12\end{tabular}  & \begin{tabular}[c]{@{}c@{}}0.032\\ 
0.028\\ 
0.028\end{tabular} & \begin{tabular}[c]{@{}c@{}}How old is the cart?\\ 
What make is the blue car?\\ 
How old is the golden retriever?\end{tabular}\\ 
\hline
\begin{tabular}[c]{@{}c@{}}13\\
14\\ 
15\end{tabular}  & \begin{tabular}[c]{@{}c@{}}0.024\\ 
0.022\\ 
0.020\end{tabular} & \begin{tabular}[c]{@{}c@{}}What is beneath the car?\\ 
Is the car behind him a police car?\\ 
How old is the pilot?\end{tabular}\\ 
\hline
\begin{tabular}[c]{@{}c@{}}16\\
17\\ 
18\end{tabular} & \begin{tabular}[c]{@{}c@{}}0.017\\ 0.016\\ 
0.016\end{tabular} & \begin{tabular}[c]{@{}c@{}}How old are you?\\ 
How old is the laptop?\\ 
How old is the television?\end{tabular}\\ 
\hline
\begin{tabular}[c]{@{}c@{}}19\\
20\\ 
21\end{tabular} & \begin{tabular}[c]{@{}c@{}}0.015\\ 0.015\\ 
0.015\end{tabular} & \begin{tabular}[c]{@{}c@{}}What make is the main car?\\ 
What type and model is the car?\\ 
What is lifting the car?\end{tabular}\\ 
\hline

\end{tabular}}}
\caption{``MQ: How old is the car?'' and image ``(a)'' in Figure \ref{fig:figure10}.}
\label{table:table51}

\vspace{0.2cm}

\resizebox{\columnwidth}{130.0pt}{
\scalebox{1}{
\begin{tabular}{|c|c|c|}
\hline
BQ ID  & \multicolumn{1}{l|}{Similarity Score} & BQ \\ \hline
\begin{tabular}[c]{@{}c@{}}01\\
02\\
03\end{tabular}  & \begin{tabular}[c]{@{}c@{}}0.281\\
0.108\\
0.055\end{tabular} & \begin{tabular}[c]{@{}c@{}}Where is the cat sitting on? \\    
What is this cat sitting on? \\   
What is cat sitting on?\end{tabular} \\
\hline
\begin{tabular}[c]{@{}c@{}}04\\
05\\ 
06\end{tabular} & \begin{tabular}[c]{@{}c@{}}0.053\\
0.050\\ 
0.047\end{tabular} & \begin{tabular}[c]{@{}c@{}}What is the cat on the left sitting on?\\ 
What is the giraffe sitting on?\\ 
What is the cat sitting in the car?\end{tabular}\\
\hline
\begin{tabular}[c]{@{}c@{}}07\\
08\\ 
09\end{tabular} & \begin{tabular}[c]{@{}c@{}}0.046\\ 
0.042\\ 
0.041\end{tabular} & \begin{tabular}[c]{@{}c@{}}That is the black cat sitting on?\\
What is the front cat sitting on?\\
What is the cat perched on?\end{tabular}\\ 
\hline
\begin{tabular}[c]{@{}c@{}}10\\
11\\ 
12\end{tabular}  & \begin{tabular}[c]{@{}c@{}}0.041\\ 
0.037\\ 
0.035\end{tabular} & \begin{tabular}[c]{@{}c@{}}What's the cat sitting on? \\ 
What is the cat leaning on? \\ 
What object is the cat sitting on?\end{tabular}\\ 
\hline
\begin{tabular}[c]{@{}c@{}}13\\
14\\ 
15\end{tabular}  & \begin{tabular}[c]{@{}c@{}}0.029\\ 
0.023\\ 
0.022\end{tabular} & \begin{tabular}[c]{@{}c@{}}What is the doll sitting on?\\ 
How is the cat standing?\\ 
What is the cat setting on?\end{tabular}\\ 
\hline
\begin{tabular}[c]{@{}c@{}}16\\
17\\ 
18\end{tabular} & \begin{tabular}[c]{@{}c@{}}0.022\\ 0.021\\ 
0.021\end{tabular} & \begin{tabular}[c]{@{}c@{}}What is the cat walking on?\\ 
What is the iPhone sitting on?\\ 
What is the cat napping on?\end{tabular}\\ 
\hline
\begin{tabular}[c]{@{}c@{}}19\\
20\\ 
21\end{tabular} & \begin{tabular}[c]{@{}c@{}}0.020\\ 0.018\\ 
0.018\end{tabular} & \begin{tabular}[c]{@{}c@{}}What is the dog sitting at?\\ 
What is the birds sitting on?\\ 
What is the sitting on?\end{tabular}\\ 
\hline

\end{tabular}}}
\caption{``MQ: What is the cat sitting on?'' and image ``(b)'' in Figure \ref{fig:figure10}.}
\label{table:table52}

\end{subtable}
\caption{``MQ: How old is the car?'' and image ``(a)'' corresponds to Figure \ref{fig:figure10}-(a). ``MQ: What is the cat sitting on?'' and image ``(b)'' corresponds to Figure \ref{fig:figure10}-(b).}
\label{table:table5}
\vspace{-0.5cm}
\end{table*}


\begin{table*}
\renewcommand\arraystretch{1.0}
\setlength\tabcolsep{11pt}
    \centering
\begin{subtable}[t]{0.3\linewidth}
\centering
\scalebox{0.5}{
    \begin{tabular}{ c | c c c c | c} 
     Task Type &    & \multicolumn{3}{c}{Open-Ended} &  \\ [0.5ex]
     \hline
     Method &    & \multicolumn{3}{c}{MUTAN without Attention} &  \\ [0.5ex]
     \hline
     Test Set&  \multicolumn{4}{c}{dev} & diff \\ [0.5ex]
     \hline
     Partition & Other & Num & Y/N & All & All \\ [0.5ex] 
     \hline
     First-dev & 37.78 & 34.93 & 68.20 & \textbf{49.96} & \textbf{10.20}  \\ 
     
     Second-dev & 37.29 & 35.03 & 65.62 & \textbf{48.67} & \textbf{11.49} \\
     
     Third-dev & 34.81 & 34.39 & 62.85 & \textbf{46.27} & \textbf{13.89}  \\
     
     Fourth-dev & 34.25 & 34.29 & 63.60 & \textbf{46.30} & \textbf{13.86}  \\
     
     Fifth-dev & 33.89 & 34.66 & 64.19 & \textbf{46.41} & \textbf{13.75} \\
     
     Sixth-dev & 33.15 & 34.68 & 64.59 & \textbf{46.22} & \textbf{13.94}  \\
     
     Seventh-dev & 32.80 & 33.99 & 63.59 & \textbf{45.57} & \textbf{14.59}  \\
     \hline
     First-std & 38.24 & 34.54 & 67.55 & \textbf{49.93} & \textbf{10.52} \\
     \hline
     
     Original-dev  & 47.16 & 37.32 & 81.45 & \textbf{60.16} & -\\
     Original-std  & 47.57 & 36.75 & 81.56 & \textbf{60.45} & - \\

     \hline
    \end{tabular}}
    \centering
    \captionsetup{justification=centering}
    \caption{MUTAN without Attention model evaluation results.}

\scalebox{0.5}{
    \begin{tabular}{ c | c c c c | c} 
     Task Type &    & \multicolumn{3}{c}{Open-Ended} &  \\ [0.5ex]
     \hline
     Method &    & \multicolumn{3}{c}{HieCoAtt (Alt,VGG19)} &  \\ [0.5ex]
     \hline
     Test Set&  \multicolumn{4}{c}{dev} & diff \\ [0.5ex]
     \hline
     Partition & Other & Num & Y/N & All & All \\ [0.5ex] 
     \hline
     First-dev & 44.44 & 37.53 & 71.11 & \textbf{54.63} & \textbf{5.85}  \\ 
     
     Second-dev & 42.62 & 36.68 & 68.67 & \textbf{52.67} & \textbf{7.81} \\
     
     Third-dev & 41.60 & 35.59 & 66.28 & \textbf{51.08} & \textbf{9.4}  \\
     
     Fourth-dev & 41.09 & 35.71 & 67.49 & \textbf{51.34} & \textbf{9.14}  \\
     
     Fifth-dev & 39.83 & 35.55 & 65.72 & \textbf{49.99} & \textbf{10.49} \\
     
     Sixth-dev & 39.60 & 35.99 & 66.56 & \textbf{50.27} & \textbf{10.21}  \\
     
     Seventh-dev & 38.33 & 35.47 & 64.89 & \textbf{48.92} & \textbf{11.56}  \\
     \hline
     First-std & 44.77 & 36.08 & 70.67 & \textbf{54.54} & \textbf{5.78} \\
     \hline

     Original-dev  & 49.14 & 38.35 & 79.63 & \textbf{60.48} & -\\
     Original-std  & 49.15 & 36.52 & 79.45 & \textbf{60.32} & - \\

     \hline
    \end{tabular}}
    \centering
    \captionsetup{justification=centering}
    \caption{HieCoAtt (Alt,VGG19) model evaluation results.}
\end{subtable}%
    \hfil
\begin{subtable}[t]{0.3\linewidth}

\centering
\scalebox{0.5}{
    \begin{tabular}{ c | c c c c | c} 
     Task Type &    & \multicolumn{3}{c}{Open-Ended} &  \\ [0.5ex]
     \hline
     Method &    & \multicolumn{3}{c}{MLB with Attention} &  \\ [0.5ex]
     \hline
     Test Set&  \multicolumn{4}{c}{dev} & diff \\ [0.5ex]
     \hline
     Partition & Other & Num & Y/N & All & All \\ [0.5ex] 
     \hline
     First-dev & 49.31 & 34.62 & 72.21 & \textbf{57.12} & \textbf{8.67}  \\ 
     
     Second-dev & 48.53 & 34.84 & 70.30 & \textbf{55.98} & \textbf{9.81} \\
     
     Third-dev & 48.01 & 33.95 & 69.15 & \textbf{55.16} & \textbf{10.63}  \\
     
     Fourth-dev & 47.20 & 34.02 & 69.31 & \textbf{54.84} & \textbf{10.95}  \\
     
     Fifth-dev & 45.85 & 34.07 & 68.95 & \textbf{54.05} & \textbf{11.74} \\
     
     Sixth-dev & 44.61 & 34.30 & 68.59 & \textbf{53.34} & \textbf{12.45}  \\
     
     Seventh-dev & 44.71 & 33.84 & 67.76 & \textbf{52.99} & \textbf{12.80}  \\
     \hline
     First-std & 49.07 & 34.13 & 71.96 & \textbf{56.95} & \textbf{8.73} \\
     \hline
     
     Original-dev  & 57.01 & 37.51 & 83.54 & \textbf{65.79} & -\\
     Original-std  & 56.60 & 36.63 & 83.68 & \textbf{65.68} & - \\

     \hline
    \end{tabular}}
    \centering
    \captionsetup{justification=centering}
    \caption{MLB with Attention model evaluation results.}
    
\scalebox{0.5}{
    \begin{tabular}{ c | c c c c | c} 
     Task Type &    & \multicolumn{3}{c}{Open-Ended} &  \\ [0.5ex]
     \hline
     Method &    & \multicolumn{3}{c}{MUTAN with Attention} &  \\ [0.5ex]
     \hline
     Test Set&  \multicolumn{4}{c}{dev} & diff \\ [0.5ex]
     \hline
     Partition & Other & Num & Y/N & All & All \\ [0.5ex] 
     \hline
     First-dev & 51.51 & 35.62 & 68.72 & \textbf{56.85} & \textbf{9.13}  \\ 
     
     Second-dev & 49.86 & 34.43 & 66.18 & \textbf{54.88} & \textbf{11.10} \\
     
     Third-dev & 49.15 & 34.50 & 64.85 & \textbf{54.00} & \textbf{11.98}  \\
     
     Fourth-dev & 47.96 & 34.26 & 64.72 & \textbf{53.35} & \textbf{12.63}  \\
     
     Fifth-dev & 47.20 & 33.93 & 64.53 & \textbf{52.88} & \textbf{13.10} \\
     
     Sixth-dev & 46.48 & 33.90 & 64.37 & \textbf{52.46} & \textbf{13.52}  \\
     
     Seventh-dev & 46.88 & 33.13 & 64.10 & \textbf{52.46} & \textbf{13.52}  \\
     \hline
     First-std & 51.34 & 35.22 & 68.32 & \textbf{56.66} & \textbf{9.11} \\
     \hline
     
     Original-dev  & 56.73 & 38.35 & 84.11 & \textbf{65.98} & -\\
     Original-std  & 56.29 & 37.47 & 84.04 & \textbf{65.77} & - \\

     \hline
    \end{tabular}}
    \centering
    \captionsetup{justification=centering}
    \caption{MUTAN with Attention model evaluation results.}
\end{subtable}%
    \hfil
\begin{subtable}[t]{0.3\linewidth}
        
\centering
\scalebox{0.5}{
    \begin{tabular}{ c | c c c c | c} 
     Task Type &    & \multicolumn{3}{c}{Open-Ended} &  \\ [0.5ex]
     \hline
     Method &    & \multicolumn{3}{c}{HieCoAtt (Alt,Resnet200)} &  \\ [0.5ex]
     \hline
     Test Set&  \multicolumn{4}{c}{dev} & diff \\ [0.5ex]
     \hline
     Partition & Other & Num & Y/N & All & All \\ [0.5ex] 
     \hline
     First-dev & 46.51 & 36.33 & 70.41 & \textbf{55.22} & \textbf{6.59}  \\ 
     
     Second-dev & 45.19 & 36.78 & 67.27 & \textbf{53.34} & \textbf{8.47} \\
     
     Third-dev & 43.87 & 36.28 & 65.29 & \textbf{51.84} & \textbf{9.97}  \\
     
     Fourth-dev & 43.41 & 36.25 & 65.94 & \textbf{51.88} & \textbf{9.93}  \\
     
     Fifth-dev & 42.02 & 35.89 & 66.09 & \textbf{51.23} & \textbf{10.58} \\
     
     Sixth-dev & 42.03 & 36.40 & 65.66 & \textbf{51.12} & \textbf{10.69}  \\
     
     Seventh-dev & 40.68 & 36.08 & 63.49 & \textbf{49.54} & \textbf{12.27}  \\
     \hline
     First-std & 46.77 & 35.22 & 69.66 & \textbf{55.00} & \textbf{7.06}\\
     \hline
     
     Original-dev  & 51.77 & 38.65 & 79.70 & \textbf{61.81} & -\\
     Original-std  & 51.95 & 38.22 & 79.95 & \textbf{62.06} & - \\

     \hline
    \end{tabular}}
    \centering
    \captionsetup{justification=centering}
    \caption{HieCoAtt (Alt,Resnet200) model evaluation results.}

\scalebox{0.5}{
    \begin{tabular}{ c | c c c c | c} 
     Task Type &    & \multicolumn{3}{c}{Open-Ended} &  \\ [0.5ex]
     \hline
     Method &    & \multicolumn{3}{c}{LSTM Q+I} &  \\ [0.5ex]
     \hline
     Test Set&  \multicolumn{4}{c}{dev} & diff \\ [0.5ex]
     \hline
     Partition & Other & Num & Y/N & All & All \\ [0.5ex] 
     \hline
     First-dev & 29.24 & 33.77 & 65.14 & \textbf{44.47} & \textbf{13.55}  \\ 
     
     Second-dev & 28.02 & 32.73 & 62.68 & \textbf{42.75} & \textbf{15.27} \\
     
     Third-dev & 26.32 & 33.10 & 60.22 & \textbf{40.97} & \textbf{17.05}  \\
     
     Fourth-dev & 25.27 & 31.70 & 61.56 & \textbf{40.86} & \textbf{17.16}  \\
     
     Fifth-dev & 24.73 & 32.63 & 61.55 & \textbf{40.70} & \textbf{17.32} \\
     
     Sixth-dev & 23.90 & 32.14 & 61.42 & \textbf{40.19} & \textbf{17.83}  \\
     
     Seventh-dev & 22.74 & 31.36 & 60.60 & \textbf{39.21} & \textbf{18.81}  \\
     \hline
     First-std & 29.68 & 33.76 & 65.09 & \textbf{44.70} & \textbf{13.48} \\
     \hline
     
     Original-dev  & 43.40 & 36.46 & 80.87 & \textbf{58.02} & -\\
     Original-std  & 43.90 & 36.67 & 80.38 & \textbf{58.18} & - \\

     \hline
    \end{tabular}}
    \centering
    \captionsetup{justification=centering}
    \caption{LSTM Q+I model evaluation results.}
\end{subtable}
\caption{The table shows the six state-of-the-art pretrained VQA models evaluation results on the GBQD and VQA dataset. ``-'' indicates the results are not available, ``-std'' represents the accuracy of VQA model evaluated on the complete testing set of GBQD and VQA dataset and ``-dev'' indicates the accuracy of VQA model evaluated on the partial testing set of GBQD and VQA dataset. In addition, $diff = Original_{dev_{All}} - X_{dev_{All}}$, where $X$ is equal to the ``First'', ``Second'', etc.}
\label{table:table6}
\end{table*}

\begin{table*}
\renewcommand\arraystretch{1.0}
\setlength\tabcolsep{11pt}
    \centering
\begin{subtable}[t]{0.3\linewidth}
\centering
\scalebox{0.5}{
    \begin{tabular}{ c | c c c c | c} 
     Task Type &    & \multicolumn{3}{c}{Open-Ended} &  \\ [0.5ex]
     \hline
     Method &    & \multicolumn{3}{c}{MUTAN without Attention} &  \\ [0.5ex]
     \hline
     Test Set&  \multicolumn{4}{c}{dev} & diff \\ [0.5ex]
     \hline
     Partition & Other & Num & Y/N & All & All \\ [0.5ex] 
     \hline
     First-dev & 33.98 & 33.50 & 73.22 & \textbf{49.96} & \textbf{10.13}  \\ 
     
     Second-dev & 32.44 & 34.47 & 72.22 & \textbf{48.67} & \textbf{11.18} \\
     
     Third-dev & 32.65 & 33.60 & 71.76 & \textbf{46.27} & \textbf{11.36}  \\
     
     Fourth-dev & 32.77 & 33.79 & 71.14 & \textbf{46.30} & \textbf{11.53}  \\
     
     Fifth-dev & 32.46 & 33.51 & 70.90 & \textbf{46.41} & \textbf{11.81} \\
     
     Sixth-dev & 33.02 & 33.18 & 69.88 & \textbf{46.22} & \textbf{12.00}  \\
     
     Seventh-dev & 32.73 & 33.28 & 69.74 & \textbf{45.57} & \textbf{12.18}  \\
     \hline
     First-std & 34.06 & 33.24 & 72.99 & \textbf{49.93} & \textbf{10.43} \\
     \hline
     
     Original-dev  & 47.16 & 37.32 & 81.45 & \textbf{60.16} & -\\
     Original-std  & 47.57 & 36.75 & 81.56 & \textbf{60.45} & - \\

     \hline
    \end{tabular}}
    \centering
    \captionsetup{justification=centering}
    \caption{MUTAN without Attention model evaluation results.}

\scalebox{0.5}{
    \begin{tabular}{ c | c c c c | c} 
     Task Type &    & \multicolumn{3}{c}{Open-Ended} &  \\ [0.5ex]
     \hline
     Method &    & \multicolumn{3}{c}{HieCoAtt (Alt,VGG19)} &  \\ [0.5ex]
     \hline
     Test Set&  \multicolumn{4}{c}{dev} & diff \\ [0.5ex]
     \hline
     Partition & Other & Num & Y/N & All & All \\ [0.5ex] 
     \hline
     First-dev & 40.80 & 30.34 & 76.92 & \textbf{54.49} & \textbf{5.99}  \\ 
     
     Second-dev & 39.63 & 30.67 & 76.49 & \textbf{53.78} & \textbf{6.70} \\
     
     Third-dev & 39.33 & 31.12 & 75.48 & \textbf{53.28} & \textbf{7.20}  \\
     
     Fourth-dev & 39.31 & 29.78 & 75.12 & \textbf{52.97} & \textbf{7.51}  \\
     
     Fifth-dev & 39.38 & 29.87 & 74.96 & \textbf{52.95} & \textbf{7.53} \\
     
     Sixth-dev & 39.13 & 30.74 & 73.95 & \textbf{52.51} & \textbf{7.97}  \\
     
     Seventh-dev & 38.90 & 31.14 & 73.80 & \textbf{52.39} & \textbf{8.09}  \\
     \hline
     First-std & 40.88 & 28.82 & 76.67 & \textbf{54.37} & \textbf{5.95} \\
     \hline

     Original-dev  & 49.14 & 38.35 & 79.63 & \textbf{60.48} & -\\
     Original-std  & 49.15 & 36.52 & 79.45 & \textbf{60.32} & - \\

     \hline
    \end{tabular}}
    \centering
    \captionsetup{justification=centering}
    \caption{HieCoAtt (Alt,VGG19) model evaluation results.}
\end{subtable}%
    \hfil
\begin{subtable}[t]{0.3\linewidth}

\centering
\scalebox{0.5}{
    \begin{tabular}{ c | c c c c | c} 
     Task Type &    & \multicolumn{3}{c}{Open-Ended} &  \\ [0.5ex]
     \hline
     Method &    & \multicolumn{3}{c}{MLB with Attention} &  \\ [0.5ex]
     \hline
     Test Set&  \multicolumn{4}{c}{dev} & diff \\ [0.5ex]
     \hline
     Partition & Other & Num & Y/N & All & All \\ [0.5ex] 
     \hline
     First-dev & 46.57 & 32.09 & 76.60 & \textbf{57.33} & \textbf{8.46}  \\ 
     
     Second-dev & 45.83 & 32.43 & 75.29 & \textbf{56.47} & \textbf{9.32} \\
     
     Third-dev & 45.17 & 32.52 & 74.87 & \textbf{55.99} & \textbf{9.80}  \\
     
     Fourth-dev & 45.11 & 32.31 & 73.73 & \textbf{55.47} & \textbf{10.32}  \\
     
     Fifth-dev & 44.35 & 31.95 & 72.93 & \textbf{54.74} & \textbf{11.05} \\
     
     Sixth-dev & 43.75 & 31.21 & 72.03 & \textbf{54.00} & \textbf{11.79}  \\
     
     Seventh-dev & 43.88 & 32.59 & 71.99 & \textbf{54.19} & \textbf{11.60}  \\
     \hline
     First-std & 46.11 & 31.46 & 76.84 & \textbf{57.25} & \textbf{8.43} \\
     \hline
     
     Original-dev  & 57.01 & 37.51 & 83.54 & \textbf{65.79} & -\\
     Original-std  & 56.60 & 36.63 & 83.68 & \textbf{65.68} & - \\

     \hline
    \end{tabular}}
    \centering
    \captionsetup{justification=centering}
    \caption{MLB with Attention model evaluation results.}
    
\scalebox{0.5}{
    \begin{tabular}{ c | c c c c | c} 
     Task Type &    & \multicolumn{3}{c}{Open-Ended} &  \\ [0.5ex]
     \hline
     Method &    & \multicolumn{3}{c}{MUTAN with Attention} &  \\ [0.5ex]
     \hline
     Test Set&  \multicolumn{4}{c}{dev} & diff \\ [0.5ex]
     \hline
     Partition & Other & Num & Y/N & All & All \\ [0.5ex] 
     \hline
     First-dev & 43.96 & 28.90 & 71.89 & \textbf{53.79} & \textbf{12.19}  \\ 
     
     Second-dev & 42.66 & 28.08 & 70.05 & \textbf{52.32} & \textbf{13.66} \\
     
     Third-dev & 41.62 & 29.12 & 69.58 & \textbf{51.74} & \textbf{14.24}  \\
     
     Fourth-dev & 41.53 & 29.30 & 67.96 & \textbf{51.06} & \textbf{14.92}  \\
     
     Fifth-dev & 40.46 & 27.66 & 68.03 & \textbf{50.39} & \textbf{15.59} \\
     
     Sixth-dev & 40.03 & 28.44 & 66.98 & \textbf{49.84} & \textbf{16.14}  \\
     
     Seventh-dev & 39.11 & 28.41 & 67.44 & \textbf{49.58} & \textbf{16.40}  \\
     \hline
     First-std & 43.55 & 28.70 & 71.76 & \textbf{53.63} & \textbf{12.14} \\
     \hline
     
     Original-dev  & 56.73 & 38.35 & 84.11 & \textbf{65.98} & -\\
     Original-std  & 56.29 & 37.47 & 84.04 & \textbf{65.77} & - \\

     \hline
    \end{tabular}}
    \centering
    \captionsetup{justification=centering}
    \caption{MUTAN with Attention model evaluation results.}
\end{subtable}%
    \hfil
\begin{subtable}[t]{0.3\linewidth}
        
\centering
\scalebox{0.5}{
    \begin{tabular}{ c | c c c c | c} 
     Task Type &    & \multicolumn{3}{c}{Open-Ended} &  \\ [0.5ex]
     \hline
     Method &    & \multicolumn{3}{c}{HieCoAtt (Alt,Resnet200)} &  \\ [0.5ex]
     \hline
     Test Set&  \multicolumn{4}{c}{dev} & diff \\ [0.5ex]
     \hline
     Partition & Other & Num & Y/N & All & All \\ [0.5ex] 
     \hline
     First-dev & 44.42 & 36.39 & 76.94 & \textbf{56.90} & \textbf{4.91}  \\ 
     
     Second-dev & 43.37 & 34.99 & 76.10 & \textbf{55.90} & \textbf{5.91} \\
     
     Third-dev & 42.22 & 33.97 & 75.80 & \textbf{55.11} & \textbf{6.70}  \\
     
     Fourth-dev & 42.52 & 34.21 & 75.33 & \textbf{55.09} & \textbf{6.72}  \\
     
     Fifth-dev & 42.81 & 34.69 & 75.21 & \textbf{55.23} & \textbf{6.58} \\
     
     Sixth-dev & 42.27 & 35.16 & 74.50 & \textbf{54.73} & \textbf{7.08}  \\
     
     Seventh-dev & 41.95 & 35.14 & 73.64 & \textbf{54.22} & \textbf{7.59}  \\
     \hline
     First-std & 44.93 & 35.59 & 76.82 & \textbf{57.10} & \textbf{4.96}\\
     \hline

     Original-dev  & 51.77 & 38.65 & 79.70 & \textbf{61.81} & -\\
     Original-std  & 51.95 & 38.22 & 79.95 & \textbf{62.06} & - \\

     \hline
    \end{tabular}}
    \centering
    \captionsetup{justification=centering}
    \caption{HieCoAtt (Alt,Resnet200) model evaluation results.}

\scalebox{0.5}{
    \begin{tabular}{ c | c c c c | c} 
     Task Type &    & \multicolumn{3}{c}{Open-Ended} &  \\ [0.5ex]
     \hline
     Method &    & \multicolumn{3}{c}{LSTM Q+I} &  \\ [0.5ex]
     \hline
     Test Set&  \multicolumn{4}{c}{dev} & diff \\ [0.5ex]
     \hline
     Partition & Other & Num & Y/N & All & All \\ [0.5ex] 
     \hline
     First-dev & 20.49 & 25.98 & 68.79 & \textbf{40.91} & \textbf{17.11}  \\ 
     
     Second-dev & 19.81 & 25.40 & 68.51 & \textbf{40.40} & \textbf{17.62} \\
     
     Third-dev & 18.58 & 24.95 & 68.53 & \textbf{39.77} & \textbf{18.25}  \\
     
     Fourth-dev & 18.50 & 24.82 & 67.83 & \textbf{39.43} & \textbf{18.59}  \\
     
     Fifth-dev & 17.68 & 24.68 & 67.99 & \textbf{39.09} & \textbf{18.93} \\
     
     Sixth-dev & 17.29 & 24.03 & 67.76 & \textbf{38.73} & \textbf{19.29}  \\
     
     Seventh-dev & 16.93 & 24.63 & 67.45 & \textbf{38.50} & \textbf{19.52}  \\
     \hline
     First-std & 20.84 & 26.14 & 68.88 & \textbf{41.19} & \textbf{16.99} \\
     \hline
     
     Original-dev  & 43.40 & 36.46 & 80.87 & \textbf{58.02} & -\\
     Original-std  & 43.90 & 36.67 & 80.38 & \textbf{58.18} & - \\

     \hline
    \end{tabular}}
    \centering
    \captionsetup{justification=centering}
    \caption{LSTM Q+I model evaluation results.}
\end{subtable}
\caption{The table shows the six state-of-the-art pretrained VQA models evaluation results on the YNBQD and VQA dataset. ``-'' indicates the results are not available, ``-std'' represents the accuracy of VQA model evaluated on the complete testing set of YNBQD and VQA dataset and ``-dev'' indicates the accuracy of VQA model evaluated on the partial testing set of YNBQD and VQA dataset. In addition, $diff = Original_{dev_{All}} - X_{dev_{All}}$, where $X$ is equal to the ``First'', ``Second'', etc.}
\label{table:table7}
\end{table*}

\begin{figure*}[t]
  \begin{subfigure}[b]{0.49\textwidth}
    \includegraphics[width=0.98\linewidth]{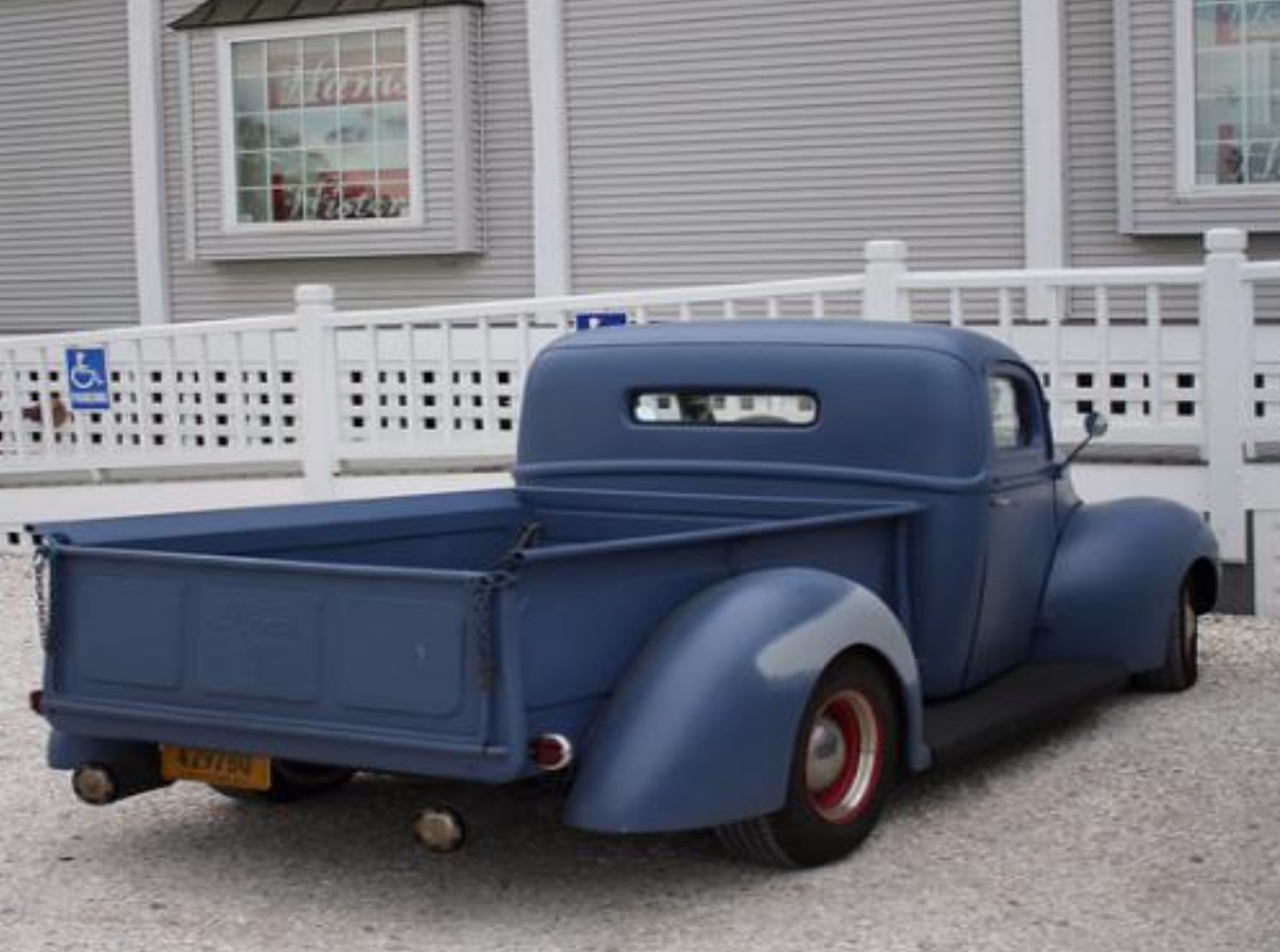}
    \caption{}
  \end{subfigure}
  \hfill
  \begin{subfigure}[b]{0.49\textwidth}
    \includegraphics[width=0.98\linewidth]{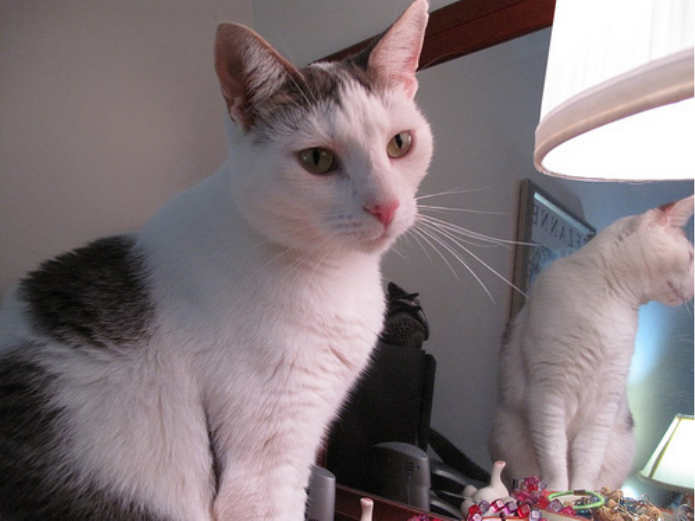}
    \caption{}
  \end{subfigure}
  \caption{Image ``(a)'' corresponds to Table \ref{table:table5}-(a). Image ``(b)'' corresponds to Table \ref{table:table5}-(b).}
\label{fig:figure10}
\end{figure*}

\vspace{3pt}\noindent\textbf{Dataset.~}

We performed experiments on the GBQD, YNBQD, and VQA datasets \cite{4}. The VQA dataset is based on the MS COCO dataset \cite{51} and comprises $248,349$ training, $121,512$ validation, and $244,302$ testing questions. Each question in the VQA dataset has ten associated answers annotated by different individuals on AMT (Amazon Mechanical Turk). Nearly $90\%$ of the answers have a single word, and $98\%$ of the answers are no more than three words long. Please refer to the \textit{Details of the Proposed Basic Question Dataset for Robustness Analysis and In-context Learning} section for further information on GBQD and YNBQD. To gain a better understanding of the datasets, we provide some examples in Table \ref{table:table5}.


\vspace{3pt}\noindent\textbf{Setup.~}

We utilize the Skip-thought Vector to encode all the training and validation questions of the VQA dataset into the columns of $\mathbf{A} \in \mathbf{R}^{4800 \times 186027}$, and represent the given main question as $\mathbf{b} \in \mathbf{R}^{4800}$. For generating our General and Yes/No BQ Datasets, we set $\lambda = 10^{-6}$ to ensure a better quality of BQs. We collect only the top $21$ ranked General and Yes/No BQs, as similarity scores beyond this limit are insignificant, and use them to create our GBQD and YNBQD. Since many state-of-the-art VQA models are trained under the assumption of a maximum of $26$ input words, we divide the $21$ top-ranked BQs into seven consecutive partitions, {\em i.e.}, $21=3*7$, for robustness analysis, as shown in Table \ref{table:table6} for GBQD and Table \ref{table:table7} for YNBQD. Note that each MQ with three BQs contains a total number of words equal to or less than $26$, under this setting.


\vspace{3pt}\noindent\textbf{BQ Generation by Popular Text Evaluation Metrics.~}

In this subsection, we compare the performance of the proposed $LASSO$-based ranking method with the non-$LASSO$-based ranking methods for generating BQs of a given MQ. Specifically, we consider seven popular sentence evaluation metrics \cite{49,67,68,69}, including BLEU-1, BLEU-2, BLEU-3, BLEU-4, ROUGE, CIDEr, and METEOR, which are commonly used to measure the similarity score between MQ and BQs. We build a general basic question dataset for each metric following the setup for building the General Basic Question Dataset (GBQD).

\begin{SCtable}
\scalebox{1.0}{
\begin{tabular}{|c|c|c|c|c|c|c|}
\hline
Model & \begin{tabular}[c]{@{}l@{}}LQI \end{tabular} &  \begin{tabular}[c]{@{}l@{}}HAV \end{tabular} & 
\begin{tabular}[c]{@{}l@{}}HAR \end{tabular} & 
\begin{tabular}[c]{@{}l@{}}MU \end{tabular} &
\begin{tabular}[c]{@{}l@{}}MUA \end{tabular} &
\begin{tabular}[c]{@{}l@{}}MLB \end{tabular}\\
\hline
$R_{score1}$ & 0.19 & \textbf{0.48} & 0.45 & 0.30 & 0.34 & 0.36\\ 
\hline
$R_{score2}$ & 0.08 & 0.48 & \textbf{0.53} & 0.30 & 0.23 & 0.37\\ 
\hline
\end{tabular}}
\caption{This table shows the robustness scores, $R_{score}$, of six state-of-the-art VQA models based on GBQD ($R_{score1}$), YNBQD ($R_{score2}$) and VQA \cite{4} dataset. LQI denotes LSTM Q+I, HAV denotes HieCoAtt (Alt,VGG19), HAR denotes HieCoAtt (Alt,Resnet200), MU denotes MUTAN without Attention, MUA denotes MUTAN with Attention and MLB denotes MLB with Attention. The $R_{score}$ parameters are $(t,~m) = (0.05,~20)$.}
\label{table:table1}
\end{SCtable}

\vspace{3pt}\noindent\textbf{Results and Analysis.~}

Next, we will present our experimental results and robustness analysis. 

\begin{figure*}[!tbp]
  \begin{subfigure}[b]{0.49\textwidth}
    \includegraphics[width=0.98\linewidth]{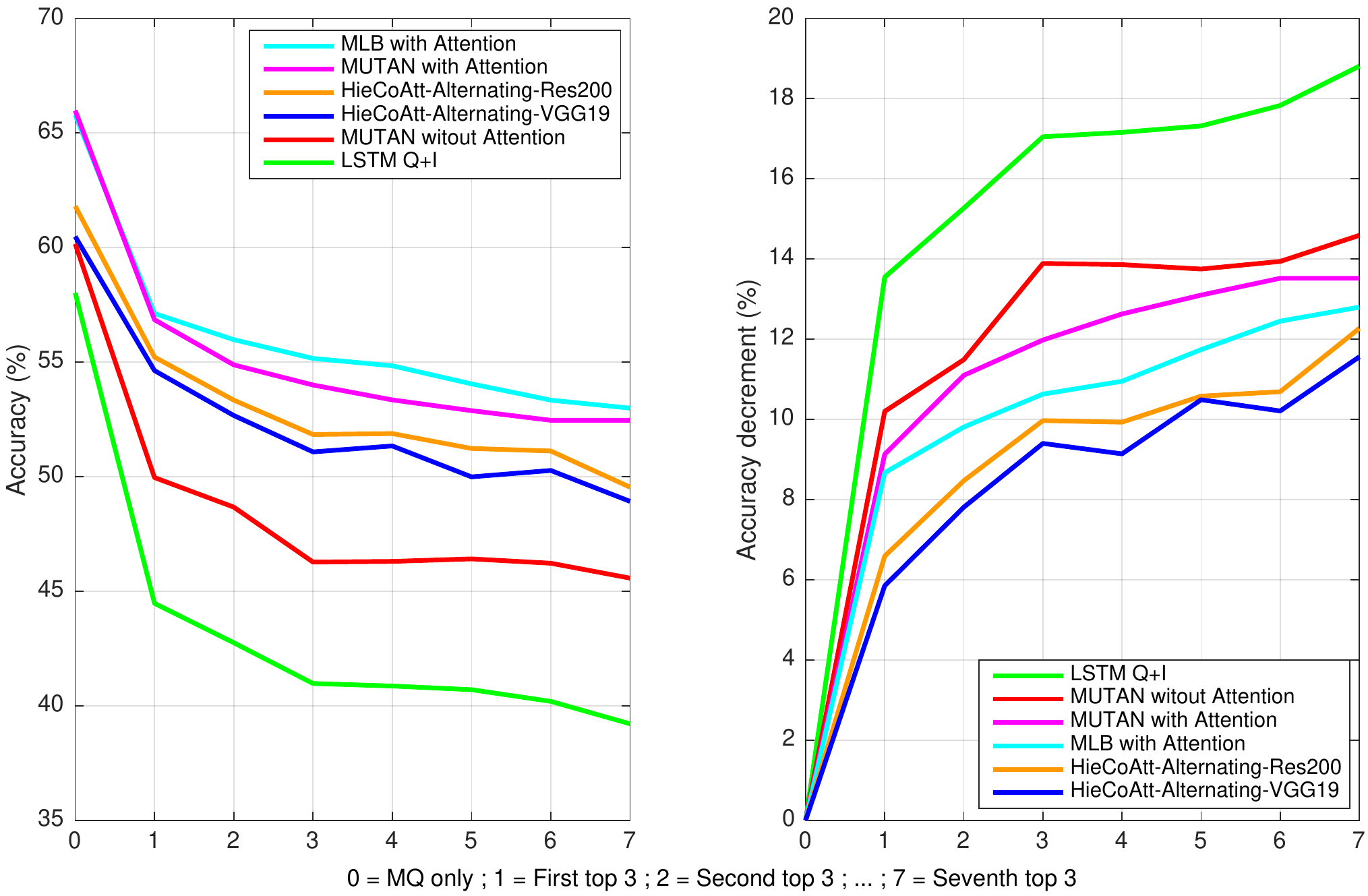}
    \caption{-1\space\space\space\space\space\space\space\space\space\space\space\space\space\space\space\space\space\space\space\space\space\space\space\space\space\space\space\space\space\space\space\space\space\space\space\space\space\space\space\space\space\space\space(a)~-2}
  \end{subfigure}
  \hfill
  \begin{subfigure}[b]{0.49\textwidth}
    \includegraphics[width=0.98\linewidth]{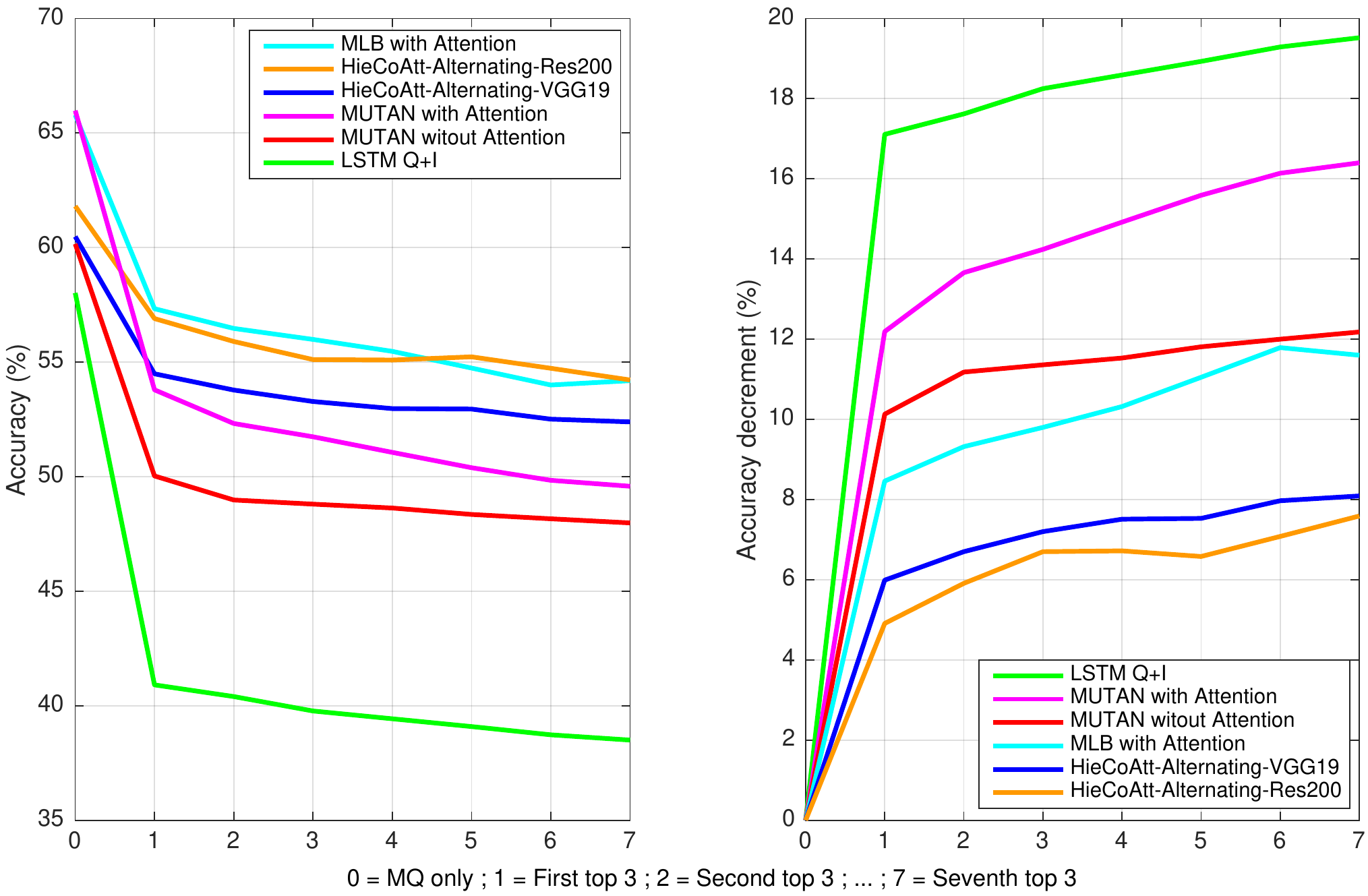}
    \caption{-1\space\space\space\space\space\space\space\space\space\space\space\space\space\space\space\space\space\space\space\space\space\space\space\space\space\space\space\space\space\space\space\space\space\space\space\space\space\space\space\space\space\space\space(b)~-2}
  \end{subfigure}
  \vspace{-0.2cm}
  \caption{The figure shows the ``accuracy'' and ``accuracy decrement'' of the six state-of-the-art pretrained VQA models evaluated on GBQD, YNBQD and VQA \cite{4} datasets. These results are based on our proposed $LASSO$ BQ ranking method. Note that we divide the top 21 ranked GBQs into 7 partitions where each partition contains 3 ranked GBQs; this is in reference to (a)-1 and (a)-2. We also divide the top 21 ranked YNBQs into 7 partitions and each partition contains 3 ranked YNBQs; this is in reference to (b)-1 and (b)-2. BQs are acting as noise, so the partitions represent the noises ranked from the least noisy to the noisiest. That is, in this figure the first partition is the least noisy partition and so on. Because the plots are monotonously decreasing in  accuracy, or, equivalently, monotonously increasing in accuracy decrement, the ranking is effective. In this figure, ``First top 3'' represents the first partition, ``Second top 3'' represents the second partition and so on.}
\label{fig:figure4}
\vspace{-0.2cm}
\end{figure*}

\begin{figure*}[!tbp]
  \begin{subfigure}[b]{0.49\textwidth}
    \includegraphics[width=0.98\linewidth]{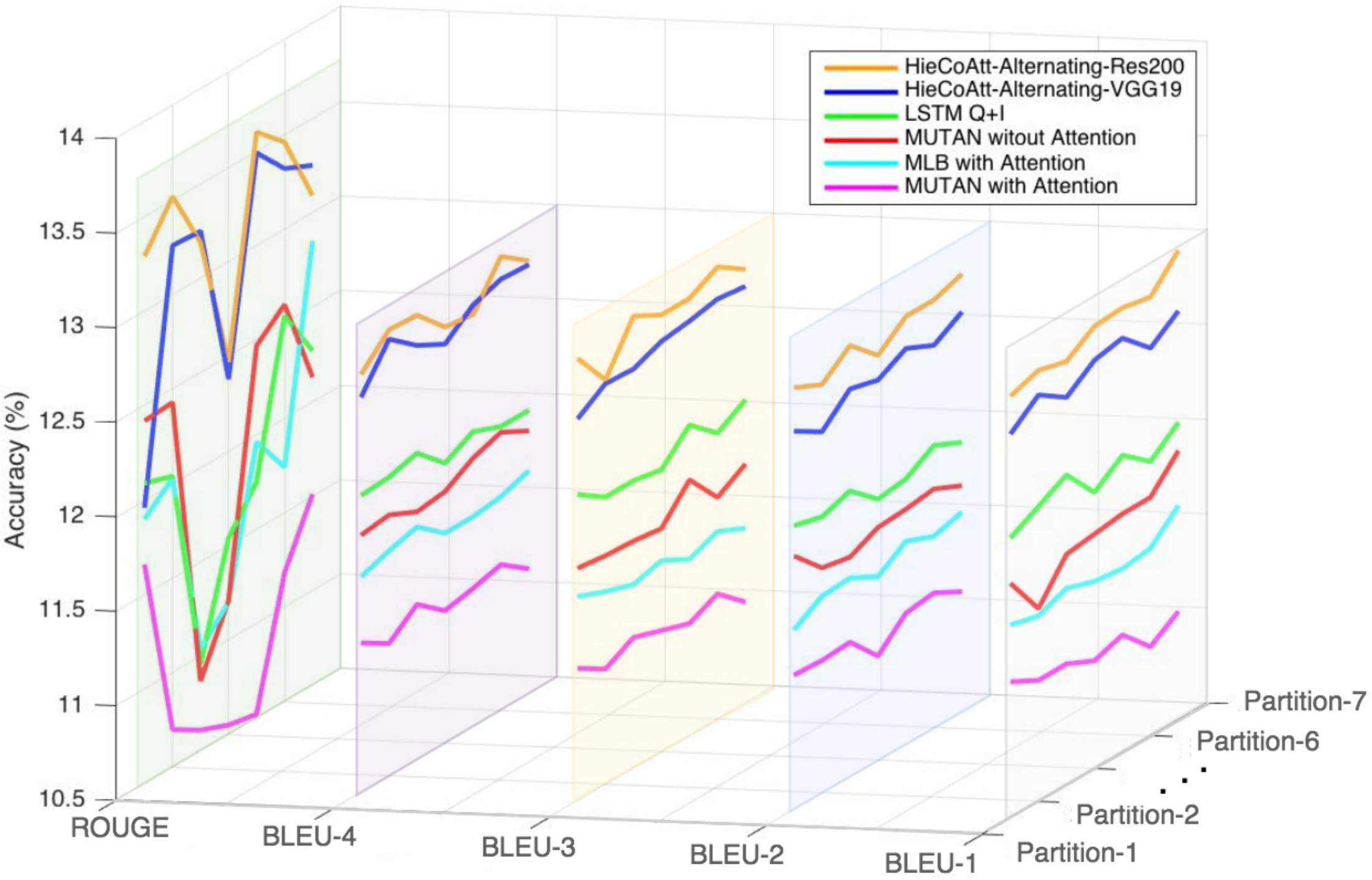}
    \caption{ROUGE, BLEU-4, BLEU-3, BLEU-2 and BLEU-1}
  \end{subfigure}
  \hfill
  \begin{subfigure}[b]{0.49\textwidth}
    \includegraphics[width=0.98\linewidth]{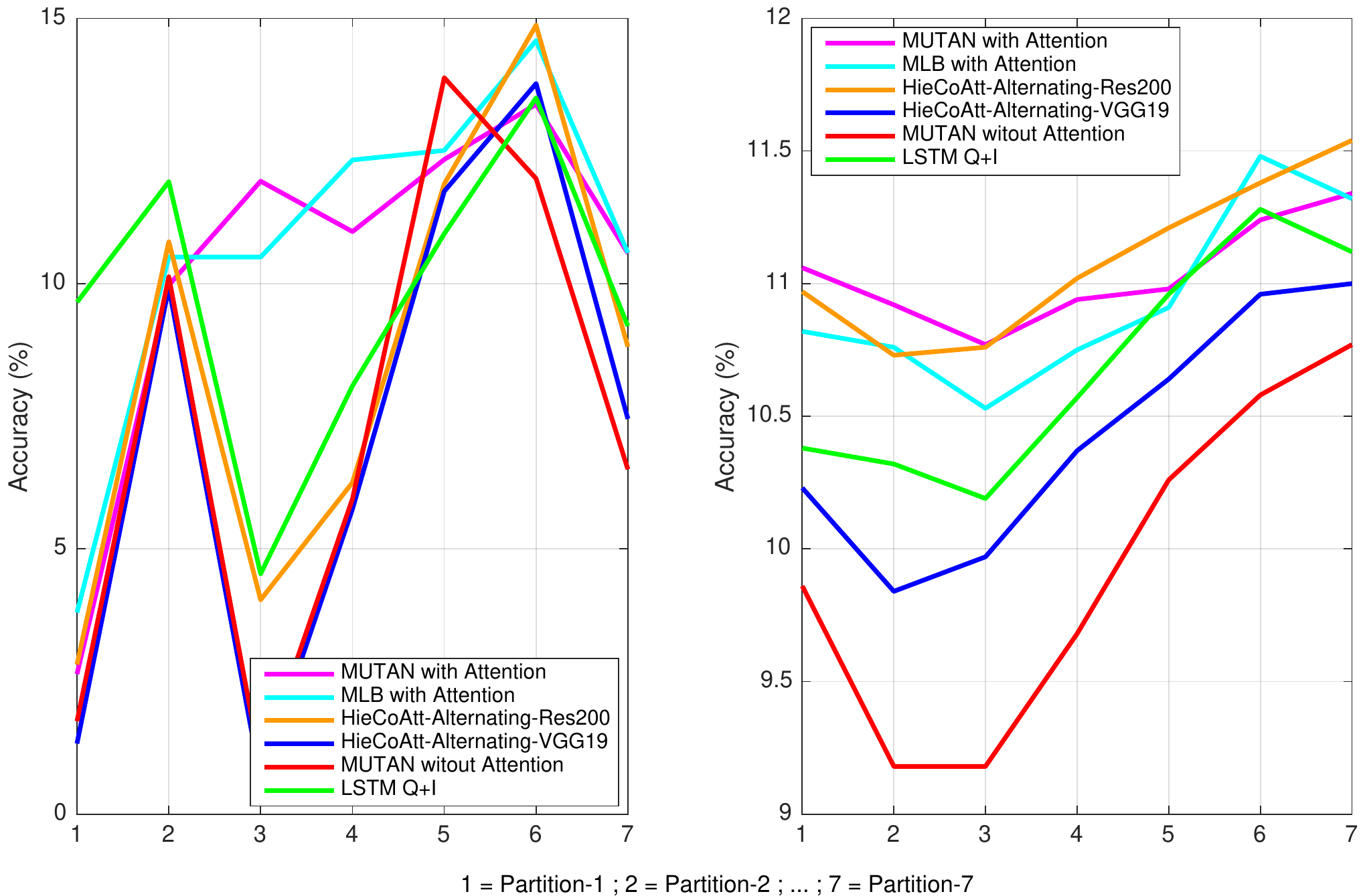}
    \caption{CIDEr\space\space\space\space\space\space\space\space\space\space\space\space\space\space\space\space\space\space\space\space\space\space\space\space\space\space\space\space\space\space\space\space\space\space(c) METEOR}
  \end{subfigure}
  \vspace{-0.2cm}
  \caption{This figure shows the accuracy of six state-of-the-art pretrained VQA models evaluated on the GBQD and VQA dataset by different BQ ranking methods, BLEU-1, BLEU-2, BLEU-3, BLEU-4, ROUGE, CIDEr and METEOR. In (a), the grey shade denotes BLEU-1, blue shade denotes BLEU-2, orange shade denotes BLEU-3, purple shade denotes BLEU-4 and green shade denotes ROUGE. In this figure, the definition of partitions are same as Figure \ref{fig:figure4}. The original accuracy of the six VQA models can be referred to Table \ref{table:table6}-(a), Table \ref{table:table6}-(b), etc. To make the figure clear, we plot the results of CIDEr and METEOR in (b) and (c), respectively. Based on this figure and Figure \ref{fig:figure4} in our paper, our $LASSO$ ranking method performance is better than those seven ranking methods.}
\label{fig:figure5}
\end{figure*}

\vspace{3pt}\noindent\textbf{(i)} \textbf{Are the rankings of BQs effective?} We divide the top $21$ ranked BQs into seven partitions, each containing three top-ranked BQs, and observe that the accuracy decreases from the first partition to the seventh partition (Figure \ref{fig:figure4}-(a)-1). Additionally, the accuracy decrement increased from the first partition to the seventh (Figure \ref{fig:figure4}-(a)-2), indicating that the similarity of BQs to the given MQ decreased from the first partition to the seventh ({\em i.e.}, the noise level increased). These trends are also observed when we use the YNBQD dataset (Figure \ref{fig:figure4}-(b)-1 and Figure \ref{fig:figure4}-(b)-2). These results suggest that the rankings by the proposed $LASSO$-based ranking method are effective. However, the accuracy of the seven similarity metrics ($\{(BLEU_1..._4,~ROUGE,~CIDEr,~METEOR)\}$) was much more random and less monotonous from the first partition to the seventh partition (Figure \ref{fig:figure5}). This indicates that the added BQs based on these metrics represent much more noise than the ones ranked by the $LASSO$-based ranking method, significantly harming the accuracy of state-of-the-art VQA models. Hence, we conclude that the rankings by these seven sentence similarity metrics are not effective in this context.

\vspace{3pt}\noindent\textbf{(ii)} \textbf{Which VQA model is the most robust?} We classify the utilized state-of-the-art VQA models into two distinct groups:: attention-based and non-attention-based, as shown in Table \ref{table:table1}. HAV, HAR, MUA, and MLB belong to the attention-based models, while LQI and MU are non-attention-based. Generally, based on Table \ref{table:table1}, attention-based VQA models are more robust than non-attention-based ones. However, when we examine MU and MUA in Table 4 ($R_{score2}$), the non-attention-based model (MU) is more robust than the attention-based model (MUA). It is worth noting that the only difference between MU and MUA is the attention mechanism. Meanwhile, in Table \ref{table:table1} ($R_{score1}$), MUA is more robust than MU, indicating that the diversity of BQ candidates affects the robustness of attention-based VQA models in some cases. Ultimately, based on the results in Table \ref{table:table1}, we conclude that HieCoAtt \cite{41} is the most robust VQA model. The HieCoAtt model employs a co-attention mechanism that repeatedly exploits the text and image information to guide each other, which enhances the robustness of VQA models \cite{41,huang2019novel}. Our experimental results show that HieCoAtt is indeed the most robust VQA model, which motivates us to conduct further experiments on this model.


\begin{SCfigure}
\scalebox{0.98}{
  \includegraphics[width=0.485\linewidth]{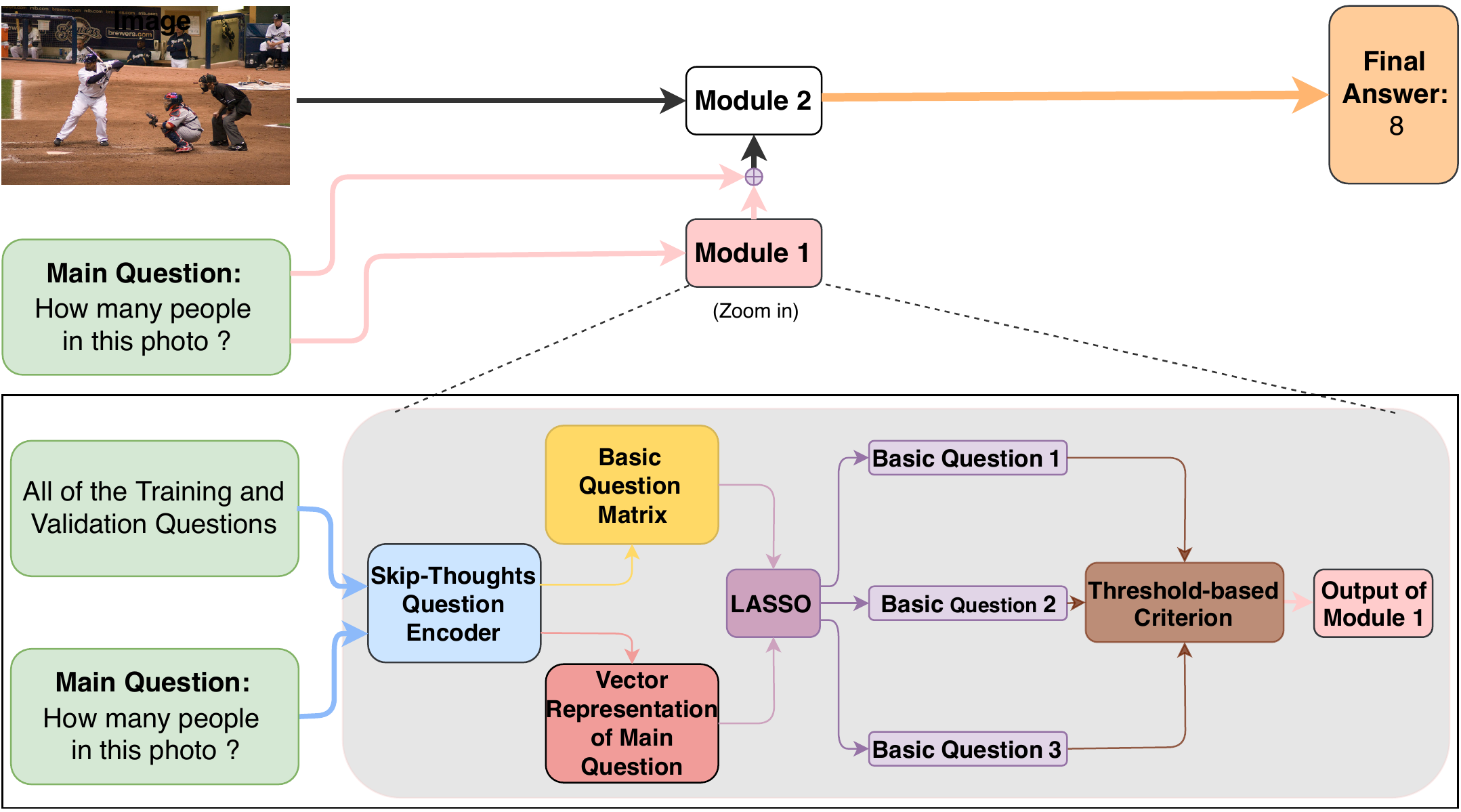}}
  \caption{Visual Question Answering by Basic Questions (VQABQ) pipeline. Note that in Module 1 all of the training and validation questions are only encoded by Skip-Thought Question Encoder once for generating the Basic Question Matrix. That is, the next input of Skip-Thought Question Encoder is only a new main question. Module 2 is a VQA model which we want to test, and it is the HieCoAtt VQA model in our case. Regarding the input question of the HieCoAtt model, it is the direct concatenation of a given main question with the corresponding selected basic questions based on the Threshold-based Criterion. ``$\oplus$'' denotes the direct concatenation of basic questions.}
\label{fig:figure8}
\end{SCfigure}



\begin{SCtable}
\scalebox{0.94}{
\begin{tabular}{|c|c|c|c|}
\hline
\multicolumn{1}{|c|}{} & score1 & score2/score1 & score3/score2 \\ \hline
avg & ~0.33  & ~~~~~~0.61 & ~~~~~~0.73 \\ 
\hline
std & ~0.20  & ~~~~~~0.27 & ~~~~~~0.21 \\
\hline
\end{tabular}}
\caption{In this table, ``avg'' denotes average and ``std'' denotes standard deviation.}
\label{table:table3}
\end{SCtable}


\begin{SCtable}
\scalebox{0.94}{
\begin{tabular}{|c|c|c|c|c|}
\hline
\multicolumn{5}{|c|}{Opend-Ended Case (Total: 244302 questions)}\\ 
\hline
& \begin{tabular}[c]{@{}l@{}}~~~0 BQ \\
(96.84\%)\end{tabular} & \begin{tabular}[c]{@{}l@{}}~~1 BQ \\ (3.07\%)\end{tabular} & \begin{tabular}[c]{@{}l@{}}~~2 BQ \\ (0.09\%)\end{tabular} & \begin{tabular}[c]{@{}l@{}}~~3 BQ \\ (0.00\%)\end{tabular} \\
\hline
\# Q & ~236570  & ~~~7512  & ~~~211  & ~~~~~9 \\ 
\hline
\end{tabular}}
\caption{The table shows how many BQs are appended. ``$X$ BQ'' means $X$ BQs are appended by MQ, where $X = 0, 1, 2, 3$, and ``\# Q'' denote number of questions.}
\label{table:table4}
\end{SCtable}

\vspace{3pt}\noindent\textbf{(iii)} \textbf{Could in-context learning through a chain of BQs improve the accuracy of the HieCoAtt model?} 

Table \ref{table:table1} shows that HieCoAtt is the most robust VQA model and was previously the state-of-the-art model in terms of accuracy \cite{41}. These factors motivate us to conduct an extended experiment and analysis of this model. We propose a framework called Visual Question Answering by Basic Questions (VQABQ) to analyze the HieCoAtt VQA model using selected high-quality BQs, as shown in Figure \ref{fig:figure8}. We use Algorithm \ref{algorithm1} to select BQs with good quality based on a threshold-based criterion. In our proposed BQD, each MQ has 21 corresponding BQs with scores and these scores are all between $[0-1]$ with the following order: 
\begin{equation}
    ~~~~~~~~~~~~~~~score1\geq score2\geq...\geq score21,
\end{equation}
where we define three thresholds ($s1$, $s2$, and $s3$) for the selection process and only consider the top three ranked BQs. We compute the averages ($avg$) and standard deviations ($std$) for $score1$, $score2/score1$, and $score3/score2$, (refer to Table \ref{table:table3}) and use $avg \pm std$ as the initial estimation of the thresholds. We find that when  $s1 = 0.60$, $s2 = 0.58$, and $s3 = 0.41$, we get the BQs that best help the accuracy of the HieCoAtt VQA model with the MQ-BQs direct concatenation method. 

However, Table \ref{table:table4} shows that only around $3.16\%$ of MQs benefit from BQs, with $96.84\%$of testing questions unable to find the proper BQs to improve the accuracy of the HieCoAtt model. Despite this, based on Table \ref{table:table100}, our method still improves the performance of the HieCoAtt model, increasing accuracy from $60.32\%$ to $60.34\%$, and answering approximately $49$ more questions correctly than the original HieCoAtt VQA model \cite{41}. Based on these results, we believe that BQs with good enough quality can help increase the accuracy of the HieCoAtt VQA model using the direct concatenation method.

\noindent\textbf{The reason why in-context learning with a chain of BQs helps the HieCoAtt model.}
By incorporating the chain of BQs into the process of VQA, the performance of the HieCoAtt model can be enhanced. The reason is that BQs are designed to capture the basic semantic concepts of images and questions. By using the BQs in a chain, the model can learn in context and leverage the knowledge gained from the BQs to better understand the input. Furthermore, the use of BQs can overcome the limitations of the HieCoAtt model, which relies heavily on co-attention between the image and question modalities. Co-attention models can struggle with complex questions that require multiple steps to answer, but by incorporating BQs, the model can break down complex questions into simpler sub-questions that it can answer more easily. The chain of BQs also serves as a form of scaffolding, guiding the model towards the correct answer by providing intermediate steps that can help the model reason about the question more effectively. Overall, in-context learning with a chain of BQs can help the HieCoAtt model make more accurate predictions by providing additional information and guidance, especially for complex questions that may be difficult for the model to answer using co-attention alone.


\begin{SCtable}
\scalebox{0.94}{
\begin{tabular}{|c|c|c}
\multicolumn{3}{c}{~~~~~~~~~~~~~~~~~~~~~~~HieCoAtt (Alt,VGG19)} \\ 
\hline
\hline
\multicolumn{2}{c|}{~~~~~(s1, s2, s3)} & (test-dev-acc, Other, Num, Y/N) \\ \hline
\multicolumn{2}{c|}{(0.60, 0.58, 0.41)} & ~~~(60.49, 49.12, 38.43, 79.65) \\ \hline
\hline
\multicolumn{2}{c|}{~~~~~(s1, s2, s3)} & (test-std-acc, Other, Num, Y/N) \\ \hline
\multicolumn{2}{c|}{(0.60, 0.58, 0.41)} & ~~~(60.34, 49.16, 36.50, 79.49) \\ \hline
\hline
\end{tabular}}
\caption{Evaluation results of HieCoAtt (Alt,VGG19) model improved by Algorithm \ref{algorithm1}. Note that the original accuracy of HieCoAtt (Alt,VGG19) VQA model for ``test-dev-acc'' is $60.48$ and for ``test-std-acc'' is $60.32$.}
\label{table:table100}
\end{SCtable}


\begin{algorithm}[t]
\caption{~~MQ-BQs Concatenation Algorithm}\label{algorithm1}
\begin{algorithmic}[1]
\State \textbf{Note that s1, s2, s3 are thresholds we can choose.}
\Procedure{MQ-BQs concatenation}{}

\If {$score1 > s1$} 
\State appending the given MQ and BQ1 with the largest score
\EndIf

\If {$score2/score1 > s2$} 
\State appending the given MQ, BQ1, and BQ2 with the second large score
\EndIf

\If {$score3/score2 > s3$} 
\State appending the given MQ, BQ1, BQ2, and BQ3 with the third large score
\EndIf

\EndProcedure
\end{algorithmic}
\end{algorithm}

\begin{SCfigure}
\scalebox{0.94}{
  \includegraphics[width=0.42 \linewidth]{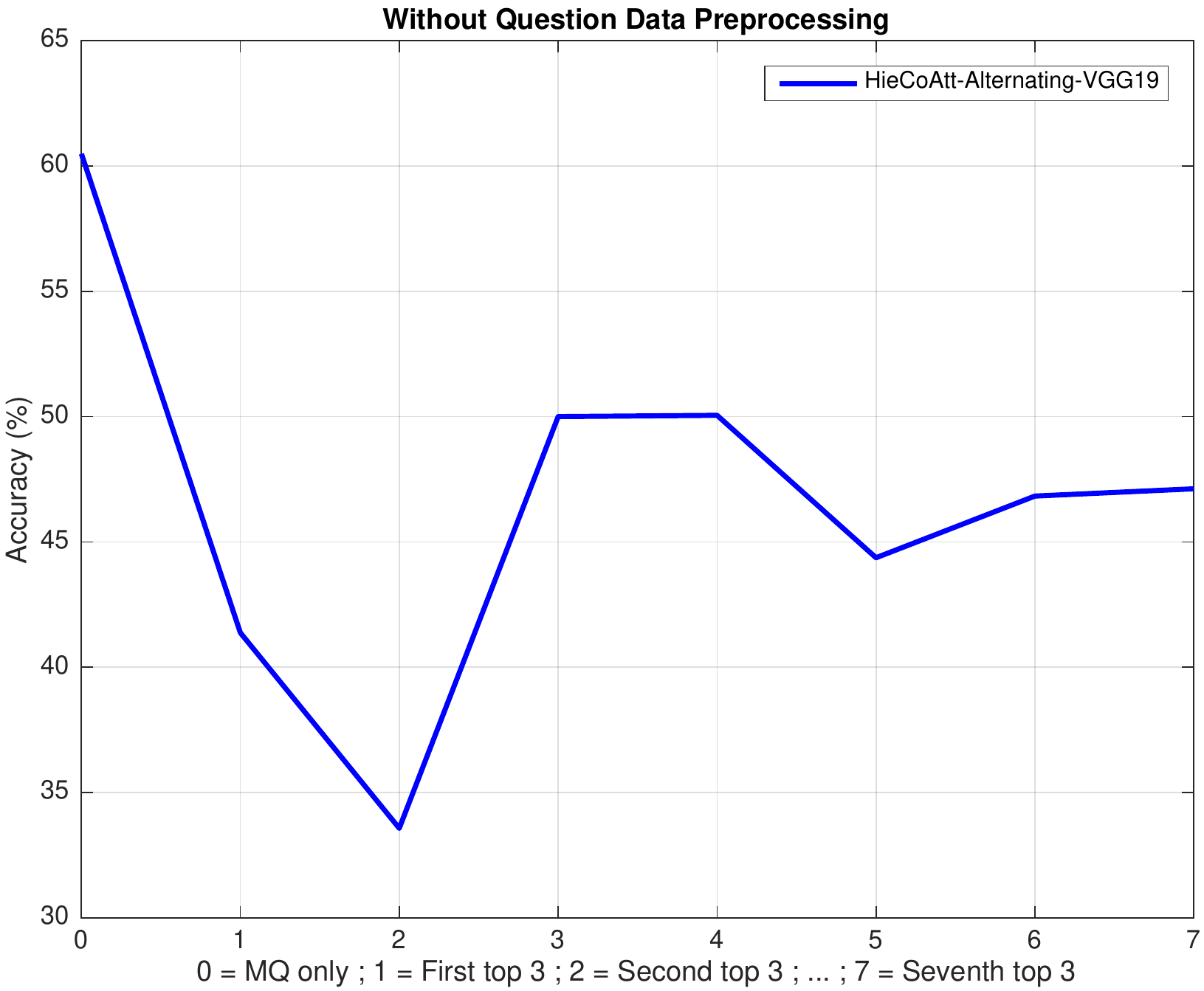}}
  \caption{This figure demonstrates what is the ranking result of jumping randomly. For convenience, we only take the most robust VQA model, HieCoAtt, to demonstrate the random jump. If we do not have question sentences preprocessing, then the proposed $LASSO$ ranking method is ineffective. That is, if we have done the question sentences preprocessing, the trend in this figure should be similar to Figure \ref{fig:figure4}-(a)-1 and Figure \ref{fig:figure4}-(b)-1. In this figure, ``MQ only'' represents the original query question and ``First top 3'' represents the first partition, ``Second top 3'' represents the second partition and so on. For the detailed numbers, please refer to Table \ref{table:table22}.}
\label{fig:figure9}
\end{SCfigure}

\noindent
\vspace{3pt}\noindent\textbf{(iv)} \textbf{Is question sentences preprocessing necessary?} We propose that preprocessing of question sentences is essential for our proposed $LASSO$-based ranking method. For convenience, we exploit the same HieCoAtt model to demonstrate the claim. In the \textit{Methodology} section, we do the preprocessing question sentences before the sentence embedding. Without question sentences preprocessing, the $LASSO$-based ranking method generates random ranking results. As illustrated in Figure \ref{fig:figure9}, the ranking result jumps randomly due to the lack of question sentences preprocessing. If the proposed method were functioning correctly, the trend of the ranking result should be monotonic, as seen in Figure \ref{fig:figure4}. Therefore, question sentences preprocessing is a necessary step for our proposed $LASSO$-based ranking method to work effectively.


\vspace{3pt}\noindent\textbf{(v)} \textbf{What are the pros and cons of each metric?} In order to compare our proposed $LASSO$-based BQ ranking method with other methods, we conduct BQ ranking experiments using seven text similarity metrics on the same BQ candidate dataset. While the performance of these metrics is not satisfactory, they are still used in various works \cite{1,19,27,28,48} due to their simple implementation. On the other hand, despite its simplicity, our $LASSO$-based ranking method shows quite effective performance. It should be noted that in practice, we will use our proposed datasets directly to test the robustness of VQA models without re-running the $LASSO$-based ranking method, so the computational complexity of the $LASSO$-based ranking method is not an issue in this case.

\begin{SCtable}
    \small
    \scalebox{0.94}{
    \begin{tabular}{ c | c c c c | c} 
     Task Type &    & \multicolumn{3}{c}{Open-Ended} &  \\ [0.5ex]
     \hline
     Method &    & \multicolumn{3}{c}{HieCoAtt (Alt,VGG19)} &  \\ [0.5ex]
     \hline
     Test Set&  \multicolumn{4}{c}{dev} & diff \\ [0.5ex]
     \hline
     Partition & Other & Num & Y/N & All & All \\ [0.5ex] 
     \hline
     First-dev & 33.83 & 37.19 & 51.34 & \textbf{41.38} & \textbf{20.43}  \\ 
     
     Second-dev & 15.46 & 31.42 & 55.38 & \textbf{33.58} & \textbf{28.23} \\
     
     Third-dev & 35.33 & 36.53 & 70.76 & \textbf{50.00} & \textbf{11.81}  \\
     
     Fourth-dev & 36.05 & 36.46 & 70.05 & \textbf{50.05} & \textbf{11.76} \\
     
     Fifth-dev & 29.89 & 30.02 & 65.14 & \textbf{44.37} & \textbf{17.44} \\
     
     Sixth-dev & 35.81 & 34.48 & 63.02 & \textbf{46.83} & \textbf{14.98}  \\
     
     Seventh-dev & 39.12 & 34.45 & 59.84 & \textbf{47.12} & \textbf{14.69}  \\
     \hline
     
     Original-dev  & 51.77 & 38.65 & 79.70 & \textbf{61.81} & -\\
     Original-std  & 51.95 & 38.22 & 79.95 & \textbf{62.06} & - \\ 
     
     \hline
    \end{tabular}}
    \caption{The HieCoAtt (Alt,VGG19) model evaluation results on BQD and VQA dataset \cite{4} without question sentences preprocessing. ``-'' indicates the results are not available, ``-std'' means that the VQA model is evaluated by the complete testing set of BQD and VQA dataset, and ``-dev'' means that the VQA model is evaluated by the partial testing set of BQD and VQA dataset. In addition, $diff = Original_{dev_{All}} - X_{dev_{All}}$, where $X$ is equal to ``First'', ``Second'',..., ``Seventh''.}
\label{table:table22}
\vspace{-0.5cm}
\end{SCtable}

\begin{figure*}[!tbp]
  \begin{subfigure}[b]{0.49\textwidth}
    \includegraphics[width=0.98\linewidth]{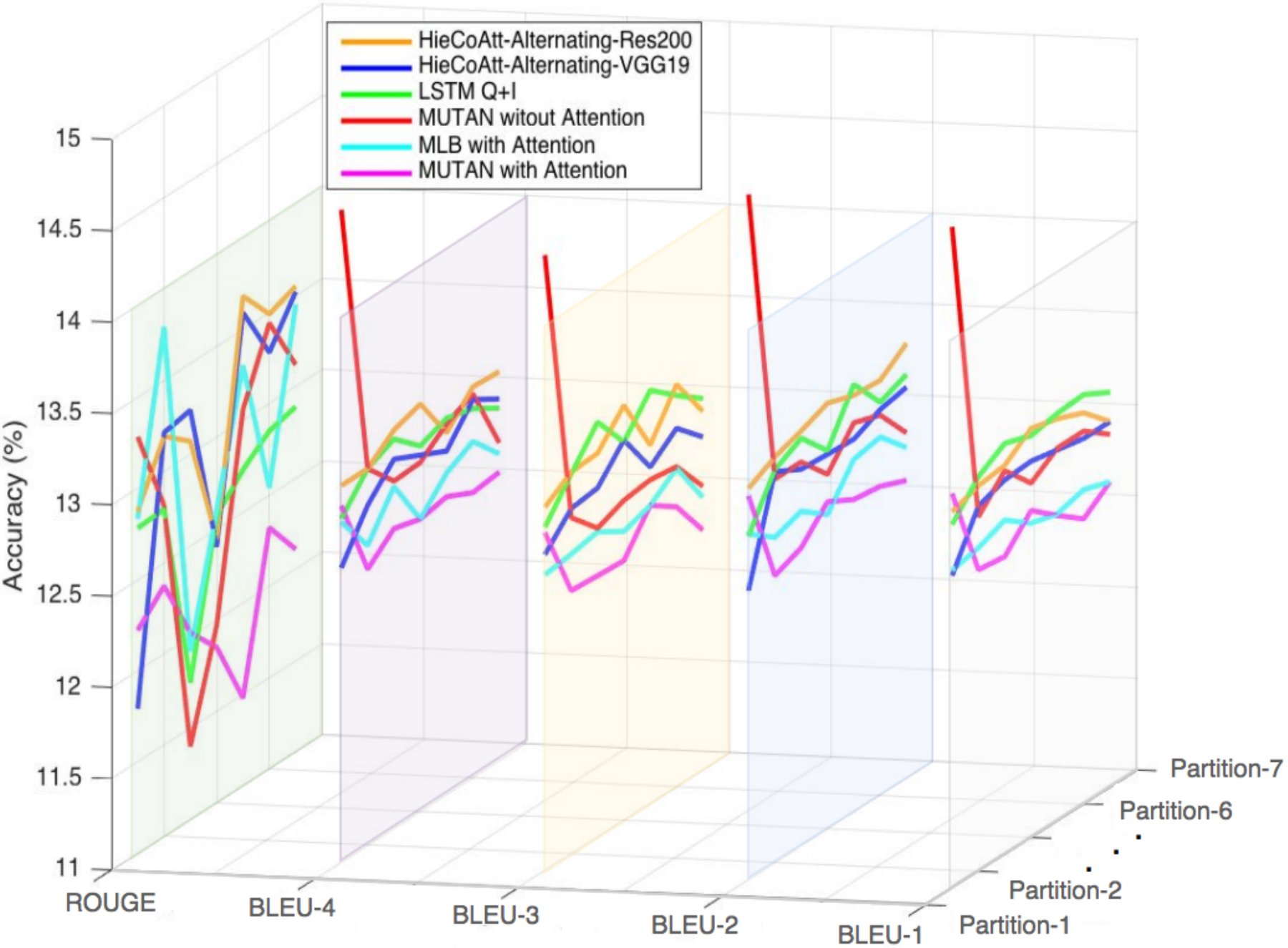}
    \caption{ROUGE, BLEU-4, BLEU-3, BLEU-2 and BLEU-1}
  \end{subfigure}
  \hfill
  \begin{subfigure}[b]{0.49\textwidth}
    \includegraphics[width=0.98\linewidth]{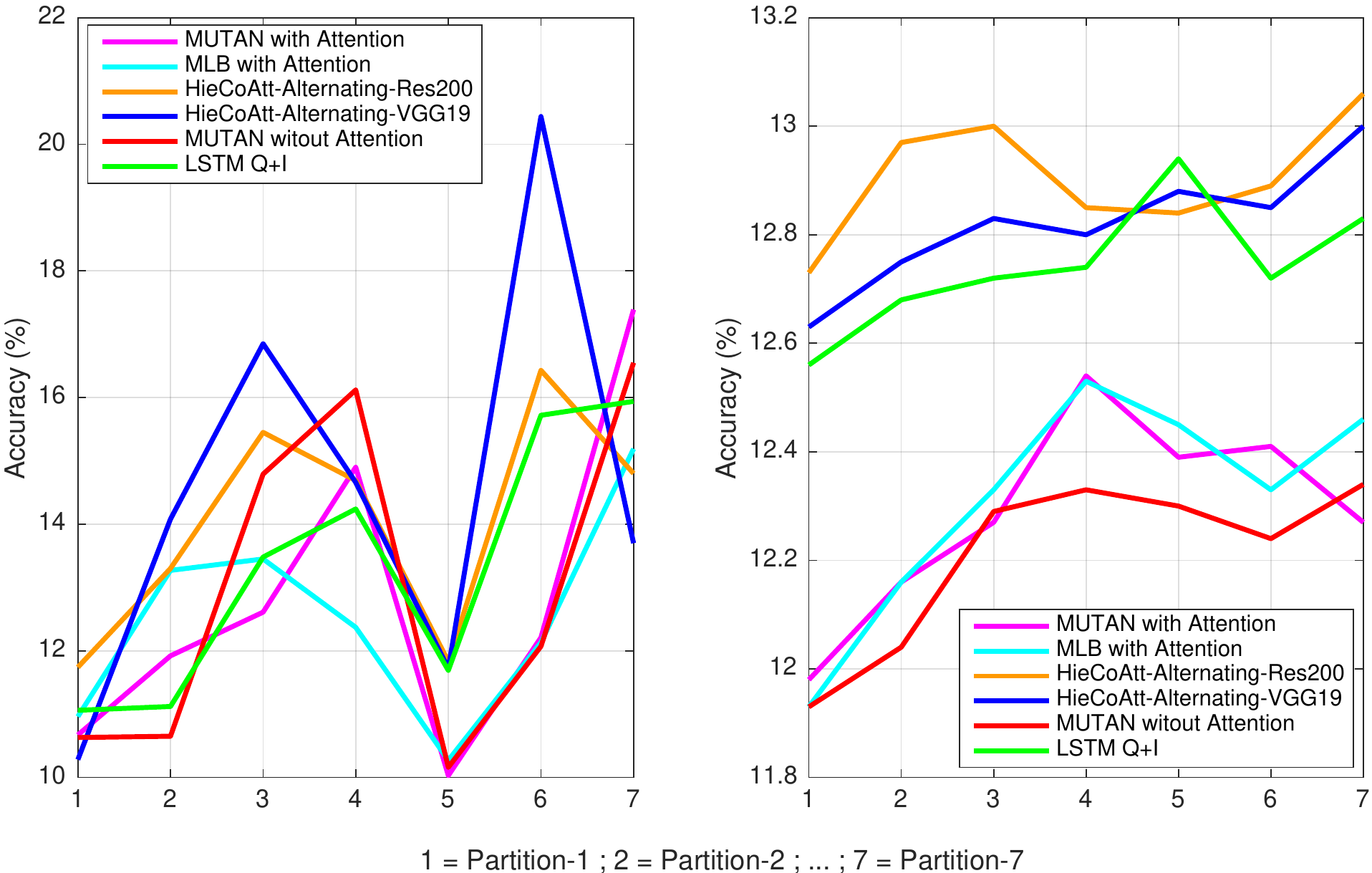}
    \caption{CIDEr~~~~~~~~~~~~~~~~~~~~~~~~~~~~~~~(c) METEOR}
  \end{subfigure}
  \vspace{-0.2cm}
  \caption{This figure shows the accuracy of six state-of-the-art pretrained VQA models evaluated on the YNBQD and VQA \cite{4} dataset by different BQ ranking methods, BLEU-1, BLEU-2, BLEU-3, BLEU-4, ROUGE, CIDEr and METEOR. The result in this figure is consistent with the result in Figure \ref{fig:figure5}. Note that in Figure \ref{fig:figure7}-(a), the grey shade denotes BLEU-1, blue shade denotes BLEU-2, orange shade denotes BLEU-3, purple shade denotes BLEU-4 and green shade denotes ROUGE. For more detailed explanation, please refer to the \textit{Extended experiments on YNBQD dataset} subsection.}
\label{fig:figure7}
\vspace{-0.2cm}
\end{figure*}


\vspace{3pt}\noindent\textbf{(vi)} \textbf{What affects the quality of BQs?} In our model, the parameter $\lambda$ plays a crucial role in determining the quality of BQs. After conducting experiments, we have discovered that the value of $\lambda \in [10^{-6}, 10^{-5}]$ produces satisfactory ranking performance, as demonstrated in Figure \ref{fig:figure4}. We have provided some ranking examples using the $LASSO$-based ranking method in Figure \ref{fig:figure10} and Table \ref{table:table5} to showcase the quality of BQs when $\lambda$ is set to $10^{-6}$.



\vspace{3pt}\noindent\textbf{(vii)} \textbf{Extended experiments on YNBQD dataset.} Although we have conducted BQ ranking experiments using seven different text similarity metrics, $BLEU_1..._4,~ROUGE,~CIDEr,$ and $METEOR$, on the GBQD dataset, we haven't done the same experiments on the YNBQD dataset. Therefore, we provide the experimental details in this subsection. We extend our experiments to include the YNBQD dataset and used the aforementioned seven metrics to rank the BQs. The definition of partitions in Figure \ref{fig:figure7} is the same as that in Figure \ref{fig:figure4}. The original accuracy of the six VQA models is given in Table \ref{table:table7}-(a) to \ref{table:table7}-(f). For convenience, we plot the results of CIDEr and METEOR in Figure \ref{fig:figure7}-(b) and Figure \ref{fig:figure7}-(c), respectively. Based on Figures \ref{fig:figure7}, \ref{fig:figure5}, and \ref{fig:figure4}, we conclude that the proposed $LASSO$-based ranking method outperforms the seven ranking methods on both the YNBQD and GBQD datasets. For detailed experiment results, please refer to Table \ref{table:table6}, \ref{table:table7},..., \ref{table:table21}.

\vspace{-0.5cm}
\section{Discussion}
In this section, we discuss our findings on the state-of-the-art VQA models among the six models that we tested, namely  \cite{4,41,57,59}, in various aspects.



\noindent
\vspace{+0.1cm}\textbf{In the sense of robustness.} 

Based on Table \ref{table:table1}, we can see that the ``HieCoAtt (Alt,VGG19)'' model achieves the highest $R_{score1}$ of $0.48$, while the ``HieCoAtt (Alt,Resnet200)'' model has the highest $R_{score2}$ of $0.53$. Therefore, among our six tested VQA models, the ``HieCoAtt (Alt,VGG19)'' model is the state-of-the-art for GBQD in terms of robustness, while the ``HieCoAtt (Alt,Resnet200)'' model is the state-of-the-art for YNBQD. On the other hand, the ``LSTM Q+I'' model performs the worst with the lowest $R_{score1}$ and $R_{score2}$. Generally, we can conclude that attention-based VQA models are more robust than non-attention-based ones.


\noindent
\vspace{+0.1cm}\textbf{In the sense of accuracy.} 

Based on the results in Table \ref{table:table6}, we can see that the ``MUTAN with Attention'' model has the highest accuracy of $65.77\%$, while `LSTM Q+I'' has the lowest accuracy of $58.18\%$. Thus, we can conclude that the ``MUTAN with Attention'' model is the state-of-the-art VQA model among the six models tested, in terms of accuracy. These findings also suggest that the attention-based VQA model performs better in terms of accuracy compared to the non-attention-based model.


\vspace{-0.5cm}
\section{Conclusion}
In this work, we introduce a novel approach consisting of several components, including the General Basic Question Dataset, Yes/No Basic Question Dataset, and a robustness measure ($R_{score}$) for assessing the robustness of VQA models. Our method is composed of two main modules, the Noise Generator and the VQA module. The former ranks the given BQs, while the latter takes the query, basic questions, and an image as input and generates a natural language answer to the query question about the image. The aim of our proposed method is to serve as a benchmark for aiding the community in developing more accurate \emph{and} robust VQA models. Furthermore, using our proposed General and Yes/No Basic Question Datasets and $R_{score}$, we demonstrate that our $LASSO$-based BQ ranking method performs better than most popular text evaluation metrics. Finally, we have presented some new methods for evaluating the robustness of VQA models, which could inspire interesting future work on building robust \emph{and} accurate VQA models.

\begin{acks}
This work is supported by competitive research funding from the University of Amsterdam and King Abdullah University of Science and Technology (KAUST).
\end{acks}

\bibliographystyle{ACM-Reference-Format}
\bibliography{sample-base}

\appendix

\section*{Appendices}
Detailed experimental results are presented in Table \ref{table:table8}, \ref{table:table9},..., \ref{table:table21}. 


\begin{table*}[ht!]
\renewcommand\arraystretch{1.0}
\setlength\tabcolsep{11pt}
    \centering
\begin{subtable}[t]{0.3\linewidth}
\centering
\scalebox{0.5}{
    \begin{tabular}{ c | c c c c | c} 
     Task Type &    & \multicolumn{3}{c}{Open-Ended (BLEU-1)} &  \\ [0.5ex]
     \hline
     Method &    & \multicolumn{3}{c}{MUTAN without Attention} &  \\ [0.5ex]
     \hline
     Test Set&  \multicolumn{4}{c}{dev} & diff \\ [0.5ex]
     \hline
     Partition & Other & Num & Y/N & All & All \\ [0.5ex] 
     \hline
     First-dev & 2.73 & 2.92 & 24.63 & 11.74 & 48.42  \\ 
     
     Second-dev & 2.57 & 2.96 & 24.28 & 11.52 & 48.64 \\
     
     Third-dev & 2.79 & 2.89 & 24.53 & 11.72 & 48.44  \\
     
     Fourth-dev & 2.68 & 2.94 & 24.67 & 11.74 & 48.42  \\
     
     Fifth-dev & 2.69 & 2.87 & 24.73 & 11.76 & 48.40 \\
     
     Sixth-dev & 2.80 & 2.79 & 24.62 & 11.76 & 48.40  \\
     
     Seventh-dev & 2.77 & 2.99 & 25.01 & 11.92 & 48.24  \\
     \hline
     First-std & 2.52 & 2.70 & 24.66 & 11.66 & 48.79 \\
     \hline
     
     Original-dev  & 47.16 & 37.32 & 81.45 & 60.16 & -\\
     Original-std  & 47.57 & 36.75 & 81.56 & 60.45 & - \\

     \hline
    \end{tabular}}
    \centering
    \captionsetup{justification=centering}
    \caption{MUTAN without Attention model evaluation results.}

\scalebox{0.5}{
    \begin{tabular}{ c | c c c c | c} 
     Task Type &    & \multicolumn{3}{c}{Open-Ended (BLEU-1)} &  \\ [0.5ex]
     \hline
     Method &    & \multicolumn{3}{c}{HieCoAtt (Alt,VGG19)} &  \\ [0.5ex]
     \hline
     Test Set&  \multicolumn{4}{c}{dev} & diff \\ [0.5ex]
     \hline
     Partition & Other & Num & Y/N & All & All \\ [0.5ex] 
     \hline
     First-dev & 2.94 & 2.93 & 26.31 & 12.53 & 47.95  \\ 
     
     Second-dev & 2.95 & 3.05 & 26.56 & 12.65 & 47.83 \\
     
     Third-dev & 3.08 & 2.95 & 26.19 & 12.55 & 47.93  \\
     
     Fourth-dev & 3.17 & 2.90 & 26.36 & 12.66 & 47.82  \\
     
     Fifth-dev & 3.17 & 3.05 & 26.39 & 12.69 & 47.79 \\
     
     Sixth-dev & 3.21 & 3.11 & 25.99 & 12.55 & 47.93  \\
     
     Seventh-dev & 3.12 & 3.14 & 26.37 & 12.66 & 47.82  \\
     \hline
     First-std & 2.76 & 2.73 & 26.10 & 12.38 & 47.94 \\
     \hline

     Original-dev  & 49.14 & 38.35 & 79.63 & 60.48 & -\\
     Original-std  & 49.15 & 36.52 & 79.45 & 60.32 & - \\

     \hline
    \end{tabular}}
    \centering
    \captionsetup{justification=centering}
    \caption{HieCoAtt (Alt,VGG19) model evaluation results.}
\end{subtable}%
    \hfil
\begin{subtable}[t]{0.3\linewidth}

\centering
\scalebox{0.5}{
    \begin{tabular}{ c | c c c c | c} 
     Task Type &    & \multicolumn{3}{c}{Open-Ended (BLEU-1)} &  \\ [0.5ex]
     \hline
     Method &    & \multicolumn{3}{c}{MLB with Attention} &  \\ [0.5ex]
     \hline
     Test Set&  \multicolumn{4}{c}{dev} & diff \\ [0.5ex]
     \hline
     Partition & Other & Num & Y/N & All & All \\ [0.5ex] 
     \hline
     First-dev & 2.65 & 2.17 & 24.38 & 11.52 & 54.27  \\ 
     
     Second-dev & 2.75 & 2.21 & 24.17 & 11.48 & 54.31 \\
     
     Third-dev & 2.89 & 2.03 & 24.19 & 11.54 & 54.25  \\
     
     Fourth-dev & 2.90 & 2.26 & 24.01 & 11.49 & 54.30  \\
     
     Fifth-dev & 2.81 & 2.17 & 24.09 & 11.47 & 54.32 \\
     
     Sixth-dev & 2.80 & 2.16 & 24.15 & 11.49 & 54.30  \\
     
     Seventh-dev & 2.91 & 2.29 & 24.32 & 11.63 & 54.16  \\
     \hline
     First-std & 2.68 & 2.03 & 23.88 & 11.35 & 54.33 \\
     \hline
     
     Original-dev  & 57.01 & 37.51 & 83.54 & 65.79 & -\\
     Original-std  & 56.60 & 36.63 & 83.68 & 65.68 & - \\

     \hline
    \end{tabular}}
    \centering
    \captionsetup{justification=centering}
    \caption{MLB with Attention model evaluation results.}
    
\scalebox{0.5}{
    \begin{tabular}{ c | c c c c | c} 
     Task Type &    & \multicolumn{3}{c}{Open-Ended (BLEU-1)} &  \\ [0.5ex]
     \hline
     Method &    & \multicolumn{3}{c}{MUTAN with Attention} &  \\ [0.5ex]
     \hline
     Test Set&  \multicolumn{4}{c}{dev} & diff \\ [0.5ex]
     \hline
     Partition & Other & Num & Y/N & All & All \\ [0.5ex] 
     \hline
     First-dev & 2.06 & 2.05 & 24.37 & 11.22 & 54.76  \\ 
     
     Second-dev & 2.13 & 2.28 & 24.05 & 11.14 & 54.84 \\
     
     Third-dev & 2.13 & 2.07 & 24.10 & 11.14 & 54.84  \\
     
     Fourth-dev & 2.11 & 2.26 & 23.09 & 11.07 & 54.91  \\
     
     Fifth-dev & 2.15 & 2.35 & 23.97 & 11.12 & 54.86 \\
     
     Sixth-dev & 2.06 & 2.24 & 23.73 & 10.97 & 55.01  \\
     
     Seventh-dev & 2.06 & 2.17 & 23.99 & 11.07 & 54.91  \\
     \hline
     First-std & 2.04 & 2.17 & 24.00 & 11.10 & 54.67 \\
     \hline
     
     Original-dev  & 56.73 & 38.35 & 84.11 & 65.98 & -\\
     Original-std  & 56.29 & 37.47 & 84.04 & 65.77 & - \\

     \hline
    \end{tabular}}
    \centering
    \captionsetup{justification=centering}
    \caption{MUTAN with Attention model evaluation results.}
\end{subtable}%
    \hfil
\begin{subtable}[t]{0.3\linewidth}
        
\centering
\scalebox{0.5}{
    \begin{tabular}{ c | c c c c | c} 
     Task Type &    & \multicolumn{3}{c}{Open-Ended (BLEU-1)} &  \\ [0.5ex]
     \hline
     Method &    & \multicolumn{3}{c}{HieCoAtt (Alt,Resnet200)} &  \\ [0.5ex]
     \hline
     Test Set&  \multicolumn{4}{c}{dev} & diff \\ [0.5ex]
     \hline
     Partition & Other & Num & Y/N & All & All \\ [0.5ex] 
     \hline
     First-dev & 3.37 & 3.09 & 26.26 & 12.73 & 49.08  \\ 
     
     Second-dev & 3.39 & 3.12 & 26.33 & 12.78 & 49.03 \\
     
     Third-dev & 3.51 & 2.98 & 26.15 & 12.74 & 49.07  \\
     
     Fourth-dev & 3.48 & 3.09 & 26.40 & 12.84 & 48.97  \\
     
     Fifth-dev & 3.56 & 2.85 & 26.37 & 12.85 & 48.96 \\
     
     Sixth-dev & 3.52 & 3.02 & 26.30 & 12.82 & 48.99  \\
     
     Seventh-dev & 3.60 & 3.22 & 26.57 & 12.98 & 48.83  \\
     \hline
     First-std & 3.22 & 2.77 & 25.95 & 12.54 & 49.52\\
     \hline

     Original-dev  & 51.77 & 38.65 & 79.70 & 61.81 & -\\
     Original-std  & 51.95 & 38.22 & 79.95 & 62.06 & - \\

     \hline
    \end{tabular}}
    \centering
    \captionsetup{justification=centering}
    \caption{HieCoAtt (Alt,Resnet200) model evaluation results.}

\scalebox{0.5}{
    \begin{tabular}{ c | c c c c | c} 
     Task Type &    & \multicolumn{3}{c}{Open-Ended (BLEU-1)} &  \\ [0.5ex]
     \hline
     Method &    & \multicolumn{3}{c}{LSTM Q+I} &  \\ [0.5ex]
     \hline
     Test Set&  \multicolumn{4}{c}{dev} & diff \\ [0.5ex]
     \hline
     Partition & Other & Num & Y/N & All & All \\ [0.5ex] 
     \hline
     First-dev & 2.08 & 3.09 & 25.95 & 11.98 & 46.04  \\ 
     
     Second-dev & 1.98 & 3.35 & 26.18 & 12.06 & 45.96 \\
     
     Third-dev & 2.04 & 3.25 & 26.32 & 12.14 & 45.88  \\
     
     Fourth-dev & 2.01 & 3.26 & 25.94 & 11.96 & 46.06  \\
     
     Fifth-dev & 2.03 & 3.31 & 26.15 & 12.07 & 45.95 \\
     
     Sixth-dev & 2.16 & 3.41 & 25.68 & 11.95 & 46.07  \\
     
     Seventh-dev & 2.10 & 3.31 & 26.08 & 12.07 & 45.95  \\
     \hline
     First-std & 2.03 & 3.31 & 25.86 & 11.98 & 46.20 \\
     \hline
     
     Original-dev  & 43.40 & 36.46 & 80.87 & 58.02 & -\\
     Original-std  & 43.90 & 36.67 & 80.38 & 58.18 & - \\

     \hline
    \end{tabular}}
    \centering
    \captionsetup{justification=centering}
    \caption{LSTM Q+I model evaluation results.}
\end{subtable}
\caption{The table shows the six state-of-the-art pretrained VQA models evaluation results on the GBQD and VQA dataset. ``-'' indicates the results are not available, ``-std'' represents the accuracy of VQA model evaluated on the complete testing set of GBQD and VQA dataset and ``-dev'' indicates the accuracy of VQA model evaluated on the partial testing set of GBQD and VQA dataset. In addition, $diff = Original_{dev_{All}} - X_{dev_{All}}$, where $X$ is equal to the ``First'', ``Second'', etc.}
\label{table:table8}
\end{table*}

\begin{table*}
\renewcommand\arraystretch{1.0}
\setlength\tabcolsep{11pt}
    \centering
\begin{subtable}[t]{0.3\linewidth}
\centering
\scalebox{0.5}{
    \begin{tabular}{ c | c c c c | c} 
     Task Type &    & \multicolumn{3}{c}{Open-Ended (BLEU-2)} &  \\ [0.5ex]
     \hline
     Method &    & \multicolumn{3}{c}{MUTAN without Attention} &  \\ [0.5ex]
     \hline
     Test Set&  \multicolumn{4}{c}{dev} & diff \\ [0.5ex]
     \hline
     Partition & Other & Num & Y/N & All & All \\ [0.5ex] 
     \hline
     First-dev & 2.55 & 2.93 & 25.09 & 11.84 & 48.32  \\ 
     
     Second-dev & 2.57 & 2.94 & 24.69 & 11.69 & 48.47 \\
     
     Third-dev & 2.66 & 2.84 & 24.54 & 11.66 & 48.50  \\
     
     Fourth-dev & 2.70 & 2.91 & 24.65 & 11.73 & 48.43  \\
     
     Fifth-dev & 2.68 & 2.80 & 24.73 & 11.74 & 48.42 \\
     
     Sixth-dev & 2.64 & 3.09 & 24.74 & 11.76 & 48.40  \\
     
     Seventh-dev & 2.59 & 2.95 & 24.66 & 11.69 & 48.47  \\
     \hline
     First-std & 2.33 & 2.63 & 24.71 & 11.59 & 48.86 \\
     \hline
     
     Original-dev  & 47.16 & 37.32 & 81.45 & 60.16 & -\\
     Original-std  & 47.57 & 36.75 & 81.56 & 60.45 & - \\

     \hline
    \end{tabular}}
    \centering
    \captionsetup{justification=centering}
    \caption{MUTAN without Attention model evaluation results.}

\scalebox{0.5}{
    \begin{tabular}{ c | c c c c | c} 
     Task Type &    & \multicolumn{3}{c}{Open-Ended (BLEU-2)} &  \\ [0.5ex]
     \hline
     Method &    & \multicolumn{3}{c}{HieCoAtt (Alt,VGG19)} &  \\ [0.5ex]
     \hline
     Test Set&  \multicolumn{4}{c}{dev} & diff \\ [0.5ex]
     \hline
     Partition & Other & Num & Y/N & All & All \\ [0.5ex] 
     \hline
     First-dev & 2.87 & 2.98 & 26.30 & 12.50 & 47.98  \\ 
     
     Second-dev & 2.87 & 2.85 & 26.12 & 12.41 & 48.07 \\
     
     Third-dev & 2.92 & 2.97 & 26.37 & 12.55 & 47.93  \\
     
     Fourth-dev & 3.04 & 2.96 & 26.14 & 12.51 & 47.97  \\
     
     Fifth-dev & 3.00 & 3.20 & 26.32 & 12.59 & 47.89 \\
     
     Sixth-dev & 3.07 & 3.02 & 26.10 & 12.52 & 47.96  \\
     
     Seventh-dev & 2.99 & 3.17 & 26.40 & 12.61 & 47.87  \\
     \hline
     First-std & 2.79 & 2.81 & 26.14 & 12.41 & 47.91 \\
     \hline

     Original-dev  & 49.14 & 38.35 & 79.63 & 60.48 & -\\
     Original-std  & 49.15 & 36.52 & 79.45 & 60.32 & - \\

     \hline
    \end{tabular}}
    \centering
    \captionsetup{justification=centering}
    \caption{HieCoAtt (Alt,VGG19) model evaluation results.}
\end{subtable}%
    \hfil
\begin{subtable}[t]{0.3\linewidth}

\centering
\scalebox{0.5}{
    \begin{tabular}{ c | c c c c | c} 
     Task Type &    & \multicolumn{3}{c}{Open-Ended (BLEU-2)} &  \\ [0.5ex]
     \hline
     Method &    & \multicolumn{3}{c}{MLB with Attention} &  \\ [0.5ex]
     \hline
     Test Set&  \multicolumn{4}{c}{dev} & diff \\ [0.5ex]
     \hline
     Partition & Other & Num & Y/N & All & All \\ [0.5ex] 
     \hline
     First-dev & 2.68 & 2.27 & 24.15 & 11.45 & 54.34  \\ 
     
     Second-dev & 2.82 & 2.28 & 24.22 & 11.54 & 54.25 \\
     
     Third-dev & 2.84 & 2.17 & 24.24 & 11.55 & 54.24  \\
     
     Fourth-dev & 2.82 & 2.15 & 24.08 & 11.47 & 54.32  \\
     
     Fifth-dev & 2.91 & 2.18 & 24.21 & 11.57 & 54.22 \\
     
     Sixth-dev & 2.83 & 2.32 & 24.12 & 11.51 & 54.28  \\
     
     Seventh-dev & 2.81 & 2.42 & 24.20 & 11.55 & 54.13  \\
     \hline
     First-std & 2.59 & 2.11 & 24.31 & 11.49 & 54.19 \\
     \hline
     
     Original-dev  & 57.01 & 37.51 & 83.54 & 65.79 & -\\
     Original-std  & 56.60 & 36.63 & 83.68 & 65.68 & - \\

     \hline
    \end{tabular}}
    \centering
    \captionsetup{justification=centering}
    \caption{MLB with Attention model evaluation results.}
    
\scalebox{0.5}{
    \begin{tabular}{ c | c c c c | c} 
     Task Type &    & \multicolumn{3}{c}{Open-Ended (BLEU-2)} &  \\ [0.5ex]
     \hline
     Method &    & \multicolumn{3}{c}{MUTAN with Attention} &  \\ [0.5ex]
     \hline
     Test Set&  \multicolumn{4}{c}{dev} & diff \\ [0.5ex]
     \hline
     Partition & Other & Num & Y/N & All & All \\ [0.5ex] 
     \hline
     First-dev & 2.03 & 2.14 & 24.38 & 11.21 & 54.77  \\ 
     
     Second-dev & 2.15 & 2.19 & 24.20 & 11.20 & 54.78 \\
     
     Third-dev & 2.07 & 2.31 & 24.29 & 11.21 & 54.77  \\
     
     Fourth-dev & 2.09 & 2.19 & 23.89 & 11.05 & 54.93  \\
     
     Fifth-dev & 2.14 & 2.30 & 24.15 & 11.19 & 54.79 \\
     
     Sixth-dev & 2.17 & 2.22 & 24.17 & 11.21 & 54.77  \\
     
     Seventh-dev & 1.95 & 2.38 & 24.20 & 11.13 & 54.85  \\
     \hline
     First-std & 1.92 & 2.16 & 24.41 & 11.21 & 54.56 \\
     \hline
     
     Original-dev  & 56.73 & 38.35 & 84.11 & 65.98 & -\\
     Original-std  & 56.29 & 37.47 & 84.04 & 65.77 & - \\

     \hline
    \end{tabular}}
    \centering
    \captionsetup{justification=centering}
    \caption{MUTAN with Attention model evaluation results.}
\end{subtable}%
    \hfil
\begin{subtable}[t]{0.3\linewidth}
        
\centering
\scalebox{0.5}{
    \begin{tabular}{ c | c c c c | c} 
     Task Type &    & \multicolumn{3}{c}{Open-Ended (BLEU-2)} &  \\ [0.5ex]
     \hline
     Method &    & \multicolumn{3}{c}{HieCoAtt (Alt,Resnet200)} &  \\ [0.5ex]
     \hline
     Test Set&  \multicolumn{4}{c}{dev} & diff \\ [0.5ex]
     \hline
     Partition & Other & Num & Y/N & All & All \\ [0.5ex] 
     \hline
     First-dev & 3.26 & 3.06 & 26.39 & 12.73 & 49.08  \\ 
     
     Second-dev & 3.22 & 3.19 & 26.22 & 12.66 & 49.15 \\
     
     Third-dev & 3.36 & 2.94 & 26.41 & 12.78 & 49.03  \\
     
     Fourth-dev & 3.43 & 3.02 & 25.97 & 12.64 & 49.17  \\
     
     Fifth-dev & 3.43 & 2.95 & 26.29 & 12.76 & 49.05 \\
     
     Sixth-dev & 3.42 & 2.88 & 26.31 & 12.76 & 49.05  \\
     
     Seventh-dev & 3.32 & 3.11 & 26.51 & 12.81 & 49.00  \\
     \hline
     First-std & 3.05 & 2.85 & 26.18 & 12.56 & 49.50\\
     \hline

     Original-dev  & 51.77 & 38.65 & 79.70 & 61.81 & -\\
     Original-std  & 51.95 & 38.22 & 79.95 & 62.06 & - \\

     \hline
    \end{tabular}}
    \centering
    \captionsetup{justification=centering}
    \caption{HieCoAtt (Alt,Resnet200) model evaluation results.}

\scalebox{0.5}{
    \begin{tabular}{ c | c c c c | c} 
     Task Type &    & \multicolumn{3}{c}{Open-Ended (BLEU-2)} &  \\ [0.5ex]
     \hline
     Method &    & \multicolumn{3}{c}{LSTM Q+I} &  \\ [0.5ex]
     \hline
     Test Set&  \multicolumn{4}{c}{dev} & diff \\ [0.5ex]
     \hline
     Partition & Other & Num & Y/N & All & All \\ [0.5ex] 
     \hline
     First-dev & 1.91 & 3.27 & 26.13 & 12.00 & 46.02  \\ 
     
     Second-dev & 1.89 & 3.25 & 26.06 & 11.96 & 46.06 \\
     
     Third-dev & 1.85 & 3.24 & 26.23 & 12.01 & 46.01  \\
     
     Fourth-dev & 1.93 & 3.34 & 25.80 & 11.88 & 46.14  \\
     
     Fifth-dev & 1.93 & 3.37 & 25.85 & 11.90 & 46.12 \\
     
     Sixth-dev & 1.95 & 3.41 & 26.04 & 11.99 & 46.03  \\
     
     Seventh-dev & 1.86 & 3.28 & 26.00 & 11.92 & 46.10  \\
     \hline
     First-std & 1.98 & 2.80 & 26.39 & 12.13 & 46.05 \\
     \hline
     
     Original-dev  & 43.40 & 36.46 & 80.87 & 58.02 & -\\
     Original-std  & 43.90 & 36.67 & 80.38 & 58.18 & - \\

     \hline
    \end{tabular}}
    \centering
    \captionsetup{justification=centering}
    \caption{LSTM Q+I model evaluation results.}
\end{subtable}
\caption{The table shows the six state-of-the-art pretrained VQA models evaluation results on the GBQD and VQA dataset. ``-'' indicates the results are not available, ``-std'' represents the accuracy of VQA model evaluated on the complete testing set of GBQD and VQA dataset and ``-dev'' indicates the accuracy of VQA model evaluated on the partial testing set of GBQD and VQA dataset. In addition, $diff = Original_{dev_{All}} - X_{dev_{All}}$, where $X$ is equal to the ``First'', ``Second'', etc.}
\label{table:table9}
\end{table*}

\begin{table*}
\renewcommand\arraystretch{1.0}
\setlength\tabcolsep{11pt}
    \centering
\begin{subtable}[t]{0.3\linewidth}
\centering
\scalebox{0.5}{
    \begin{tabular}{ c | c c c c | c} 
     Task Type &    & \multicolumn{3}{c}{Open-Ended (BLEU-3)} &  \\ [0.5ex]
     \hline
     Method &    & \multicolumn{3}{c}{MUTAN without Attention} &  \\ [0.5ex]
     \hline
     Test Set&  \multicolumn{4}{c}{dev} & diff \\ [0.5ex]
     \hline
     Partition & Other & Num & Y/N & All & All \\ [0.5ex] 
     \hline
     First-dev & 2.63 & 2.72 & 24.77 & 11.73 & 48.43  \\ 
     
     Second-dev & 2.66 & 2.67 & 24.72 & 11.71 & 48.45 \\
     
     Third-dev & 2.71 & 2.53 & 24.66 & 11.70 & 48.46  \\
     
     Fourth-dev & 2.66 & 2.81 & 24.59 & 11.68 & 48.48  \\
     
     Fifth-dev & 2.72 & 2.64 & 25.00 & 11.85 & 48.31 \\
     
     Sixth-dev & 2.58 & 2.64 & 24.72 & 11.67 & 48.49  \\
     
     Seventh-dev & 2.73 & 2.56 & 24.78 & 11.76 & 48.40  \\
     \hline
     First-std & 2.60 & 3.04 & 24.33 & 11.60 & 48.85 \\
     \hline
     
     Original-dev  & 47.16 & 37.32 & 81.45 & 60.16 & -\\
     Original-std  & 47.57 & 36.75 & 81.56 & 60.45 & - \\

     \hline
    \end{tabular}}
    \centering
    \captionsetup{justification=centering}
    \caption{MUTAN without Attention model evaluation results.}

\scalebox{0.5}{
    \begin{tabular}{ c | c c c c | c} 
     Task Type &    & \multicolumn{3}{c}{Open-Ended (BLEU-3)} &  \\ [0.5ex]
     \hline
     Method &    & \multicolumn{3}{c}{HieCoAtt (Alt,VGG19)} &  \\ [0.5ex]
     \hline
     Test Set&  \multicolumn{4}{c}{dev} & diff \\ [0.5ex]
     \hline
     Partition & Other & Num & Y/N & All & All \\ [0.5ex] 
     \hline
     First-dev & 2.85 & 2.81 & 26.42 & 12.52 & 47.96  \\ 
     
     Second-dev & 2.96 & 2.90 & 26.52 & 12.62 & 47.86 \\
     
     Third-dev & 2.98 & 2.91 & 26.47 & 12.61 & 47.87  \\
     
     Fourth-dev & 3.03 & 3.05 & 26.52 & 12.67 & 47.81  \\
     
     Fifth-dev & 3.02 & 3.19 & 26.55 & 12.69 & 47.79 \\
     
     Sixth-dev & 3.17 & 3.27 & 26.41 & 12.72 & 47.76  \\
     
     Seventh-dev & 3.21 & 3.03 & 26.36 & 12.70 & 47.78  \\
     \hline
     First-std & 2.80 & 2.91 & 25.99 & 12.37 & 47.95 \\
     \hline

     Original-dev  & 49.14 & 38.35 & 79.63 & 60.48 & -\\
     Original-std  & 49.15 & 36.52 & 79.45 & 60.32 & - \\

     \hline
    \end{tabular}}
    \centering
    \captionsetup{justification=centering}
    \caption{HieCoAtt (Alt,VGG19) model evaluation results.}
\end{subtable}%
    \hfil
\begin{subtable}[t]{0.3\linewidth}

\centering
\scalebox{0.5}{
    \begin{tabular}{ c | c c c c | c} 
     Task Type &    & \multicolumn{3}{c}{Open-Ended (BLEU-3)} &  \\ [0.5ex]
     \hline
     Method &    & \multicolumn{3}{c}{MLB with Attention} &  \\ [0.5ex]
     \hline
     Test Set&  \multicolumn{4}{c}{dev} & diff \\ [0.5ex]
     \hline
     Partition & Other & Num & Y/N & All & All \\ [0.5ex] 
     \hline
     First-dev & 2.77 & 2.00 & 24.43 & 11.58 & 54.21  \\ 
     
     Second-dev & 2.84 & 2.09 & 24.19 & 11.52 & 54.27 \\
     
     Third-dev & 2.92 & 1.93 & 24.01 & 11.47 & 54.32  \\
     
     Fourth-dev & 2.97 & 1.97 & 24.03 & 11.51 & 54.28  \\
     
     Fifth-dev & 2.90 & 1.97 & 23.92 & 11.43 & 54.36 \\
     
     Sixth-dev & 2.90 & 2.12 & 24.02 & 11.49 & 54.30  \\
     
     Seventh-dev & 2.96 & 2.06 & 23.80 & 11.42 & 54.37  \\
     \hline
     First-std & 2.65 & 2.23 & 24.20 & 11.48 & 54.20 \\
     \hline
     
     Original-dev  & 57.01 & 37.51 & 83.54 & 65.79 & -\\
     Original-std  & 56.60 & 36.63 & 83.68 & 65.68 & - \\

     \hline
    \end{tabular}}
    \centering
    \captionsetup{justification=centering}
    \caption{MLB with Attention model evaluation results.}
    
\scalebox{0.5}{
    \begin{tabular}{ c | c c c c | c} 
     Task Type &    & \multicolumn{3}{c}{Open-Ended (BLEU-3)} &  \\ [0.5ex]
     \hline
     Method &    & \multicolumn{3}{c}{MUTAN with Attention} &  \\ [0.5ex]
     \hline
     Test Set&  \multicolumn{4}{c}{dev} & diff \\ [0.5ex]
     \hline
     Partition & Other & Num & Y/N & All & All \\ [0.5ex] 
     \hline
     First-dev & 2.01 & 2.18 & 24.36 & 11.20 & 54.78  \\ 
     
     Second-dev & 2.09 & 2.12 & 24.06 & 11.11 & 54.87 \\
     
     Third-dev & 2.08 & 2.15 & 24.25 & 11.19 & 54.79  \\
     
     Fourth-dev & 2.14 & 2.09 & 24.08 & 11.14 & 54.84  \\
     
     Fifth-dev & 2.05 & 2.00 & 24.10 & 11.09 & 54.89 \\
     
     Sixth-dev & 2.04 & 2.25 & 24.20 & 11.16 & 54.82  \\
     
     Seventh-dev & 2.06 & 2.26 & 23.87 & 11.03 & 54.95  \\
     \hline
     First-std & 2.06 & 2.15 & 24.13 & 11.16 & 54.61 \\
     \hline
     
     Original-dev  & 56.73 & 38.35 & 84.11 & 65.98 & -\\
     Original-std  & 56.29 & 37.47 & 84.04 & 65.77 & - \\

     \hline
    \end{tabular}}
    \centering
    \captionsetup{justification=centering}
    \caption{MUTAN with Attention model evaluation results.}
\end{subtable}%
    \hfil
\begin{subtable}[t]{0.3\linewidth}
        
\centering
\scalebox{0.5}{
    \begin{tabular}{ c | c c c c | c} 
     Task Type &    & \multicolumn{3}{c}{Open-Ended (BLEU-3)} &  \\ [0.5ex]
     \hline
     Method &    & \multicolumn{3}{c}{HieCoAtt (Alt,Resnet200)} &  \\ [0.5ex]
     \hline
     Test Set&  \multicolumn{4}{c}{dev} & diff \\ [0.5ex]
     \hline
     Partition & Other & Num & Y/N & All & All \\ [0.5ex] 
     \hline
     First-dev & 3.33 & 3.07 & 26.58 & 12.84 & 48.97  \\ 
     
     Second-dev & 3.25 & 3.04 & 26.09 & 12.64 & 49.17 \\
     
     Third-dev & 3.48 & 3.00 & 26.53 & 12.89 & 48.92  \\
     
     Fourth-dev & 3.43 & 2.99 & 26.40 & 12.81 & 49.00  \\
     
     Fifth-dev & 3.45 & 3.09 & 26.35 & 12.81 & 49.00 \\
     
     Sixth-dev & 3.41 & 2.99 & 26.62 & 12.89 & 48.92  \\
     
     Seventh-dev & 3.46 & 2.95 & 26.32 & 12.79 & 49.02  \\
     \hline
     First-std & 3.27 & 2.90 & 26.23 & 12.69 & 49.37\\
     \hline

     Original-dev  & 51.77 & 38.65 & 79.70 & 61.81 & -\\
     Original-std  & 51.95 & 38.22 & 79.95 & 62.06 & - \\

     \hline
    \end{tabular}}
    \centering
    \captionsetup{justification=centering}
    \caption{HieCoAtt (Alt,Resnet200) model evaluation results.}

\scalebox{0.5}{
    \begin{tabular}{ c | c c c c | c} 
     Task Type &    & \multicolumn{3}{c}{Open-Ended (BLEU-3)} &  \\ [0.5ex]
     \hline
     Method &    & \multicolumn{3}{c}{LSTM Q+I} &  \\ [0.5ex]
     \hline
     Test Set&  \multicolumn{4}{c}{dev} & diff \\ [0.5ex]
     \hline
     Partition & Other & Num & Y/N & All & All \\ [0.5ex] 
     \hline
     First-dev & 2.02 & 3.23 & 26.32 & 12.12 & 45.90  \\ 
     
     Second-dev & 2.08 & 3.14 & 26.01 & 12.02 & 46.00 \\
     
     Third-dev & 1.96 & 3.26 & 26.12 & 12.02 & 46.00  \\
     
     Fourth-dev & 2.05 & 3.28 & 25.95 & 11.99 & 46.03  \\
     
     Fifth-dev & 2.07 & 3.36 & 26.26 & 12.14 & 45.88 \\
     
     Sixth-dev & 2.10 & 3.29 & 25.93 & 12.01 & 46.01  \\
     
     Seventh-dev & 2.15 & 3.19 & 26.12 & 12.10 & 45.92  \\
     \hline
     First-std & 1.88 & 3.26 & 25.96 & 11.95 & 46.23 \\
     \hline
     
     Original-dev  & 43.40 & 36.46 & 80.87 & 58.02 & -\\
     Original-std  & 43.90 & 36.67 & 80.38 & 58.18 & - \\

     \hline
    \end{tabular}}
    \centering
    \captionsetup{justification=centering}
    \caption{LSTM Q+I model evaluation results.}
\end{subtable}
\caption{The table shows the six state-of-the-art pretrained VQA models evaluation results on the GBQD and VQA dataset. ``-'' indicates the results are not available, ``-std'' represents the accuracy of VQA model evaluated on the complete testing set of GBQD and VQA dataset and ``-dev'' indicates the accuracy of VQA model evaluated on the partial testing set of GBQD and VQA dataset. In addition, $diff = Original_{dev_{All}} - X_{dev_{All}}$, where $X$ is equal to the ``First'', ``Second'', etc.}
\label{table:table10}
\end{table*}

\begin{table*}
\renewcommand\arraystretch{1.0}
\setlength\tabcolsep{11pt}
    \centering
\begin{subtable}[t]{0.3\linewidth}
\centering
\scalebox{0.5}{
    \begin{tabular}{ c | c c c c | c} 
     Task Type &    & \multicolumn{3}{c}{Open-Ended (BLEU-4)} &  \\ [0.5ex]
     \hline
     Method &    & \multicolumn{3}{c}{MUTAN without Attention} &  \\ [0.5ex]
     \hline
     Test Set&  \multicolumn{4}{c}{dev} & diff \\ [0.5ex]
     \hline
     Partition & Other & Num & Y/N & All & All \\ [0.5ex] 
     \hline
     First-dev & 2.46 & 2.98 & 25.22 & 11.86 & 48.30  \\ 
     
     Second-dev & 2.47 & 3.05 & 25.23 & 11.88 & 48.28 \\
     
     Third-dev & 2.62 & 2.90 & 24.95 & 11.81 & 48.35  \\
     
     Fourth-dev & 2.71 & 2.96 & 24.87 & 11.83 & 48.33  \\
     
     Fifth-dev & 2.70 & 3.03 & 25.08 & 11.92 & 48.24 \\
     
     Sixth-dev & 2.65 & 2.84 & 25.30 & 11.97 & 48.19  \\
     
     Seventh-dev & 2.71 & 2.99 & 25.01 & 11.89 & 48.27  \\
     \hline
     First-std & 2.51 & 2.36 & 24.36 & 11.50 & 48.95 \\
     \hline
     
     Original-dev  & 47.16 & 37.32 & 81.45 & 60.16 & -\\
     Original-std  & 47.57 & 36.75 & 81.56 & 60.45 & - \\

     \hline
    \end{tabular}}
    \centering
    \captionsetup{justification=centering}
    \caption{MUTAN without Attention model evaluation results.}

\scalebox{0.5}{
    \begin{tabular}{ c | c c c c | c} 
     Task Type &    & \multicolumn{3}{c}{Open-Ended (BLEU-4)} &  \\ [0.5ex]
     \hline
     Method &    & \multicolumn{3}{c}{HieCoAtt (Alt,VGG19)} &  \\ [0.5ex]
     \hline
     Test Set&  \multicolumn{4}{c}{dev} & diff \\ [0.5ex]
     \hline
     Partition & Other & Num & Y/N & All & All \\ [0.5ex] 
     \hline
     First-dev & 2.80 & 3.17 & 26.55 & 12.59 & 47.89  \\ 
     
     Second-dev & 2.87 & 3.14 & 27.02 & 12.81 & 47.67 \\
     
     Third-dev & 3.02 & 2.93 & 26.60 & 12.69 & 47.79  \\
     
     Fourth-dev & 3.08 & 3.14 & 26.29 & 12.61 & 47.87  \\
     
     Fifth-dev & 3.09 & 3.28 & 26.52 & 12.73 & 47.75 \\
     
     Sixth-dev & 3.11 & 3.20 & 26.66 & 12.78 & 47.70  \\
     
     Seventh-dev & 3.03 & 3.26 & 26.71 & 12.77 & 47.71  \\
     \hline
     First-std & 2.73 & 2.46 & 25.81 & 12.21 & 48.11 \\
     \hline

     Original-dev  & 49.14 & 38.35 & 79.63 & 60.48 & -\\
     Original-std  & 49.15 & 36.52 & 79.45 & 60.32 & - \\

     \hline
    \end{tabular}}
    \centering
    \captionsetup{justification=centering}
    \caption{HieCoAtt (Alt,VGG19) model evaluation results.}
\end{subtable}%
    \hfil
\begin{subtable}[t]{0.3\linewidth}

\centering
\scalebox{0.5}{
    \begin{tabular}{ c | c c c c | c} 
     Task Type &    & \multicolumn{3}{c}{Open-Ended (BLEU-4)} &  \\ [0.5ex]
     \hline
     Method &    & \multicolumn{3}{c}{MLB with Attention} &  \\ [0.5ex]
     \hline
     Test Set&  \multicolumn{4}{c}{dev} & diff \\ [0.5ex]
     \hline
     Partition & Other & Num & Y/N & All & All \\ [0.5ex] 
     \hline
     First-dev & 2.65 & 2.41 & 24.63 & 11.64 & 54.15  \\ 
     
     Second-dev & 2.72 & 2.47 & 24.63 & 11.69 & 54.10 \\
     
     Third-dev & 2.83 & 2.40 & 24.62 & 11.73 & 54.06  \\
     
     Fourth-dev & 2.88 & 2.38 & 24.28 & 11.61 & 54.18  \\
     
     Fifth-dev & 2.79 & 2.31 & 24.40 & 11.61 & 54.18 \\
     
     Sixth-dev & 2.89 & 2.36 & 24.31 & 11.63 & 54.16  \\
     
     Seventh-dev & 2.80 & 2.51 & 24.52 & 11.68 & 54.11  \\
     \hline
     First-std & 2.58 & 1.85 & 23.54 & 11.14 & 54.54 \\
     \hline
     
     Original-dev  & 57.01 & 37.51 & 83.54 & 65.79 & -\\
     Original-std  & 56.60 & 36.63 & 83.68 & 65.68 & - \\

     \hline
    \end{tabular}}
    \centering
    \captionsetup{justification=centering}
    \caption{MLB with Attention model evaluation results.}
    
\scalebox{0.5}{
    \begin{tabular}{ c | c c c c | c} 
     Task Type &    & \multicolumn{3}{c}{Open-Ended (BLEU-4)} &  \\ [0.5ex]
     \hline
     Method &    & \multicolumn{3}{c}{MUTAN with Attention} &  \\ [0.5ex]
     \hline
     Test Set&  \multicolumn{4}{c}{dev} & diff \\ [0.5ex]
     \hline
     Partition & Other & Num & Y/N & All & All \\ [0.5ex] 
     \hline
     First-dev & 2.00 & 2.35 & 24.55 & 11.29 & 54.69  \\ 
     
     Second-dev & 2.05 & 2.21 & 24.30 & 11.20 & 54.78 \\
     
     Third-dev & 2.01 & 2.32 & 24.61 & 11.32 & 54.66  \\
     
     Fourth-dev & 2.11 & 2.39 & 24.18 & 11.20 & 54.78  \\
     
     Fifth-dev & 1.94 & 2.37 & 24.47 & 11.23 & 54.75 \\
     
     Sixth-dev & 2.08 & 2.43 & 24.39 & 11.27 & 54.71  \\
     
     Seventh-dev & 2.00 & 2.35 & 24.23 & 11.16 & 54.82  \\
     \hline
     First-std & 1.98 & 1.94 & 23.62 & 10.90 & 54.87 \\
     \hline
     
     Original-dev  & 56.73 & 38.35 & 84.11 & 65.98 & -\\
     Original-std  & 56.29 & 37.47 & 84.04 & 65.77 & - \\

     \hline
    \end{tabular}}
    \centering
    \captionsetup{justification=centering}
    \caption{MUTAN with Attention model evaluation results.}
\end{subtable}%
    \hfil
\begin{subtable}[t]{0.3\linewidth}
        
\centering
\scalebox{0.5}{
    \begin{tabular}{ c | c c c c | c} 
     Task Type &    & \multicolumn{3}{c}{Open-Ended (BLEU-4)} &  \\ [0.5ex]
     \hline
     Method &    & \multicolumn{3}{c}{HieCoAtt (Alt,Resnet200)} &  \\ [0.5ex]
     \hline
     Test Set&  \multicolumn{4}{c}{dev} & diff \\ [0.5ex]
     \hline
     Partition & Other & Num & Y/N & All & All \\ [0.5ex] 
     \hline
     First-dev & 3.01& 3.37 & 26.55 & 12.71 & 49.10  \\ 
     
     Second-dev & 3.08 & 3.34 & 26.84 & 12.86 & 48.95 \\
     
     Third-dev & 3.23 & 3.03 & 26.71 & 12.85 & 48.96  \\
     
     Fourth-dev & 3.24 & 3.16 & 26.31 & 12.70 & 49.11  \\
     
     Fifth-dev & 3.35 & 3.10 & 26.16 & 12.68 & 49.13 \\
     
     Sixth-dev & 3.34 & 3.25 & 26.66 & 12.90 & 48.91  \\
     
     Seventh-dev & 3.21 & 3.23 & 26.56 & 12.79 & 49.02  \\
     \hline
     First-std & 3.25 & 2.52 & 25.84 & 12.49 & 49.57\\
     \hline

     Original-dev  & 51.77 & 38.65 & 79.70 & 61.81 & -\\
     Original-std  & 51.95 & 38.22 & 79.95 & 62.06 & - \\

     \hline
    \end{tabular}}
    \centering
    \captionsetup{justification=centering}
    \caption{HieCoAtt (Alt,Resnet200) model evaluation results.}

\scalebox{0.5}{
    \begin{tabular}{ c | c c c c | c} 
     Task Type &    & \multicolumn{3}{c}{Open-Ended (BLEU-4)} &  \\ [0.5ex]
     \hline
     Method &    & \multicolumn{3}{c}{LSTM Q+I} &  \\ [0.5ex]
     \hline
     Test Set&  \multicolumn{4}{c}{dev} & diff \\ [0.5ex]
     \hline
     Partition & Other & Num & Y/N & All & All \\ [0.5ex] 
     \hline
     First-dev & 1.84 & 3.20 & 26.41 & 12.07 & 45.95  \\ 
     
     Second-dev & 1.87 & 3.22 & 26.38 & 12.08 & 45.94 \\
     
     Third-dev & 1.93 & 3.28 & 26.41 & 12.12 & 45.90  \\
     
     Fourth-dev & 1.85 & 3.24 & 26.16 & 11.98 & 46.04  \\
     
     Fifth-dev & 1.91 & 3.32 & 26.26 & 12.06 & 45.96 \\
     
     Sixth-dev & 1.90 & 3.27 & 26.16 & 12.00 & 46.02  \\
     
     Seventh-dev & 1.97 & 3.31 & 26.07 & 12.00 & 46.02  \\
     \hline
     First-std & 2.03 & 2.86 & 25.73 & 11.88 & 46.30 \\
     \hline
     
     Original-dev  & 43.40 & 36.46 & 80.87 & 58.02 & -\\
     Original-std  & 43.90 & 36.67 & 80.38 & 58.18 & - \\

     \hline
    \end{tabular}}
    \centering
    \captionsetup{justification=centering}
    \caption{LSTM Q+I model evaluation results.}
\end{subtable}
\caption{The table shows the six state-of-the-art pretrained VQA models evaluation results on the GBQD and VQA dataset. ``-'' indicates the results are not available, ``-std'' represents the accuracy of VQA model evaluated on the complete testing set of GBQD and VQA dataset and ``-dev'' indicates the accuracy of VQA model evaluated on the partial testing set of GBQD and VQA dataset. In addition, $diff = Original_{dev_{All}} - X_{dev_{All}}$, where $X$ is equal to the ``First'', ``Second'', etc.}
\label{table:table11}
\end{table*}

\begin{table*}
\renewcommand\arraystretch{1.0}
\setlength\tabcolsep{11pt}
    \centering
\begin{subtable}[t]{0.3\linewidth}
\centering
\scalebox{0.5}{
    \begin{tabular}{ c | c c c c | c} 
     Task Type &    & \multicolumn{3}{c}{Open-Ended (ROUGE)} &  \\ [0.5ex]
     \hline
     Method &    & \multicolumn{3}{c}{MUTAN without Attention} &  \\ [0.5ex]
     \hline
     Test Set&  \multicolumn{4}{c}{dev} & diff \\ [0.5ex]
     \hline
     Partition & Other & Num & Y/N & All & All \\ [0.5ex] 
     \hline
     First-dev & 2.68 & 2.66 & 26.41 & 12.42 & 47.74  \\ 
     
     Second-dev & 3.40 & 3.12 & 25.47 & 12.43 & 47.73 \\
     
     Third-dev & 3.38 & 2.60 & 21.83 & 10.87 & 49.29  \\
     
     Fourth-dev & 3.04 & 2.17 & 23.19 & 11.21 & 48.95  \\
     
     Fifth-dev & 2.93 & 2.77 & 26.22 & 12.47 & 47.69 \\
     
     Sixth-dev & 2.43 & 2.66 & 27.14 & 12.60 & 47.56  \\
     
     Seventh-dev & 1.66 & 2.73 & 26.90 & 12.13 & 48.03  \\
     \hline
     First-std & 2.69 & 2.57 & 26.20 & 12.36 & 48.09 \\
     \hline
     
     Original-dev  & 47.16 & 37.32 & 81.45 & 60.16 & -\\
     Original-std  & 47.57 & 36.75 & 81.56 & 60.45 & - \\

     \hline
    \end{tabular}}
    \centering
    \captionsetup{justification=centering}
    \caption{MUTAN without Attention model evaluation results.}

\scalebox{0.5}{
    \begin{tabular}{ c | c c c c | c} 
     Task Type &    & \multicolumn{3}{c}{Open-Ended (ROUGE)} &  \\ [0.5ex]
     \hline
     Method &    & \multicolumn{3}{c}{HieCoAtt (Alt,VGG19)} &  \\ [0.5ex]
     \hline
     Test Set&  \multicolumn{4}{c}{dev} & diff \\ [0.5ex]
     \hline
     Partition & Other & Num & Y/N & All & All \\ [0.5ex] 
     \hline
     First-dev & 2.88 & 3.73 & 24.78 & 11.96 & 48.52  \\ 
     
     Second-dev & 3.26 & 3.75 & 27.49 & 13.26 & 47.22 \\
     
     Third-dev & 3.11 & 3.41 & 27.73 & 13.25 & 47.23  \\
     
     Fourth-dev & 3.05 & 3.20 & 25.74 & 12.38 & 48.10  \\
     
     Fifth-dev & 3.13 & 3.56 & 28.27 & 13.49 & 46.99 \\
     
     Sixth-dev & 3.33 & 3.35 & 27.67 & 13.32 & 47.16  \\
     
     Seventh-dev & 2.78 & 3.58 & 28.09 & 13.25 & 47.23  \\
     \hline
     First-std & 2.73 & 3.41 & 24.01 & 11.57 & 48.75 \\
     \hline

     Original-dev  & 49.14 & 38.35 & 79.63 & 60.48 & -\\
     Original-std  & 49.15 & 36.52 & 79.45 & 60.32 & - \\

     \hline
    \end{tabular}}
    \centering
    \captionsetup{justification=centering}
    \caption{HieCoAtt (Alt,VGG19) model evaluation results.}
\end{subtable}%
    \hfil
\begin{subtable}[t]{0.3\linewidth}

\centering
\scalebox{0.5}{
    \begin{tabular}{ c | c c c c | c} 
     Task Type &    & \multicolumn{3}{c}{Open-Ended (ROUGE)} &  \\ [0.5ex]
     \hline
     Method &    & \multicolumn{3}{c}{MLB with Attention} &  \\ [0.5ex]
     \hline
     Test Set&  \multicolumn{4}{c}{dev} & diff \\ [0.5ex]
     \hline
     Partition & Other & Num & Y/N & All & All \\ [0.5ex] 
     \hline
     First-dev & 3.08 & 2.59 & 24.70 & 11.90 & 53.89  \\ 
     
     Second-dev & 3.20 & 2.88 & 24.81 & 12.03 & 53.76 \\
     
     Third-dev & 3.10 & 2.63 & 22.57 & 11.04 & 54.75  \\
     
     Fourth-dev & 3.23 & 2.60 & 22.82 & 11.20 & 54.59  \\
     
     Fifth-dev & 3.20 & 2.42 & 24.75 & 11.96 & 53.83 \\
     
     Sixth-dev & 2.92 & 2.61 & 24.49 & 11.74 & 54.05  \\
     
     Seventh-dev & 2.67 & 2.62 & 27.49 & 12.85 & 52.94  \\
     \hline
     First-std & 2.94 & 2.38 & 24.16 & 11.63 & 54.05 \\
     \hline
     
     Original-dev  & 57.01 & 37.51 & 83.54 & 65.79 & -\\
     Original-std  & 56.60 & 36.63 & 83.68 & 65.68 & - \\

     \hline
    \end{tabular}}
    \centering
    \captionsetup{justification=centering}
    \caption{MLB with Attention model evaluation results.}
    
\scalebox{0.5}{
    \begin{tabular}{ c | c c c c | c} 
     Task Type &    & \multicolumn{3}{c}{Open-Ended (ROUGE)} &  \\ [0.5ex]
     \hline
     Method &    & \multicolumn{3}{c}{MUTAN with Attention} &  \\ [0.5ex]
     \hline
     Test Set&  \multicolumn{4}{c}{dev} & diff \\ [0.5ex]
     \hline
     Partition & Other & Num & Y/N & All & All \\ [0.5ex] 
     \hline
     First-dev & 2.14 & 2.02 & 25.38 & 11.66 & 54.32  \\ 
     
     Second-dev & 2.12 & 2.01 & 22.84 & 10.70 & 55.28 \\
     
     Third-dev & 2.13 & 2.37 & 22.73 & 10.61 & 55.37  \\
     
     Fourth-dev & 2.04 & 2.26 & 22.70 & 10.55 & 55.43  \\
     
     Fifth-dev & 1.97 & 2.26 & 22.72 & 10.52 & 55.46 \\
     
     Sixth-dev & 2.25 & 2.63 & 23.91 & 11.18 & 54.80  \\
     
     Seventh-dev & 1.93 & 2.63 & 25.10 & 11.51 & 54.47  \\
     \hline
     First-std & 2.04 & 1.88 & 24.83 & 11.42 & 54.35 \\
     \hline
     
     Original-dev  & 56.73 & 38.35 & 84.11 & 65.98 & -\\
     Original-std  & 56.29 & 37.47 & 84.04 & 65.77 & - \\

     \hline
    \end{tabular}}
    \centering
    \captionsetup{justification=centering}
    \caption{MUTAN with Attention model evaluation results.}
\end{subtable}%
    \hfil
\begin{subtable}[t]{0.3\linewidth}
        
\centering
\scalebox{0.5}{
    \begin{tabular}{ c | c c c c | c} 
     Task Type &    & \multicolumn{3}{c}{Open-Ended (ROUGE)} &  \\ [0.5ex]
     \hline
     Method &    & \multicolumn{3}{c}{HieCoAtt (Alt,Resnet200)} &  \\ [0.5ex]
     \hline
     Test Set&  \multicolumn{4}{c}{dev} & diff \\ [0.5ex]
     \hline
     Partition & Other & Num & Y/N & All & All \\ [0.5ex] 
     \hline
     First-dev & 3.68 & 3.13 & 27.23 & 13.29 & 48.52  \\ 
     
     Second-dev & 3.83 & 3.76 & 27.45 & 13.52 & 48.29 \\
     
     Third-dev & 3.41 & 3.28 & 27.26 & 13.19 & 48.62  \\
     
     Fourth-dev & 3.25 & 3.37 & 25.69 & 12.47 & 49.34  \\
     
     Fifth-dev & 3.33 & 3.47 & 28.33 & 13.60 & 48.21 \\
     
     Sixth-dev & 3.56 & 2.99 & 27.83 & 13.46 & 48.35  \\
     
     Seventh-dev & 2.45 & 3.38 & 28.14 & 13.09 & 48.72  \\
     \hline
     First-std & 3.39 & 3.01 & 26.51 & 12.88 & 49.18\\
     \hline

     Original-dev  & 51.77 & 38.65 & 79.70 & 61.81 & -\\
     Original-std  & 51.95 & 38.22 & 79.95 & 62.06 & - \\

     \hline
    \end{tabular}}
    \centering
    \captionsetup{justification=centering}
    \caption{HieCoAtt (Alt,Resnet200) model evaluation results.}

\scalebox{0.5}{
    \begin{tabular}{ c | c c c c | c} 
     Task Type &    & \multicolumn{3}{c}{Open-Ended (ROUGE)} &  \\ [0.5ex]
     \hline
     Method &    & \multicolumn{3}{c}{LSTM Q+I} &  \\ [0.5ex]
     \hline
     Test Set&  \multicolumn{4}{c}{dev} & diff \\ [0.5ex]
     \hline
     Partition & Other & Num & Y/N & All & All \\ [0.5ex] 
     \hline
     First-dev & 1.71 & 3.56 & 26.51 & 12.09 & 45.93  \\ 
     
     Second-dev & 2.01 & 3.40 & 26.09 & 12.04 & 45.98 \\
     
     Third-dev & 1.91 & 2.92 & 23.70 & 10.96 & 47.06  \\
     
     Fourth-dev & 1.61 & 3.37 & 25.35 & 11.54 & 46.48  \\
     
     Fifth-dev & 1.57 & 3.32 & 25.92 & 11.75 & 46.27 \\
     
     Sixth-dev & 2.21 & 2.79 & 27.24 & 12.54 & 45.48  \\
     
     Seventh-dev & 1.58 & 2.99 & 27.26 & 12.27 & 45.75  \\
     \hline
     First-std & 1.79 & 3.42 & 26.42 & 12.11 & 46.07 \\
     \hline
     
     Original-dev  & 43.40 & 36.46 & 80.87 & 58.02 & -\\
     Original-std  & 43.90 & 36.67 & 80.38 & 58.18 & - \\

     \hline
    \end{tabular}}
    \centering
    \captionsetup{justification=centering}
    \caption{LSTM Q+I model evaluation results.}
\end{subtable}
\caption{The table shows the six state-of-the-art pretrained VQA models evaluation results on the GBQD and VQA dataset. ``-'' indicates the results are not available, ``-std'' represents the accuracy of VQA model evaluated on the complete testing set of GBQD and VQA dataset and ``-dev'' indicates the accuracy of VQA model evaluated on the partial testing set of GBQD and VQA dataset. In addition, $diff = Original_{dev_{All}} - X_{dev_{All}}$, where $X$ is equal to the ``First'', ``Second'', etc.}
\label{table:table12}
\end{table*}

\begin{table*}
\renewcommand\arraystretch{1.0}
\setlength\tabcolsep{11pt}
    \centering
\begin{subtable}[t]{0.3\linewidth}
\centering
\scalebox{0.5}{
    \begin{tabular}{ c | c c c c | c} 
     Task Type &    & \multicolumn{3}{c}{Open-Ended (CIDEr)} &  \\ [0.5ex]
     \hline
     Method &    & \multicolumn{3}{c}{MUTAN without Attention} &  \\ [0.5ex]
     \hline
     Test Set&  \multicolumn{4}{c}{dev} & diff \\ [0.5ex]
     \hline
     Partition & Other & Num & Y/N & All & All \\ [0.5ex] 
     \hline
     First-dev & 0.62 & 12.76 & 0.18 & 1.75 & 58.41  \\ 
     
     Second-dev & 2.47 & 2.89 & 21.03 & 10.13 & 50.03 \\
     
     Third-dev & 0.96 & 0.33 & 1.07 & 0.94 & 59.22  \\
     
     Fourth-dev & 1.44 & 1.79 & 12.32 & 5.94 & 54.22  \\
     
     Fifth-dev & 2.73 & 3.09 & 29.80 & 13.88 & 46.28 \\
     
     Sixth-dev & 1.56 & 2.26 & 31.64 & 11.98 & 48.18  \\
     
     Seventh-dev & 0.83 & 1.10 & 14.58 & 6.50 & 53.66  \\
     \hline
     First-std & 0.66 & 12.92 & 0.12 & 1.72 & 58.73 \\
     \hline
     
     Original-dev  & 47.16 & 37.32 & 81.45 & 60.16 & -\\
     Original-std  & 47.57 & 36.75 & 81.56 & 60.45 & - \\

     \hline
    \end{tabular}}
    \centering
    \captionsetup{justification=centering}
    \caption{MUTAN without Attention model evaluation results.}

\scalebox{0.5}{
    \begin{tabular}{ c | c c c c | c} 
     Task Type &    & \multicolumn{3}{c}{Open-Ended (CIDEr)} &  \\ [0.5ex]
     \hline
     Method &    & \multicolumn{3}{c}{HieCoAtt (Alt,VGG19)} &  \\ [0.5ex]
     \hline
     Test Set&  \multicolumn{4}{c}{dev} & diff \\ [0.5ex]
     \hline
     Partition & Other & Num & Y/N & All & All \\ [0.5ex] 
     \hline
     First-dev & 1.60 & 4.80 & 0.09 & 1.33 & 59.15  \\ 
     
     Second-dev & 2.51 & 2.00 & 20.72 & 9.93 & 50.55 \\
     
     Third-dev & 1.21 & 0.35 & 0.06 & 0.65 & 59.83  \\
     
     Fourth-dev & 5.53 & 2.03 & 6.98 & 5.75 & 54.73  \\
     
     Fifth-dev & 2.34 & 2.78 & 25.13 & 11.74 & 48.74 \\
     
     Sixth-dev & 2.43 & 3.61 & 29.75 & 13.77 & 46.71  \\
     
     Seventh-dev & 1.66 & 1.89 & 15.72 & 7.45 & 53.03  \\
     \hline
     First-std & 1.51 & 5.25 & 0.08 & 12.38 & 59.01 \\
     \hline

     Original-dev  & 49.14 & 38.35 & 79.63 & 60.48 & -\\
     Original-std  & 49.15 & 36.52 & 79.45 & 60.32 & - \\

     \hline
    \end{tabular}}
    \centering
    \captionsetup{justification=centering}
    \caption{HieCoAtt (Alt,VGG19) model evaluation results.}
\end{subtable}%
    \hfil
\begin{subtable}[t]{0.3\linewidth}

\centering
\scalebox{0.5}{
    \begin{tabular}{ c | c c c c | c} 
     Task Type &    & \multicolumn{3}{c}{Open-Ended (CIDEr)} &  \\ [0.5ex]
     \hline
     Method &    & \multicolumn{3}{c}{MLB with Attention} &  \\ [0.5ex]
     \hline
     Test Set&  \multicolumn{4}{c}{dev} & diff \\ [0.5ex]
     \hline
     Partition & Other & Num & Y/N & All & All \\ [0.5ex] 
     \hline
     First-dev & 1.57 & 21.95 & 1.61 & 3.80 & 61.99  \\ 
     
     Second-dev & 2.48 & 2.90 & 21.91 & 10.50 & 55.29 \\
     
     Third-dev & 2.09 & 2.26 & 22.52 & 10.50 & 55.29  \\
     
     Fourth-dev & 3.20 & 2.91 & 25.53 & 12.33 & 53.46  \\
     
     Fifth-dev & 2.18 & 3.37 & 27.05 & 12.51 & 53.28 \\
     
     Sixth-dev & 2.48 & 2.19 & 32.04 & 14.58 & 51.21  \\
     
     Seventh-dev & 1.68 & 2.07 & 23.26 & 10.58 & 55.21  \\
     \hline
     First-std & 1.60 & 22.19 & 1.44 & 3.68 & 62.00 \\
     \hline
     
     Original-dev  & 57.01 & 37.51 & 83.54 & 65.79 & -\\
     Original-std  & 56.60 & 36.63 & 83.68 & 65.68 & - \\

     \hline
    \end{tabular}}
    \centering
    \captionsetup{justification=centering}
    \caption{MLB with Attention model evaluation results.}
    
\scalebox{0.5}{
    \begin{tabular}{ c | c c c c | c} 
     Task Type &    & \multicolumn{3}{c}{Open-Ended (CIDEr)} &  \\ [0.5ex]
     \hline
     Method &    & \multicolumn{3}{c}{MUTAN with Attention} &  \\ [0.5ex]
     \hline
     Test Set&  \multicolumn{4}{c}{dev} & diff \\ [0.5ex]
     \hline
     Partition & Other & Num & Y/N & All & All \\ [0.5ex] 
     \hline
     First-dev & 0.81 & 15.35 & 1.44 & 2.64 & 63.34  \\ 
     
     Second-dev & 1.20 & 2.79 & 22.16 & 9.97 & 56.01 \\
     
     Third-dev & 1.73 & 3.46 & 26.13 & 11.93 & 54.05  \\
     
     Fourth-dev & 2.01 & 2.63 & 23.70 & 10.98 & 55.00  \\
     
     Fifth-dev & 1.27 & 2.86 & 27.83 & 12.34 & 53.64 \\
     
     Sixth-dev & 1.46 & 1.92 & 30.38 & 13.38 & 52.60  \\
     
     Seventh-dev & 1.42 & 2.33 & 23.47 & 10.57 & 55.41  \\
     \hline
     First-std & 0.82 & 15.92 & 1.33 & 2.60 & 63.17 \\
     \hline
     
     Original-dev  & 56.73 & 38.35 & 84.11 & 65.98 & -\\
     Original-std  & 56.29 & 37.47 & 84.04 & 65.77 & - \\

     \hline
    \end{tabular}}
    \centering
    \captionsetup{justification=centering}
    \caption{MUTAN with Attention model evaluation results.}
\end{subtable}%
    \hfil
\begin{subtable}[t]{0.3\linewidth}
        
\centering
\scalebox{0.5}{
    \begin{tabular}{ c | c c c c | c} 
     Task Type &    & \multicolumn{3}{c}{Open-Ended (CIDEr)} &  \\ [0.5ex]
     \hline
     Method &    & \multicolumn{3}{c}{HieCoAtt (Alt,Resnet200)} &  \\ [0.5ex]
     \hline
     Test Set&  \multicolumn{4}{c}{dev} & diff \\ [0.5ex]
     \hline
     Partition & Other & Num & Y/N & All & All \\ [0.5ex] 
     \hline
     First-dev & 1.36 & 17.21 & 0.74 & 2.82 & 58.99  \\ 
     
     Second-dev & 2.57 & 3.49 & 22.36 & 10.79 & 51.02 \\
     
     Third-dev & 1.36 & 2.74 & 7.53 & 4.04 & 57.77  \\
     
     Fourth-dev & 7.02 & 3.16 & 6.16 & 6.25 & 55.56  \\
     
     Fifth-dev & 2.62 & 4.43 & 24.71 & 11.88 & 49.93 \\
     
     Sixth-dev & 2.96 & 3.70 & 31.80 & 14.87 & 46.94  \\
     
     Seventh-dev & 2.20 & 3.13 & 18.06 & 8.81 & 53.00  \\
     \hline
     First-std & 1.44 & 17.21 & 0.78 & 2.81 & 59.25\\
     \hline

     Original-dev  & 51.77 & 38.65 & 79.70 & 61.81 & -\\
     Original-std  & 51.95 & 38.22 & 79.95 & 62.06 & - \\

     \hline
    \end{tabular}}
    \centering
    \captionsetup{justification=centering}
    \caption{HieCoAtt (Alt,Resnet200) model evaluation results.}

\scalebox{0.5}{
    \begin{tabular}{ c | c c c c | c} 
     Task Type &    & \multicolumn{3}{c}{Open-Ended (CIDEr)} &  \\ [0.5ex]
     \hline
     Method &    & \multicolumn{3}{c}{LSTM Q+I} &  \\ [0.5ex]
     \hline
     Test Set&  \multicolumn{4}{c}{dev} & diff \\ [0.5ex]
     \hline
     Partition & Other & Num & Y/N & All & All \\ [0.5ex] 
     \hline
     First-dev & 0.81 & 2.63 & 21.86 & 9.65 & 48.37  \\ 
     
     Second-dev & 1.88 & 2.61 & 26.16 & 11.92 & 46.10 \\
     
     Third-dev & 1.24 & 0.78 & 9.38 & 4.53 & 53.49  \\
     
     Fourth-dev & 1.65 & 1.67 & 17.29 & 8.07 & 49.95  \\
     
     Fifth-dev & 2.31 & 2.66 & 23.24 & 10.94 & 47.08 \\
     
     Sixth-dev & 1.22 & 2.28 & 30.87 & 13.50 & 44.52  \\
     
     Seventh-dev & 1.12 & 1.60 & 20.68 & 9.20 & 48.82  \\
     \hline
     First-std & 0.86 & 2.61 & 21.88 & 9.70 & 48.48 \\
     \hline
     
     Original-dev  & 43.40 & 36.46 & 80.87 & 58.02 & -\\
     Original-std  & 43.90 & 36.67 & 80.38 & 58.18 & - \\

     \hline
    \end{tabular}}
    \centering
    \captionsetup{justification=centering}
    \caption{LSTM Q+I model evaluation results.}
\end{subtable}
\caption{The table shows the six state-of-the-art pretrained VQA models evaluation results on the GBQD and VQA dataset. ``-'' indicates the results are not available, ``-std'' represents the accuracy of VQA model evaluated on the complete testing set of GBQD and VQA dataset and ``-dev'' indicates the accuracy of VQA model evaluated on the partial testing set of GBQD and VQA dataset. In addition, $diff = Original_{dev_{All}} - X_{dev_{All}}$, where $X$ is equal to the ``First'', ``Second'', etc.}
\label{table:table13}
\end{table*}

\begin{table*}
\renewcommand\arraystretch{1.0}
\setlength\tabcolsep{11pt}
    \centering
\begin{subtable}[t]{0.3\linewidth}
\centering
\scalebox{0.5}{
    \begin{tabular}{ c | c c c c | c} 
     Task Type &    & \multicolumn{3}{c}{Open-Ended (METEOR)} &  \\ [0.5ex]
     \hline
     Method &    & \multicolumn{3}{c}{MUTAN without Attention} &  \\ [0.5ex]
     \hline
     Test Set&  \multicolumn{4}{c}{dev} & diff \\ [0.5ex]
     \hline
     Partition & Other & Num & Y/N & All & All \\ [0.5ex] 
     \hline
     First-dev & 2.46 & 2.64 & 20.43 & 9.86 & 50.30  \\ 
     
     Second-dev & 2.55 & 2.56 & 18.70 & 9.18 & 50.98 \\
     
     Third-dev & 2.39 & 2.56 & 18.90 & 9.18 & 50.98  \\
     
     Fourth-dev & 2.18 & 2.43 & 20.38 & 9.68 & 50.48  \\
     
     Fifth-dev & 2.19 & 2.54 & 21.77 & 10.26 & 49.90 \\
     
     Sixth-dev & 2.05 & 2.68 & 22.66 & 10.58 & 49.58  \\
     
     Seventh-dev & 1.99 & 2.76 & 23.18 & 10.77 & 49.39  \\
     \hline
     First-std & 2.37 & 2.48 & 20.11 & 9.69 & 50.76 \\
     \hline
     
     Original-dev  & 47.16 & 37.32 & 81.45 & 60.16 & -\\
     Original-std  & 47.57 & 36.75 & 81.56 & 60.45 & - \\

     \hline
    \end{tabular}}
    \centering
    \captionsetup{justification=centering}
    \caption{MUTAN without Attention model evaluation results.}

\scalebox{0.5}{
    \begin{tabular}{ c | c c c c | c} 
     Task Type &    & \multicolumn{3}{c}{Open-Ended (METEOR)} &  \\ [0.5ex]
     \hline
     Method &    & \multicolumn{3}{c}{HieCoAtt (Alt,VGG19)} &  \\ [0.5ex]
     \hline
     Test Set&  \multicolumn{4}{c}{dev} & diff \\ [0.5ex]
     \hline
     Partition & Other & Num & Y/N & All & All \\ [0.5ex] 
     \hline
     First-dev & 2.67 & 3.09 & 20.98 & 10.23 & 50.25  \\ 
     
     Second-dev & 2.91 & 2.88 & 19.80 & 9.84 & 50.64 \\
     
     Third-dev & 2.83 & 3.16 & 20.14 & 9.97 & 50.51  \\
     
     Fourth-dev & 2.68 & 3.31 & 21.24 & 10.37 & 50.11  \\
     
     Fifth-dev & 2.60 & 3.13 & 22.05 & 10.64 & 49.84 \\
     
     Sixth-dev & 2.54 & 3.23 & 22.88 & 10.96 & 49.52  \\
     
     Seventh-dev & 2.56 & 3.23 & 22.94 & 11.00 & 49.48  \\
     \hline
     First-std & 2.65 & 2.89 & 20.92 & 10.21 & 50.11 \\
     \hline

     Original-dev  & 49.14 & 38.35 & 79.63 & 60.48 & -\\
     Original-std  & 49.15 & 36.52 & 79.45 & 60.32 & - \\

     \hline
    \end{tabular}}
    \centering
    \captionsetup{justification=centering}
    \caption{HieCoAtt (Alt,VGG19) model evaluation results.}
\end{subtable}%
    \hfil
\begin{subtable}[t]{0.3\linewidth}

\centering
\scalebox{0.5}{
    \begin{tabular}{ c | c c c c | c} 
     Task Type &    & \multicolumn{3}{c}{Open-Ended (METEOR)} &  \\ [0.5ex]
     \hline
     Method &    & \multicolumn{3}{c}{MLB with Attention} &  \\ [0.5ex]
     \hline
     Test Set&  \multicolumn{4}{c}{dev} & diff \\ [0.5ex]
     \hline
     Partition & Other & Num & Y/N & All & All \\ [0.5ex] 
     \hline
     First-dev & 2.33 & 2.55 & 22.96 & 10.82 & 54.97  \\ 
     
     Second-dev & 2.54 & 2.37 & 22.60 & 10.76 & 55.03 \\
     
     Third-dev & 2.51 & 2.03 & 22.17 & 10.53 & 55.26  \\
     
     Fourth-dev & 2.33 & 2.29 & 22.86 & 10.75 & 55.04  \\
     
     Fifth-dev & 2.33 & 2.24 & 23.27 & 10.91 & 54.88 \\
     
     Sixth-dev & 2.36 & 2.15 & 24.65 & 11.48 & 54.31  \\
     
     Seventh-dev & 2.29 & 2.12 & 24.33 & 11.32 & 54.47  \\
     \hline
     First-std & 2.29 & 2.17 & 23.14 & 10.87 & 54.81 \\
     \hline
     
     Original-dev  & 57.01 & 37.51 & 83.54 & 65.79 & -\\
     Original-std  & 56.60 & 36.63 & 83.68 & 65.68 & - \\

     \hline
    \end{tabular}}
    \centering
    \captionsetup{justification=centering}
    \caption{MLB with Attention model evaluation results.}
    
\scalebox{0.5}{
    \begin{tabular}{ c | c c c c | c} 
     Task Type &    & \multicolumn{3}{c}{Open-Ended (METEOR)} &  \\ [0.5ex]
     \hline
     Method &    & \multicolumn{3}{c}{MUTAN with Attention} &  \\ [0.5ex]
     \hline
     Test Set&  \multicolumn{4}{c}{dev} & diff \\ [0.5ex]
     \hline
     Partition & Other & Num & Y/N & All & All \\ [0.5ex] 
     \hline
     First-dev & 1.87 & 2.19 & 24.18 & 11.06 & 54.92  \\ 
     
     Second-dev & 1.89 & 2.18 & 23.81 & 10.92 & 55.06 \\
     
     Third-dev & 1.88 & 2.11 & 23.48 & 10.77 & 55.21  \\
     
     Fourth-dev & 1.82 & 2.20 & 23.94 & 10.94 & 55.04  \\
     
     Fifth-dev & 1.67 & 2.28 & 24.18 & 10.98 & 55.00 \\
     
     Sixth-dev & 1.74 & 2.20 & 24.77 & 11.24 & 54.74  \\
     
     Seventh-dev & 1.69 & 2.30 & 25.05 & 11.34 & 54.64  \\
     \hline
     First-std & 1.83 & 2.02 & 24.28 & 11.10 & 54.67 \\
     \hline
     
     Original-dev  & 56.73 & 38.35 & 84.11 & 65.98 & -\\
     Original-std  & 56.29 & 37.47 & 84.04 & 65.77 & - \\

     \hline
    \end{tabular}}
    \centering
    \captionsetup{justification=centering}
    \caption{MUTAN with Attention model evaluation results.}
\end{subtable}%
    \hfil
\begin{subtable}[t]{0.3\linewidth}
        
\centering
\scalebox{0.5}{
    \begin{tabular}{ c | c c c c | c} 
     Task Type &    & \multicolumn{3}{c}{Open-Ended (METEOR)} &  \\ [0.5ex]
     \hline
     Method &    & \multicolumn{3}{c}{HieCoAtt (Alt,Resnet200)} &  \\ [0.5ex]
     \hline
     Test Set&  \multicolumn{4}{c}{dev} & diff \\ [0.5ex]
     \hline
     Partition & Other & Num & Y/N & All & All \\ [0.5ex] 
     \hline
     First-dev & 3.34 & 3.38 & 21.92 & 10.97 & 50.84  \\ 
     
     Second-dev & 3.21 & 3.17 & 21.55 & 10.73 & 51.08 \\
     
     Third-dev & 3.29 & 3.51 & 21.44 & 10.76 & 51.05  \\
     
     Fourth-dev & 3.20 & 3.41 & 22.21 & 11.02 & 50.79  \\
     
     Fifth-dev & 3.24 & 3.36 & 22.64 & 11.21 & 50.60 \\
     
     Sixth-dev & 3.05 & 3.34 & 23.26 & 11.38 & 50.43  \\
     
     Seventh-dev & 3.12 & 3.53 & 23.52 & 11.54 & 50.27  \\
     \hline
     First-std & 3.32 & 3.18 & 21.78 & 10.92 & 51.14\\
     \hline

     Original-dev  & 51.77 & 38.65 & 79.70 & 61.81 & -\\
     Original-std  & 51.95 & 38.22 & 79.95 & 62.06 & - \\

     \hline
    \end{tabular}}
    \centering
    \captionsetup{justification=centering}
    \caption{HieCoAtt (Alt,Resnet200) model evaluation results.}

\scalebox{0.5}{
    \begin{tabular}{ c | c c c c | c} 
     Task Type &    & \multicolumn{3}{c}{Open-Ended (METEOR)} &  \\ [0.5ex]
     \hline
     Method &    & \multicolumn{3}{c}{LSTM Q+I} &  \\ [0.5ex]
     \hline
     Test Set&  \multicolumn{4}{c}{dev} & diff \\ [0.5ex]
     \hline
     Partition & Other & Num & Y/N & All & All \\ [0.5ex] 
     \hline
     First-dev & 2.15 & 3.26 & 21.91 & 10.38 & 47.64  \\ 
     
     Second-dev & 2.04 & 3.24 & 21.91 & 10.32 & 47.70 \\
     
     Third-dev & 2.08 & 2.68 & 21.67 & 10.19 & 47.83  \\
     
     Fourth-dev & 2.12 & 2.62 & 22.57 & 10.57 & 47.45  \\
     
     Fifth-dev & 1.94 & 2.86 & 23.67 & 10.96 & 47.06 \\
     
     Sixth-dev & 1.96 & 2.69 & 24.49 & 11.28 & 46.74  \\
     
     Seventh-dev & 1.94 & 2.61 & 24.12 & 11.12 & 46.90  \\
     \hline
     First-std & 2.00 & 2.76 & 22.00 & 10.32 & 47.86 \\
     \hline
     
     Original-dev  & 43.40 & 36.46 & 80.87 & 58.02 & -\\
     Original-std  & 43.90 & 36.67 & 80.38 & 58.18 & - \\

     \hline
    \end{tabular}}
    \centering
    \captionsetup{justification=centering}
    \caption{LSTM Q+I model evaluation results.}
\end{subtable}
\caption{The table shows the six state-of-the-art pretrained VQA models evaluation results on the GBQD and VQA dataset. ``-'' indicates the results are not available, ``-std'' represents the accuracy of VQA model evaluated on the complete testing set of GBQD and VQA dataset and ``-dev'' indicates the accuracy of VQA model evaluated on the partial testing set of GBQD and VQA dataset. In addition, $diff = Original_{dev_{All}} - X_{dev_{All}}$, where $X$ is equal to the ``First'', ``Second'', etc.}
\label{table:table14}
\end{table*}

\begin{table*}
\renewcommand\arraystretch{1.0}
\setlength\tabcolsep{11pt}
    \centering
\begin{subtable}[t]{0.3\linewidth}
\centering
\scalebox{0.5}{
    \begin{tabular}{ c | c c c c | c} 
     Task Type &    & \multicolumn{3}{c}{Open-Ended (BLEU-1)} &  \\ [0.5ex]
     \hline
     Method &    & \multicolumn{3}{c}{MUTAN without Attention} &  \\ [0.5ex]
     \hline
     Test Set&  \multicolumn{4}{c}{dev} & diff \\ [0.5ex]
     \hline
     Partition & Other & Num & Y/N & All & All \\ [0.5ex] 
     \hline
     First-dev & 1.37 & 2.63 & 33.31 & 14.62 & 45.54  \\ 
     
     Second-dev & 1.61 & 2.71 & 28.93 & 12.94 & 47.22 \\
     
     Third-dev & 1.56 & 2.91 & 29.32 & 13.10 & 47.06  \\
     
     Fourth-dev & 1.51 & 2.76 & 29.03 & 12.94 & 47.22  \\
     
     Fifth-dev & 1.62 & 2.85 & 29.11 & 13.04 & 47.12 \\
     
     Sixth-dev & 1.63 & 2.69 & 29.16 & 13.04 & 47.12  \\
     
     Seventh-dev & 1.57 & 2.81 & 28.93 & 12.93 & 47.23  \\
     \hline
     First-std & 1.48 & 2.60 & 33.10 & 14.63 & 45.82 \\
     \hline
     
     Original-dev  & 47.16 & 37.32 & 81.45 & 60.16 & -\\
     Original-std  & 47.57 & 36.75 & 81.56 & 60.45 & - \\

     \hline
    \end{tabular}}
    \centering
    \captionsetup{justification=centering}
    \caption{MUTAN without Attention model evaluation results.}

\scalebox{0.5}{
    \begin{tabular}{ c | c c c c | c} 
     Task Type &    & \multicolumn{3}{c}{Open-Ended (BLEU-1)} &  \\ [0.5ex]
     \hline
     Method &    & \multicolumn{3}{c}{HieCoAtt (Alt,VGG19)} &  \\ [0.5ex]
     \hline
     Test Set&  \multicolumn{4}{c}{dev} & diff \\ [0.5ex]
     \hline
     Partition & Other & Num & Y/N & All & All \\ [0.5ex] 
     \hline
     First-dev & 2.28 & 2.62 & 27.61 & 12.71 & 47.77  \\ 
     
     Second-dev & 2.33 & 2.68 & 28.23 & 13.00 & 47.48 \\
     
     Third-dev & 2.25 & 2.76 & 28.43 & 13.05 & 47.43  \\
     
     Fourth-dev & 2.23 & 2.84 & 28.46 & 13.06 & 47.42  \\
     
     Fifth-dev & 2.22 & 2.47 & 28.48 & 13.03 & 47.45 \\
     
     Sixth-dev & 2.23 & 2.65 & 28.37 & 13.00 & 47.48  \\
     
     Seventh-dev & 2.24 & 2.68 & 28.34 & 13.00 & 47.48  \\
     \hline
     First-std & 2.35 & 2.64 & 27.33 & 12.68 & 47.64 \\
     \hline

     Original-dev  & 49.14 & 38.35 & 79.63 & 60.48 & -\\
     Original-std  & 49.15 & 36.52 & 79.45 & 60.32 & - \\

     \hline
    \end{tabular}}
    \centering
    \captionsetup{justification=centering}
    \caption{HieCoAtt (Alt,VGG19) model evaluation results.}
\end{subtable}%
    \hfil
\begin{subtable}[t]{0.3\linewidth}

\centering
\scalebox{0.5}{
    \begin{tabular}{ c | c c c c | c} 
     Task Type &    & \multicolumn{3}{c}{Open-Ended (BLEU-1)} &  \\ [0.5ex]
     \hline
     Method &    & \multicolumn{3}{c}{MLB with Attention} &  \\ [0.5ex]
     \hline
     Test Set&  \multicolumn{4}{c}{dev} & diff \\ [0.5ex]
     \hline
     Partition & Other & Num & Y/N & All & All \\ [0.5ex] 
     \hline
     First-dev & 1.99 & 2.38 & 28.08 & 12.74 & 53.05  \\ 
     
     Second-dev & 1.90 & 2.22 & 28.31 & 12.77 & 53.02 \\
     
     Third-dev & 2.00 & 2.48 & 28.26 & 12.83 & 52.96  \\
     
     Fourth-dev & 1.97 & 2.53 & 28.00 & 12.72 & 53.07  \\
     
     Fifth-dev & 1.89 & 2.23 & 28.10 & 12.68 & 53.11 \\
     
     Sixth-dev & 1.90 & 2.24 & 28.18 & 12.72 & 53.07  \\
     
     Seventh-dev & 1.81 & 2.37 & 28.13 & 12.67 & 53.12  \\
     \hline
     First-std & 2.13 & 2.43 & 28.05 & 12.84 & 52.84 \\
     \hline
     
     Original-dev  & 57.01 & 37.51 & 83.54 & 65.79 & -\\
     Original-std  & 56.60 & 36.63 & 83.68 & 65.68 & - \\

     \hline
    \end{tabular}}
    \centering
    \captionsetup{justification=centering}
    \caption{MLB with Attention model evaluation results.}
    
\scalebox{0.5}{
    \begin{tabular}{ c | c c c c | c} 
     Task Type &    & \multicolumn{3}{c}{Open-Ended (BLEU-1)} &  \\ [0.5ex]
     \hline
     Method &    & \multicolumn{3}{c}{MUTAN with Attention} &  \\ [0.5ex]
     \hline
     Test Set&  \multicolumn{4}{c}{dev} & diff \\ [0.5ex]
     \hline
     Partition & Other & Num & Y/N & All & All \\ [0.5ex] 
     \hline
     First-dev & 1.52 & 2.32 & 29.67 & 13.16 & 52.82  \\ 
     
     Second-dev & 1.38 & 1.98 & 28.68 & 12.65 & 53.33 \\
     
     Third-dev & 1.48 & 2.03 & 28.50 & 12.63 & 53.35  \\
     
     Fourth-dev & 1.54 & 2.19 & 28.78 & 12.79 & 53.19  \\
     
     Fifth-dev & 1.49 & 2.27 & 28.52 & 12.67 & 53.31 \\
     
     Sixth-dev & 1.48 & 2.24 & 28.27 & 12.56 & 53.42  \\
     
     Seventh-dev & 1.45 & 2.04 & 28.64 & 12.67 & 53.31  \\
     \hline
     First-std & 1.62 & 2.34 & 29.50 & 13.19 & 52.58 \\
     \hline
     
     Original-dev  & 56.73 & 38.35 & 84.11 & 65.98 & -\\
     Original-std  & 56.29 & 37.47 & 84.04 & 65.77 & - \\

     \hline
    \end{tabular}}
    \centering
    \captionsetup{justification=centering}
    \caption{MUTAN with Attention model evaluation results.}
\end{subtable}%
    \hfil
\begin{subtable}[t]{0.3\linewidth}
        
\centering
\scalebox{0.5}{
    \begin{tabular}{ c | c c c c | c} 
     Task Type &    & \multicolumn{3}{c}{Open-Ended (BLEU-1)} &  \\ [0.5ex]
     \hline
     Method &    & \multicolumn{3}{c}{HieCoAtt (Alt,Resnet200)} &  \\ [0.5ex]
     \hline
     Test Set&  \multicolumn{4}{c}{dev} & diff \\ [0.5ex]
     \hline
     Partition & Other & Num & Y/N & All & All \\ [0.5ex] 
     \hline
     First-dev & 2.57 & 2.93 & 28.04 & 13.06 & 48.75  \\ 
     
     Second-dev & 2.60 & 2.79 & 28.16 & 13.11 & 48.70 \\
     
     Third-dev & 2.62 & 2.83 & 28.18 & 13.13 & 48.68  \\
     
     Fourth-dev & 2.68 & 2.97 & 28.34 & 13.24 & 48.57  \\
     
     Fifth-dev & 2.72 & 2.97 & 28.19 & 13.20 & 48.61 \\
     
     Sixth-dev & 2.71 & 3.07 & 28.03 & 13.14 & 48.67  \\
     
     Seventh-dev & 2.60 & 2.92 & 27.85 & 13.00 & 48.81  \\
     \hline
     First-std & 2.82 & 3.04 & 27.74 & 13.11 & 48.95\\
     \hline

     Original-dev  & 51.77 & 38.65 & 79.70 & 61.81 & -\\
     Original-std  & 51.95 & 38.22 & 79.95 & 62.06 & - \\

     \hline
    \end{tabular}}
    \centering
    \captionsetup{justification=centering}
    \caption{HieCoAtt (Alt,Resnet200) model evaluation results.}

\scalebox{0.5}{
    \begin{tabular}{ c | c c c c | c} 
     Task Type &    & \multicolumn{3}{c}{Open-Ended (BLEU-1)} &  \\ [0.5ex]
     \hline
     Method &    & \multicolumn{3}{c}{LSTM Q+I} &  \\ [0.5ex]
     \hline
     Test Set&  \multicolumn{4}{c}{dev} & diff \\ [0.5ex]
     \hline
     Partition & Other & Num & Y/N & All & All \\ [0.5ex] 
     \hline
     First-dev & 1.51 & 2.29 & 29.27 & 12.99 & 45.03  \\ 
     
     Second-dev & 1.57 & 2.29 & 29.61 & 13.16 & 44.86 \\
     
     Third-dev & 1.63 & 2.43 & 29.74 & 13.25 & 44.77  \\
     
     Fourth-dev & 1.61 & 2.39 & 29.65 & 13.20 & 44.82  \\
     
     Fifth-dev & 1.61 & 2.22 & 29.78 & 13.23 & 44.79 \\
     
     Sixth-dev & 1.53 & 2.53 & 29.80 & 13.24 & 44.78  \\
     
     Seventh-dev & 1.53 & 2.43 & 29.63 & 13.16 & 44.86  \\
     \hline
     First-std & 1.56 & 2.44 & 28.76 & 12.86 & 45.32 \\
     \hline
     
     Original-dev  & 43.40 & 36.46 & 80.87 & 58.02 & -\\
     Original-std  & 43.90 & 36.67 & 80.38 & 58.18 & - \\

     \hline
    \end{tabular}}
    \centering
    \captionsetup{justification=centering}
    \caption{LSTM Q+I model evaluation results.}
\end{subtable}
\caption{The table shows the six state-of-the-art pretrained VQA models evaluation results on the YNBQD and VQA dataset. ``-'' indicates the results are not available, ``-std'' represents the accuracy of VQA model evaluated on the complete testing set of YNBQD and VQA dataset and ``-dev'' indicates the accuracy of VQA model evaluated on the partial testing set of YNBQD and VQA dataset. In addition, $diff = Original_{dev_{All}} - X_{dev_{All}}$, where $X$ is equal to the ``First'', ``Second'', etc.}
\label{table:table15}
\end{table*}

\begin{table*}
\renewcommand\arraystretch{1.0}
\setlength\tabcolsep{11pt}
    \centering
\begin{subtable}[t]{0.3\linewidth}
\centering
\scalebox{0.5}{
    \begin{tabular}{ c | c c c c | c} 
     Task Type &    & \multicolumn{3}{c}{Open-Ended (BLEU-2)} &  \\ [0.5ex]
     \hline
     Method &    & \multicolumn{3}{c}{MUTAN without Attention} &  \\ [0.5ex]
     \hline
     Test Set&  \multicolumn{4}{c}{dev} & diff \\ [0.5ex]
     \hline
     Partition & Other & Num & Y/N & All & All \\ [0.5ex] 
     \hline
     First-dev & 1.41 & 2.89 & 33.51 & 14.75 & 45.41  \\ 
     
     Second-dev & 1.66 & 2.86 & 29.20 & 13.10 & 47.06 \\
     
     Third-dev & 1.56 & 2.91 & 29.32 & 13.10 & 47.06  \\
     
     Fourth-dev & 1.51 & 2.76 & 29.03 & 12.94 & 47.22  \\
     
     Fifth-dev & 1.61 & 2.85 & 29.34 & 13.13 & 47.03 \\
     
     Sixth-dev & 1.63 & 2.86 & 29.20 & 13.08 & 47.08  \\
     
     Seventh-dev & 1.50 & 2.97 & 28.87 & 12.89 & 47.27  \\
     \hline
     First-std & 1.35 & 2.86 & 32.94 & 14.52 & 45.93 \\
     \hline
     
     Original-dev  & 47.16 & 37.32 & 81.45 & 60.16 & -\\
     Original-std  & 47.57 & 36.75 & 81.56 & 60.45 & - \\

     \hline
    \end{tabular}}
    \centering
    \captionsetup{justification=centering}
    \caption{MUTAN without Attention model evaluation results.}

\scalebox{0.5}{
    \begin{tabular}{ c | c c c c | c} 
     Task Type &    & \multicolumn{3}{c}{Open-Ended (BLEU-2)} &  \\ [0.5ex]
     \hline
     Method &    & \multicolumn{3}{c}{HieCoAtt (Alt,VGG19)} &  \\ [0.5ex]
     \hline
     Test Set&  \multicolumn{4}{c}{dev} & diff \\ [0.5ex]
     \hline
     Partition & Other & Num & Y/N & All & All \\ [0.5ex] 
     \hline
     First-dev & 2.23 & 3.01 & 27.25 & 12.58 & 47.90  \\ 
     
     Second-dev & 2.62 & 3.50 & 28.01 & 13.14 & 47.34 \\
     
     Third-dev & 2.27 & 3.10 & 28.33 & 13.06 & 47.42  \\
     
     Fourth-dev & 2.25 & 3.11 & 28.33 & 13.05 & 47.43  \\
     
     Fifth-dev & 2.24 & 3.05 & 28.34 & 13.04 & 47.44 \\
     
     Sixth-dev & 2.29 & 3.34 & 28.36 & 13.11 & 47.37  \\
     
     Seventh-dev & 2.24 & 3.05 & 28.58 & 13.14 & 47.34  \\
     \hline
     First-std & 2.36 & 2.29 & 27.51 & 12.78 & 47.54 \\
     \hline

     Original-dev  & 49.14 & 38.35 & 79.63 & 60.48 & -\\
     Original-std  & 49.15 & 36.52 & 79.45 & 60.32 & - \\

     \hline
    \end{tabular}}
    \centering
    \captionsetup{justification=centering}
    \caption{HieCoAtt (Alt,VGG19) model evaluation results.}
\end{subtable}%
    \hfil
\begin{subtable}[t]{0.3\linewidth}

\centering
\scalebox{0.5}{
    \begin{tabular}{ c | c c c c | c} 
     Task Type &    & \multicolumn{3}{c}{Open-Ended (BLEU-2)} &  \\ [0.5ex]
     \hline
     Method &    & \multicolumn{3}{c}{MLB with Attention} &  \\ [0.5ex]
     \hline
     Test Set&  \multicolumn{4}{c}{dev} & diff \\ [0.5ex]
     \hline
     Partition & Other & Num & Y/N & All & All \\ [0.5ex] 
     \hline
     First-dev & 1.98 & 2.64 & 28.38 & 12.89 & 52.90  \\ 
     
     Second-dev & 2.08 & 2.45 & 28.05 & 12.78 & 53.01 \\
     
     Third-dev & 2.00 & 2.48 & 28.26 & 12.83 & 52.96  \\
     
     Fourth-dev & 1.97 & 2.53 & 28.00 & 12.72 & 53.07  \\
     
     Fifth-dev & 1.92 & 2.34 & 28.64 & 12.93 & 52.86 \\
     
     Sixth-dev & 1.95 & 2.37 & 28.66 & 12.96 & 52.83  \\
     
     Seventh-dev & 1.90 & 2.44 & 28.34 & 12.81 & 52.98  \\
     \hline
     First-std & 2.06 & 2.70 & 28.50 & 13.02 & 52.66 \\
     \hline
     
     Original-dev  & 57.01 & 37.51 & 83.54 & 65.79 & -\\
     Original-std  & 56.60 & 36.63 & 83.68 & 65.68 & - \\

     \hline
    \end{tabular}}
    \centering
    \captionsetup{justification=centering}
    \caption{MLB with Attention model evaluation results.}
    
\scalebox{0.5}{
    \begin{tabular}{ c | c c c c | c} 
     Task Type &    & \multicolumn{3}{c}{Open-Ended (BLEU-2)} &  \\ [0.5ex]
     \hline
     Method &    & \multicolumn{3}{c}{MUTAN with Attention} &  \\ [0.5ex]
     \hline
     Test Set&  \multicolumn{4}{c}{dev} & diff \\ [0.5ex]
     \hline
     Partition & Other & Num & Y/N & All & All \\ [0.5ex] 
     \hline
     First-dev & 1.45 & 2.24 & 29.63 & 13.10 & 52.88  \\ 
     
     Second-dev & 1.38 & 2.22 & 28.41 & 12.57 & 53.41 \\
     
     Third-dev & 1.48 & 2.03 & 28.50 & 12.63 & 53.35  \\
     
     Fourth-dev & 1.54 & 2.19 & 28.78 & 12.79 & 53.19  \\
     
     Fifth-dev & 1.39 & 2.39 & 28.71 & 12.71 & 53.27 \\
     
     Sixth-dev & 1.47 & 2.08 & 28.65 & 12.69 & 53.29  \\
     
     Seventh-dev & 1.45 & 2.16 & 28.50 & 12.63 & 53.35  \\
     \hline
     First-std & 1.58 & 2.38 & 29.15 & 13.02 & 52.75 \\
     \hline
     
     Original-dev  & 56.73 & 38.35 & 84.11 & 65.98 & -\\
     Original-std  & 56.29 & 37.47 & 84.04 & 65.77 & - \\

     \hline
    \end{tabular}}
    \centering
    \captionsetup{justification=centering}
    \caption{MUTAN with Attention model evaluation results.}
\end{subtable}%
    \hfil
\begin{subtable}[t]{0.3\linewidth}
        
\centering
\scalebox{0.5}{
    \begin{tabular}{ c | c c c c | c} 
     Task Type &    & \multicolumn{3}{c}{Open-Ended (BLEU-2)} &  \\ [0.5ex]
     \hline
     Method &    & \multicolumn{3}{c}{HieCoAtt (Alt,Resnet200)} &  \\ [0.5ex]
     \hline
     Test Set&  \multicolumn{4}{c}{dev} & diff \\ [0.5ex]
     \hline
     Partition & Other & Num & Y/N & All & All \\ [0.5ex] 
     \hline
     First-dev & 2.62 & 3.50 & 28.01 & 13.14 & 48.67  \\ 
     
     Second-dev & 2.72 & 3.48 & 28.12 & 13.22 & 48.59 \\
     
     Third-dev & 2.79 & 3.39 & 28.17 & 13.27 & 48.54  \\
     
     Fourth-dev & 2.76 & 3.44 & 28.32 & 13.33 & 48.48  \\
     
     Fifth-dev & 2.74 & 3.48 & 28.24 & 13.28 & 48.53 \\
     
     Sixth-dev & 2.68 & 3.38 & 28.30 & 13.27 & 48.54  \\
     
     Seventh-dev & 2.79 & 3.30 & 28.46 & 13.38 & 48.43  \\
     \hline
     First-std & 2.61 & 3.20 & 28.00 & 13.13 & 48.93\\
     \hline

     Original-dev  & 51.77 & 38.65 & 79.70 & 61.81 & -\\
     Original-std  & 51.95 & 38.22 & 79.95 & 62.06 & - \\

     \hline
    \end{tabular}}
    \centering
    \captionsetup{justification=centering}
    \caption{HieCoAtt (Alt,Resnet200) model evaluation results.}

\scalebox{0.5}{
    \begin{tabular}{ c | c c c c | c} 
     Task Type &    & \multicolumn{3}{c}{Open-Ended (BLEU-2)} &  \\ [0.5ex]
     \hline
     Method &    & \multicolumn{3}{c}{LSTM Q+I} &  \\ [0.5ex]
     \hline
     Test Set&  \multicolumn{4}{c}{dev} & diff \\ [0.5ex]
     \hline
     Partition & Other & Num & Y/N & All & All \\ [0.5ex] 
     \hline
     First-dev & 1.60 & 2.65 & 28.81 & 12.88 & 45.14  \\ 
     
     Second-dev & 1.65 & 2.33 & 29.48 & 13.15 & 44.87 \\
     
     Third-dev & 1.63 & 2.54 & 29.66 & 13.23 & 44.79  \\
     
     Fourth-dev & 1.60 & 2.63 & 29.26 & 13.07 & 44.95  \\
     
     Fifth-dev & 1.61 & 2.55 & 29.96 & 13.34 & 44.68 \\
     
     Sixth-dev & 1.68 & 2.60 & 29.39 & 13.15 & 44.87  \\
     
     Seventh-dev & 1.53 & 2.44 & 29.75 & 13.21 & 44.81  \\
     \hline
     First-std & 1.55 & 2.59 & 29.22 & 13.06 & 45.12 \\
     \hline
     
     Original-dev  & 43.40 & 36.46 & 80.87 & 58.02 & -\\
     Original-std  & 43.90 & 36.67 & 80.38 & 58.18 & - \\

     \hline
    \end{tabular}}
    \centering
    \captionsetup{justification=centering}
    \caption{LSTM Q+I model evaluation results.}
\end{subtable}
\caption{The table shows the six state-of-the-art pretrained VQA models evaluation results on the YNBQD and VQA dataset. ``-'' indicates the results are not available, ``-std'' represents the accuracy of VQA model evaluated on the complete testing set of YNBQD and VQA dataset and ``-dev'' indicates the accuracy of VQA model evaluated on the partial testing set of YNBQD and VQA dataset. In addition, $diff = Original_{dev_{All}} - X_{dev_{All}}$, where $X$ is equal to the ``First'', ``Second'', etc.}
\label{table:table16}
\end{table*}

\begin{table*}
\renewcommand\arraystretch{1.0}
\setlength\tabcolsep{11pt}
    \centering
\begin{subtable}[t]{0.3\linewidth}
\centering
\scalebox{0.5}{
    \begin{tabular}{ c | c c c c | c} 
     Task Type &    & \multicolumn{3}{c}{Open-Ended (BLEU-3)} &  \\ [0.5ex]
     \hline
     Method &    & \multicolumn{3}{c}{MUTAN without Attention} &  \\ [0.5ex]
     \hline
     Test Set&  \multicolumn{4}{c}{dev} & diff \\ [0.5ex]
     \hline
     Partition & Other & Num & Y/N & All & All \\ [0.5ex] 
     \hline
     First-dev & 1.33 & 2.69 & 32.75 & 14.37 & 45.79  \\ 
     
     Second-dev & 1.58 & 2.67 & 28.73 & 12.84 & 47.32 \\
     
     Third-dev & 1.56 & 2.82 & 28.35 & 12.69 & 47.47  \\
     
     Fourth-dev & 1.58 & 2.83 & 28.47 & 12.75 & 47.41  \\
     
     Fifth-dev & 1.57 & 2.62 & 28.57 & 12.77 & 47.39 \\
     
     Sixth-dev & 1.53 & 2.51 & 28.61 & 12.75 & 47.41  \\
     
     Seventh-dev & 1.57 & 2.68 & 28.03 & 12.55 & 47.61  \\
     \hline
     First-std & 1.22 & 2.81 & 32.85 & 14.42 & 46.03 \\
     \hline
     
     Original-dev  & 47.16 & 37.32 & 81.45 & 60.16 & -\\
     Original-std  & 47.57 & 36.75 & 81.56 & 60.45 & - \\

     \hline
    \end{tabular}}
    \centering
    \captionsetup{justification=centering}
    \caption{MUTAN without Attention model evaluation results.}

\scalebox{0.5}{
    \begin{tabular}{ c | c c c c | c} 
     Task Type &    & \multicolumn{3}{c}{Open-Ended (BLEU-3)} &  \\ [0.5ex]
     \hline
     Method &    & \multicolumn{3}{c}{HieCoAtt (Alt,VGG19)} &  \\ [0.5ex]
     \hline
     Test Set&  \multicolumn{4}{c}{dev} & diff \\ [0.5ex]
     \hline
     Partition & Other & Num & Y/N & All & All \\ [0.5ex] 
     \hline
     First-dev & 2.45 & 3.03 & 27.34 & 12.73 & 47.75  \\ 
     
     Second-dev & 2.29 & 3.06 & 27.91 & 12.89 & 47.59 \\
     
     Third-dev & 2.33 & 2.90 & 27.97 & 12.91 & 47.57  \\
     
     Fourth-dev & 2.30 & 2.97 & 28.38 & 13.08 & 47.40  \\
     
     Fifth-dev & 2.34 & 2.79 & 27.81 & 12.84 & 47.64 \\
     
     Sixth-dev & 2.27 & 3.01 & 28.13 & 12.96 & 47.52  \\
     
     Seventh-dev & 2.24 & 2.90 & 27.84 & 12.82 & 47.66  \\
     \hline
     First-std & 2.31 & 2.97 & 26.98 & 12.55 & 47.77 \\
     \hline

     Original-dev  & 49.14 & 38.35 & 79.63 & 60.48 & -\\
     Original-std  & 49.15 & 36.52 & 79.45 & 60.32 & - \\

     \hline
    \end{tabular}}
    \centering
    \captionsetup{justification=centering}
    \caption{HieCoAtt (Alt,VGG19) model evaluation results.}
\end{subtable}%
    \hfil
\begin{subtable}[t]{0.3\linewidth}

\centering
\scalebox{0.5}{
    \begin{tabular}{ c | c c c c | c} 
     Task Type &    & \multicolumn{3}{c}{Open-Ended (BLEU-3)} &  \\ [0.5ex]
     \hline
     Method &    & \multicolumn{3}{c}{MLB with Attention} &  \\ [0.5ex]
     \hline
     Test Set&  \multicolumn{4}{c}{dev} & diff \\ [0.5ex]
     \hline
     Partition & Other & Num & Y/N & All & All \\ [0.5ex] 
     \hline
     First-dev & 2.05 & 2.59 & 27.65 & 12.62 & 53.17  \\ 
     
     Second-dev & 2.02 & 2.52 & 27.77 & 12.64 & 53.15 \\
     
     Third-dev & 1.92 & 2.37 & 28.01 & 12.67 & 53.12  \\
     
     Fourth-dev & 1.94 & 2.58 & 27.70 & 12.58 & 53.21  \\
     
     Fifth-dev & 1.85 & 2.51 & 27.94 & 12.63 & 53.16 \\
     
     Sixth-dev & 1.89 & 2.54 & 28.14 & 12.74 & 53.05  \\
     
     Seventh-dev & 1.94 & 2.18 & 27.58 & 12.49 & 53.30  \\
     \hline
     First-std & 1.91 & 2.78 & 28.31 & 12.88 & 52.80 \\
     \hline
     
     Original-dev  & 57.01 & 37.51 & 83.54 & 65.79 & -\\
     Original-std  & 56.60 & 36.63 & 83.68 & 65.68 & - \\

     \hline
    \end{tabular}}
    \centering
    \captionsetup{justification=centering}
    \caption{MLB with Attention model evaluation results.}
    
\scalebox{0.5}{
    \begin{tabular}{ c | c c c c | c} 
     Task Type &    & \multicolumn{3}{c}{Open-Ended (BLEU-3)} &  \\ [0.5ex]
     \hline
     Method &    & \multicolumn{3}{c}{MUTAN with Attention} &  \\ [0.5ex]
     \hline
     Test Set&  \multicolumn{4}{c}{dev} & diff \\ [0.5ex]
     \hline
     Partition & Other & Num & Y/N & All & All \\ [0.5ex] 
     \hline
     First-dev & 1.50 & 2.31 & 28.94 & 12.85 & 53.13  \\ 
     
     Second-dev & 1.41 & 2.37 & 28.03 & 12.44 & 53.54 \\
     
     Third-dev & 1.47 & 2.26 & 27.96 & 12.43 & 53.55  \\
     
     Fourth-dev & 1.45 & 1.91 & 28.04 & 12.42 & 53.56  \\
     
     Fifth-dev & 1.46 & 2.33 & 28.45 & 12.63 & 53.35 \\
     
     Sixth-dev & 1.47 & 2.12 & 28.25 & 12.53 & 53.45  \\
     
     Seventh-dev & 1.41 & 1.95 & 27.83 & 12.31 & 53.67  \\
     \hline
     First-std & 1.47 & 2.44 & 29.26 & 13.02 & 52.75 \\
     \hline
     
     Original-dev  & 56.73 & 38.35 & 84.11 & 65.98 & -\\
     Original-std  & 56.29 & 37.47 & 84.04 & 65.77 & - \\

     \hline
    \end{tabular}}
    \centering
    \captionsetup{justification=centering}
    \caption{MUTAN with Attention model evaluation results.}
\end{subtable}%
    \hfil
\begin{subtable}[t]{0.3\linewidth}
        
\centering
\scalebox{0.5}{
    \begin{tabular}{ c | c c c c | c} 
     Task Type &    & \multicolumn{3}{c}{Open-Ended (BLEU-3)} &  \\ [0.5ex]
     \hline
     Method &    & \multicolumn{3}{c}{HieCoAtt (Alt,Resnet200)} &  \\ [0.5ex]
     \hline
     Test Set&  \multicolumn{4}{c}{dev} & diff \\ [0.5ex]
     \hline
     Partition & Other & Num & Y/N & All & All \\ [0.5ex] 
     \hline
     First-dev & 2.55 & 3.26 & 27.80 & 12.99 & 48.82  \\ 
     
     Second-dev & 2.56 & 3.20 & 28.05 & 13.09 & 48.72 \\
     
     Third-dev & 2.79 & 3.08 & 27.84 & 13.10 & 48.71  \\
     
     Fourth-dev & 2.71 & 3.24 & 28.31 & 13.27 & 48.54  \\
     
     Fifth-dev & 2.69 & 3.08 & 27.62 & 12.96 & 48.85 \\
     
     Sixth-dev & 2.83 & 3.12 & 28.02 & 13.20 & 48.61  \\
     
     Seventh-dev & 2.68 & 3.06 & 27.62 & 12.96 & 48.85  \\
     \hline
     First-std & 2.45 & 3.14 & 27.61 & 12.89 & 49.17\\
     \hline

     Original-dev  & 51.77 & 38.65 & 79.70 & 61.81 & -\\
     Original-std  & 51.95 & 38.22 & 79.95 & 62.06 & - \\

     \hline
    \end{tabular}}
    \centering
    \captionsetup{justification=centering}
    \caption{HieCoAtt (Alt,Resnet200) model evaluation results.}

\scalebox{0.5}{
    \begin{tabular}{ c | c c c c | c} 
     Task Type &    & \multicolumn{3}{c}{Open-Ended (BLEU-3)} &  \\ [0.5ex]
     \hline
     Method &    & \multicolumn{3}{c}{LSTM Q+I} &  \\ [0.5ex]
     \hline
     Test Set&  \multicolumn{4}{c}{dev} & diff \\ [0.5ex]
     \hline
     Partition & Other & Num & Y/N & All & All \\ [0.5ex] 
     \hline
     First-dev & 1.46 & 2.66 & 28.97 & 12.88 & 45.14  \\ 
     
     Second-dev & 1.54 & 2.72 & 29.35 & 13.08 & 44.94 \\
     
     Third-dev & 1.57 & 2.91 & 29.73 & 13.27 & 44.75  \\
     
     Fourth-dev & 1.54 & 2.68 & 29.34 & 13.07 & 44.95  \\
     
     Fifth-dev & 1.46 & 2.70 & 29.88 & 13.26 & 44.76 \\
     
     Sixth-dev & 1.48 & 2.67 & 29.58 & 13.14 & 44.88  \\
     
     Seventh-dev & 1.56 & 2.47 & 29.26 & 13.03 & 44.99  \\
     \hline
     First-std & 1.49 & 2.61 & 29.11 & 12.99 & 45.19 \\
     \hline
     
     Original-dev  & 43.40 & 36.46 & 80.87 & 58.02 & -\\
     Original-std  & 43.90 & 36.67 & 80.38 & 58.18 & - \\

     \hline
    \end{tabular}}
    \centering
    \captionsetup{justification=centering}
    \caption{LSTM Q+I model evaluation results.}
\end{subtable}
\caption{The table shows the six state-of-the-art pretrained VQA models evaluation results on the YNBQD and VQA dataset. ``-'' indicates the results are not available, ``-std'' represents the accuracy of VQA model evaluated on the complete testing set of YNBQD and VQA dataset and ``-dev'' indicates the accuracy of VQA model evaluated on the partial testing set of YNBQD and VQA dataset. In addition, $diff = Original_{dev_{All}} - X_{dev_{All}}$, where $X$ is equal to the ``First'', ``Second'', etc.}
\label{table:table17}
\end{table*}

\begin{table*}
\renewcommand\arraystretch{1.0}
\setlength\tabcolsep{11pt}
    \centering
\begin{subtable}[t]{0.3\linewidth}
\centering
\scalebox{0.5}{
    \begin{tabular}{ c | c c c c | c} 
     Task Type &    & \multicolumn{3}{c}{Open-Ended (BLEU-4)} &  \\ [0.5ex]
     \hline
     Method &    & \multicolumn{3}{c}{MUTAN without Attention} &  \\ [0.5ex]
     \hline
     Test Set&  \multicolumn{4}{c}{dev} & diff \\ [0.5ex]
     \hline
     Partition & Other & Num & Y/N & All & All \\ [0.5ex] 
     \hline
     First-dev & 1.31 & 2.63 & 33.27 & 14.57 & 45.59  \\ 
     
     Second-dev & 1.58 & 2.63 & 29.26 & 13.06 & 47.10 \\
     
     Third-dev & 1.53 & 2.68 & 28.93 & 12.90 & 47.26  \\
     
     Fourth-dev & 1.58 & 2.59 & 28.92 & 12.91 & 47.25  \\
     
     Fifth-dev & 1.51 & 2.69 & 29.28 & 13.03 & 47.13 \\
     
     Sixth-dev & 1.59 & 2.47 & 29.40 & 13.10 & 47.06  \\
     
     Seventh-dev & 1.54 & 2.53 & 28.56 & 12.74 & 47.42  \\
     \hline
     First-std & 1.41 & 2.67 & 33.06 & 14.58 & 45.87 \\
     \hline
     
     Original-dev  & 47.16 & 37.32 & 81.45 & 60.16 & -\\
     Original-std  & 47.57 & 36.75 & 81.56 & 60.45 & - \\

     \hline
    \end{tabular}}
    \centering
    \captionsetup{justification=centering}
    \caption{MUTAN without Attention model evaluation results.}

\scalebox{0.5}{
    \begin{tabular}{ c | c c c c | c} 
     Task Type &    & \multicolumn{3}{c}{Open-Ended (BLEU-4)} &  \\ [0.5ex]
     \hline
     Method &    & \multicolumn{3}{c}{HieCoAtt (Alt,VGG19)} &  \\ [0.5ex]
     \hline
     Test Set&  \multicolumn{4}{c}{dev} & diff \\ [0.5ex]
     \hline
     Partition & Other & Num & Y/N & All & All \\ [0.5ex] 
     \hline
     First-dev & 2.19 & 2.64 & 27.47 & 12.61 & 47.87  \\ 
     
     Second-dev & 2.17 & 2.78 & 28.07 & 12.86 & 47.62 \\
     
     Third-dev & 2.17 & 2.73 & 28.46 & 13.02 & 47.46  \\
     
     Fourth-dev & 2.17 & 2.79 & 28.29 & 12.95 & 47.53  \\
     
     Fifth-dev & 2.23 & 2.72 & 28.06 & 12.88 & 47.60 \\
     
     Sixth-dev & 2.24 & 3.03 & 28.44 & 13.07 & 47.41  \\
     
     Seventh-dev & 2.06 & 2.60 & 28.53 & 12.98 & 47.50  \\
     \hline
     First-std & 2.29 & 2.85 & 27.52 & 12.74 & 47.48 \\
     \hline

     Original-dev  & 49.14 & 38.35 & 79.63 & 60.48 & -\\
     Original-std  & 49.15 & 36.52 & 79.45 & 60.32 & - \\

     \hline
    \end{tabular}}
    \centering
    \captionsetup{justification=centering}
    \caption{HieCoAtt (Alt,VGG19) model evaluation results.}
\end{subtable}%
    \hfil
\begin{subtable}[t]{0.3\linewidth}

\centering
\scalebox{0.5}{
    \begin{tabular}{ c | c c c c | c} 
     Task Type &    & \multicolumn{3}{c}{Open-Ended (BLEU-4)} &  \\ [0.5ex]
     \hline
     Method &    & \multicolumn{3}{c}{MLB with Attention} &  \\ [0.5ex]
     \hline
     Test Set&  \multicolumn{4}{c}{dev} & diff \\ [0.5ex]
     \hline
     Partition & Other & Num & Y/N & All & All \\ [0.5ex] 
     \hline
     First-dev & 1.92 & 2.36 & 28.46 & 12.86 & 52.93  \\ 
     
     Second-dev & 1.88 & 2.02 & 28.06 & 12.64 & 53.15 \\
     
     Third-dev & 1.92 & 2.26 & 28.50 & 12.87 & 52.92  \\
     
     Fourth-dev & 1.84 & 2.29 & 27.93 & 12.60 & 53.19  \\
     
     Fifth-dev & 1.86 & 2.13 & 28.34 & 12.76 & 53.03 \\
     
     Sixth-dev & 1.86 & 2.29 & 28.49 & 12.84 & 52.95  \\
     
     Seventh-dev & 1.83 & 2.25 & 28.17 & 12.68 & 53.11  \\
     \hline
     First-std & 2.01 & 2.62 & 28.09 & 12.82 & 52.86 \\
     \hline
     
     Original-dev  & 57.01 & 37.51 & 83.54 & 65.79 & -\\
     Original-std  & 56.60 & 36.63 & 83.68 & 65.68 & - \\

     \hline
    \end{tabular}}
    \centering
    \captionsetup{justification=centering}
    \caption{MLB with Attention model evaluation results.}
    
\scalebox{0.5}{
    \begin{tabular}{ c | c c c c | c} 
     Task Type &    & \multicolumn{3}{c}{Open-Ended (BLEU-4)} &  \\ [0.5ex]
     \hline
     Method &    & \multicolumn{3}{c}{MUTAN with Attention} &  \\ [0.5ex]
     \hline
     Test Set&  \multicolumn{4}{c}{dev} & diff \\ [0.5ex]
     \hline
     Partition & Other & Num & Y/N & All & All \\ [0.5ex] 
     \hline
     First-dev & 1.42 & 1.96 & 29.38 & 12.95 & 53.03  \\ 
     
     Second-dev & 1.38 & 1.97 & 28.35 & 12.51 & 53.47 \\
     
     Third-dev & 1.35 & 1.66 & 28.78 & 12.64 & 53.34  \\
     
     Fourth-dev & 1.33 & 2.12 & 28.57 & 12.60 & 53.38  \\
     
     Fifth-dev & 1.32 & 1.90 & 28.72 & 12.63 & 53.35 \\
     
     Sixth-dev & 1.43 & 1.76 & 28.46 & 12.56 & 53.42  \\
     
     Seventh-dev & 1.38 & 1.87 & 28.54 & 12.58 & 53.40  \\
     \hline
     First-std & 1.53 & 2.26 & 29.29 & 13.05 & 52.72 \\
     \hline
     
     Original-dev  & 56.73 & 38.35 & 84.11 & 65.98 & -\\
     Original-std  & 56.29 & 37.47 & 84.04 & 65.77 & - \\

     \hline
    \end{tabular}}
    \centering
    \captionsetup{justification=centering}
    \caption{MUTAN with Attention model evaluation results.}
\end{subtable}%
    \hfil
\begin{subtable}[t]{0.3\linewidth}
        
\centering
\scalebox{0.5}{
    \begin{tabular}{ c | c c c c | c} 
     Task Type &    & \multicolumn{3}{c}{Open-Ended (BLEU-4)} &  \\ [0.5ex]
     \hline
     Method &    & \multicolumn{3}{c}{HieCoAtt (Alt,Resnet200)} &  \\ [0.5ex]
     \hline
     Test Set&  \multicolumn{4}{c}{dev} & diff \\ [0.5ex]
     \hline
     Partition & Other & Num & Y/N & All & All \\ [0.5ex] 
     \hline
     First-dev & 2.41 & 3.19 & 28.14 & 13.06 & 48.75  \\ 
     
     Second-dev & 2.46 & 2.92 & 28.18 & 13.06 & 48.75 \\
     
     Third-dev & 2.62 & 2.89 & 28.29 & 13.18 & 48.63  \\
     
     Fourth-dev & 2.58 & 3.15 & 28.36 & 13.23 & 48.58  \\
     
     Fifth-dev & 2.55 & 2.97 & 27.86 & 12.98 & 48.83 \\
     
     Sixth-dev & 2.60 & 3.12 & 28.14 & 13.14 & 48.67  \\
     
     Seventh-dev & 2.60 & 2.87 & 28.18 & 13.13 & 48.68  \\
     \hline
     First-std & 2.61 & 2.90 & 28.06 & 13.13 & 48.93\\
     \hline

     Original-dev  & 51.77 & 38.65 & 79.70 & 61.81 & -\\
     Original-std  & 51.95 & 38.22 & 79.95 & 62.06 & - \\

     \hline
    \end{tabular}}
    \centering
    \captionsetup{justification=centering}
    \caption{HieCoAtt (Alt,Resnet200) model evaluation results.}

\scalebox{0.5}{
    \begin{tabular}{ c | c c c c | c} 
     Task Type &    & \multicolumn{3}{c}{Open-Ended (BLEU-4)} &  \\ [0.5ex]
     \hline
     Method &    & \multicolumn{3}{c}{LSTM Q+I} &  \\ [0.5ex]
     \hline
     Test Set&  \multicolumn{4}{c}{dev} & diff \\ [0.5ex]
     \hline
     Partition & Other & Num & Y/N & All & All \\ [0.5ex] 
     \hline
     First-dev & 1.59 & 2.17 & 28.94 & 12.88 & 45.14  \\ 
     
     Second-dev & 1.57 & 2.49 & 29.33 & 13.06 & 44.96 \\
     
     Third-dev & 1.54 & 2.40 & 29.56 & 13.13 & 44.89  \\
     
     Fourth-dev & 1.56 & 2.43 & 29.20 & 13.00 & 45.02  \\
     
     Fifth-dev & 1.51 & 2.44 & 29.41 & 13.06 & 44.96 \\
     
     Sixth-dev & 1.44 & 2.30 & 29.42 & 13.02 & 45.00  \\
     
     Seventh-dev & 1.58 & 2.16 & 29.10 & 12.93 & 45.09  \\
     \hline
     First-std & 1.53 & 2.59 & 29.29 & 13.08 & 45.10 \\
     \hline
     
     Original-dev  & 43.40 & 36.46 & 80.87 & 58.02 & -\\
     Original-std  & 43.90 & 36.67 & 80.38 & 58.18 & - \\

     \hline
    \end{tabular}}
    \centering
    \captionsetup{justification=centering}
    \caption{LSTM Q+I model evaluation results.}
\end{subtable}
\caption{The table shows the six state-of-the-art pretrained VQA models evaluation results on the YNBQD and VQA dataset. ``-'' indicates the results are not available, ``-std'' represents the accuracy of VQA model evaluated on the complete testing set of YNBQD and VQA dataset and ``-dev'' indicates the accuracy of VQA model evaluated on the partial testing set of YNBQD and VQA dataset. In addition, $diff = Original_{dev_{All}} - X_{dev_{All}}$, where $X$ is equal to the ``First'', ``Second'', etc.}
\label{table:table18}
\end{table*}

\begin{table*}
\renewcommand\arraystretch{1.0}
\setlength\tabcolsep{11pt}
    \centering
\begin{subtable}[t]{0.3\linewidth}
\centering
\scalebox{0.5}{
    \begin{tabular}{ c | c c c c | c} 
     Task Type &    & \multicolumn{3}{c}{Open-Ended (ROUGE)} &  \\ [0.5ex]
     \hline
     Method &    & \multicolumn{3}{c}{MUTAN without Attention} &  \\ [0.5ex]
     \hline
     Test Set&  \multicolumn{4}{c}{dev} & diff \\ [0.5ex]
     \hline
     Partition & Other & Num & Y/N & All & All \\ [0.5ex] 
     \hline
     First-dev & 1.55 & 2.35 & 29.92 & 13.28 & 46.88  \\ 
     
     Second-dev & 1.47 & 2.67 & 28.79 & 12.81 & 47.35 \\
     
     Third-dev & 1.40 & 2.48 & 25.47 & 11.40 & 48.76  \\
     
     Fourth-dev & 1.67 & 2.59 & 26.52 & 11.97 & 48.19  \\
     
     Fifth-dev & 1.59 & 2.92 & 29.18 & 13.06 & 47.10 \\
     
     Sixth-dev & 1.97 & 2.62 & 29.74 & 13.44 & 46.72  \\
     
     Seventh-dev & 1.69 & 2.60 & 29.30 & 13.12 & 47.04  \\
     \hline
     First-std & 1.38 & 2.56 & 29.67 & 13.16 & 47.29 \\
     \hline
     
     Original-dev  & 47.16 & 37.32 & 81.45 & 60.16 & -\\
     Original-std  & 47.57 & 36.75 & 81.56 & 60.45 & - \\

     \hline
    \end{tabular}}
    \centering
    \captionsetup{justification=centering}
    \caption{MUTAN without Attention model evaluation results.}

\scalebox{0.5}{
    \begin{tabular}{ c | c c c c | c} 
     Task Type &    & \multicolumn{3}{c}{Open-Ended (ROUGE)} &  \\ [0.5ex]
     \hline
     Method &    & \multicolumn{3}{c}{HieCoAtt (Alt,VGG19)} &  \\ [0.5ex]
     \hline
     Test Set&  \multicolumn{4}{c}{dev} & diff \\ [0.5ex]
     \hline
     Partition & Other & Num & Y/N & All & All \\ [0.5ex] 
     \hline
     First-dev & 2.09 & 2.54 & 25.59 & 11.79 & 48.69  \\ 
     
     Second-dev & 2.19 & 3.17 & 28.79 & 13.21 & 47.27 \\
     
     Third-dev & 2.17 & 2.68 & 29.00 & 13.24 & 47.24  \\
     
     Fourth-dev & 2.16 & 2.69 & 26.97 & 12.40 & 48.08  \\
     
     Fifth-dev & 2.34 & 3.00 & 29.58 & 13.59 & 46.89 \\
     
     Sixth-dev & 2.39 & 2.91 & 28.80 & 13.28 & 47.20  \\
     
     Seventh-dev & 2.36 & 3.01 & 29.38 & 13.52 & 46.96  \\
     \hline
     First-std & 2.15 & 3.11 & 25.19 & 11.75 & 48.57 \\
     \hline

     Original-dev  & 49.14 & 38.35 & 79.63 & 60.48 & -\\
     Original-std  & 49.15 & 36.52 & 79.45 & 60.32 & - \\

     \hline
    \end{tabular}}
    \centering
    \captionsetup{justification=centering}
    \caption{HieCoAtt (Alt,VGG19) model evaluation results.}
\end{subtable}%
    \hfil
\begin{subtable}[t]{0.3\linewidth}

\centering
\scalebox{0.5}{
    \begin{tabular}{ c | c c c c | c} 
     Task Type &    & \multicolumn{3}{c}{Open-Ended (ROUGE)} &  \\ [0.5ex]
     \hline
     Method &    & \multicolumn{3}{c}{MLB with Attention} &  \\ [0.5ex]
     \hline
     Test Set&  \multicolumn{4}{c}{dev} & diff \\ [0.5ex]
     \hline
     Partition & Other & Num & Y/N & All & All \\ [0.5ex] 
     \hline
     First-dev & 1.89 & 2.33 & 28.44 & 12.83 & 52.96  \\ 
     
     Second-dev & 1.64 & 2.50 & 31.01 & 13.79 & 52.00 \\
     
     Third-dev & 1.83 & 2.19 & 26.32 & 11.92 & 53.87  \\
     
     Fourth-dev & 1.85 & 2.61 & 27.86 & 12.60 & 53.19  \\
     
     Fifth-dev & 1.92 & 2.37 & 29.53 & 13.30 & 52.49 \\
     
     Sixth-dev & 2.28 & 2.50 & 27.24 & 12.54 & 53.25  \\
     
     Seventh-dev & 2.01 & 2.43 & 29.77 & 13.45 & 52.34  \\
     \hline
     First-std & 1.71 & 2.37 & 28.41 & 12.78 & 52.90 \\
     \hline
     
     Original-dev  & 57.01 & 37.51 & 83.54 & 65.79 & -\\
     Original-std  & 56.60 & 36.63 & 83.68 & 65.68 & - \\

     \hline
    \end{tabular}}
    \centering
    \captionsetup{justification=centering}
    \caption{MLB with Attention model evaluation results.}
    
\scalebox{0.5}{
    \begin{tabular}{ c | c c c c | c} 
     Task Type &    & \multicolumn{3}{c}{Open-Ended (ROUGE)} &  \\ [0.5ex]
     \hline
     Method &    & \multicolumn{3}{c}{MUTAN with Attention} &  \\ [0.5ex]
     \hline
     Test Set&  \multicolumn{4}{c}{dev} & diff \\ [0.5ex]
     \hline
     Partition & Other & Num & Y/N & All & All \\ [0.5ex] 
     \hline
     First-dev & 1.65 & 1.97 & 27.33 & 12.22 & 53.76  \\ 
     
     Second-dev & 1.11 & 1.94 & 28.33 & 12.37 & 53.61 \\
     
     Third-dev & 1.23 & 1.90 & 27.34 & 12.02 & 53.96  \\
     
     Fourth-dev & 1.38 & 2.09 & 26.72 & 11.85 & 54.13  \\
     
     Fifth-dev & 1.46 & 2.14 & 25.69 & 11.48 & 54.50 \\
     
     Sixth-dev & 1.73 & 2.17 & 27.42 & 12.32 & 53.66  \\
     
     Seventh-dev & 1.47 & 1.97 & 27.27 & 12.11 & 53.87  \\
     \hline
     First-std & 1.45 & 2.17 & 27.04 & 12.07 & 53.70 \\
     \hline
     
     Original-dev  & 56.73 & 38.35 & 84.11 & 65.98 & -\\
     Original-std  & 56.29 & 37.47 & 84.04 & 65.77 & - \\

     \hline
    \end{tabular}}
    \centering
    \captionsetup{justification=centering}
    \caption{MUTAN with Attention model evaluation results.}
\end{subtable}%
    \hfil
\begin{subtable}[t]{0.3\linewidth}
        
\centering
\scalebox{0.5}{
    \begin{tabular}{ c | c c c c | c} 
     Task Type &    & \multicolumn{3}{c}{Open-Ended (ROUGE)} &  \\ [0.5ex]
     \hline
     Method &    & \multicolumn{3}{c}{HieCoAtt (Alt,Resnet200)} &  \\ [0.5ex]
     \hline
     Test Set&  \multicolumn{4}{c}{dev} & diff \\ [0.5ex]
     \hline
     Partition & Other & Num & Y/N & All & All \\ [0.5ex] 
     \hline
     First-dev & 2.72 & 2.76 & 27.45 & 12.87 & 48.94  \\ 
     
     Second-dev & 2.68 & 2.77 & 28.25 & 13.19 & 48.62 \\
     
     Third-dev & 2.74 & 2.98 & 27.84 & 13.07 & 48.74  \\
     
     Fourth-dev & 2.43 & 2.87 & 26.70 & 12.44 & 49.37  \\
     
     Fifth-dev & 2.71 & 2.83 & 29.40 & 13.68 & 48.13 \\
     
     Sixth-dev & 3.06 & 2.82 & 28.54 & 13.49 & 48.32  \\
     
     Seventh-dev & 2.69 & 3.08 & 29.06 & 13.55 & 48.26  \\
     \hline
     First-std & 2.54 & 3.10 & 27.66 & 12.95 & 49.11 \\
     \hline

     Original-dev  & 51.77 & 38.65 & 79.70 & 61.81 & -\\
     Original-std  & 51.95 & 38.22 & 79.95 & 62.06 & - \\

     \hline
    \end{tabular}}
    \centering
    \captionsetup{justification=centering}
    \caption{HieCoAtt (Alt,Resnet200) model evaluation results.}

\scalebox{0.5}{
    \begin{tabular}{ c | c c c c | c} 
     Task Type &    & \multicolumn{3}{c}{Open-Ended (ROUGE)} &  \\ [0.5ex]
     \hline
     Method &    & \multicolumn{3}{c}{LSTM Q+I} &  \\ [0.5ex]
     \hline
     Test Set&  \multicolumn{4}{c}{dev} & diff \\ [0.5ex]
     \hline
     Partition & Other & Num & Y/N & All & All \\ [0.5ex] 
     \hline
     First-dev & 1.60 & 2.35 & 28.65 & 12.78 & 45.24  \\ 
     
     Second-dev & 1.64 & 2.37 & 28.63 & 12.79 & 45.23 \\
     
     Third-dev & 1.64 & 2.51 & 26.05 & 11.75 & 46.27  \\
     
     Fourth-dev & 1.50 & 2.49 & 28.21 & 12.57 & 45.45  \\
     
     Fifth-dev & 1.41 & 2.37 & 28.74 & 12.73 & 45.29 \\
     
     Sixth-dev & 1.60 & 2.40 & 28.81 & 12.85 & 45.17  \\
     
     Seventh-dev & 1.53 & 2.57 & 28.94 & 12.89 & 45.13  \\
     \hline
     First-std & 1.51 & 2.59 & 28.37 & 12.69 & 45.49 \\
     \hline
     
     Original-dev  & 43.40 & 36.46 & 80.87 & 58.02 & -\\
     Original-std  & 43.90 & 36.67 & 80.38 & 58.18 & - \\

     \hline
    \end{tabular}}
    \centering
    \captionsetup{justification=centering}
    \caption{LSTM Q+I model evaluation results.}
\end{subtable}
\caption{The table shows the six state-of-the-art pretrained VQA models evaluation results on the YNBQD and VQA dataset. ``-'' indicates the results are not available, ``-std'' represents the accuracy of VQA model evaluated on the complete testing set of YNBQD and VQA dataset and ``-dev'' indicates the accuracy of VQA model evaluated on the partial testing set of YNBQD and VQA dataset. In addition, $diff = Original_{dev_{All}} - X_{dev_{All}}$, where $X$ is equal to the ``First'', ``Second'', etc.}
\label{table:table19}
\end{table*}

\begin{table*}
\renewcommand\arraystretch{1.0}
\setlength\tabcolsep{11pt}
    \centering
\begin{subtable}[t]{0.3\linewidth}
\centering
\scalebox{0.5}{
    \begin{tabular}{ c | c c c c | c} 
     Task Type &    & \multicolumn{3}{c}{Open-Ended (CIDEr)} &  \\ [0.5ex]
     \hline
     Method &    & \multicolumn{3}{c}{MUTAN without Attention} &  \\ [0.5ex]
     \hline
     Test Set&  \multicolumn{4}{c}{dev} & diff \\ [0.5ex]
     \hline
     Partition & Other & Num & Y/N & All & All \\ [0.5ex] 
     \hline
     First-dev & 0.98 & 1.97 & 24.24 & 10.63 & 49.53  \\ 
     
     Second-dev & 1.18 & 1.90 & 24.06 & 10.65 & 49.51 \\
     
     Third-dev & 1.43 & 2.37 & 33.75 & 14.79 & 45.37  \\
     
     Fourth-dev & 1.28 & 2.46 & 37.13 & 16.12 & 44.04  \\
     
     Fifth-dev & 1.27 & 2.02 & 22.73 & 10.16 & 50.00 \\
     
     Sixth-dev & 1.25 & 1.73 & 27.49 & 12.07 & 48.09  \\
     
     Seventh-dev & 1.38 & 2.33 & 38.10 & 16.55 & 43.61  \\
     \hline
     First-std & 1.07 & 2.24 & 23.83 & 10.57 & 49.88 \\
     \hline
     
     Original-dev  & 47.16 & 37.32 & 81.45 & 60.16 & -\\
     Original-std  & 47.57 & 36.75 & 81.56 & 60.45 & - \\

     \hline
    \end{tabular}}
    \centering
    \captionsetup{justification=centering}
    \caption{MUTAN without Attention model evaluation results.}

\scalebox{0.5}{
    \begin{tabular}{ c | c c c c | c} 
     Task Type &    & \multicolumn{3}{c}{Open-Ended (CIDEr)} &  \\ [0.5ex]
     \hline
     Method &    & \multicolumn{3}{c}{HieCoAtt (Alt,VGG19)} &  \\ [0.5ex]
     \hline
     Test Set&  \multicolumn{4}{c}{dev} & diff \\ [0.5ex]
     \hline
     Partition & Other & Num & Y/N & All & All \\ [0.5ex] 
     \hline
     First-dev & 1.77 & 1.71 & 22.52 & 10.28 & 50.20  \\ 
     
     Second-dev & 2.36 & 1.75 & 31.07 & 14.08 & 46.40 \\
     
     Third-dev & 2.28 & 2.78 & 37.65 & 16.85 & 43.63  \\
     
     Fourth-dev & 2.61 & 3.60 & 31.71 & 14.66 & 45.82  \\
     
     Fifth-dev & 1.95 & 1.96 & 25.75 & 11.72 & 48.76 \\
     
     Sixth-dev & 2.28 & 2.08 & 46.58 & 20.44 & 40.04  \\
     
     Seventh-dev & 2.20 & 3.12 & 29.98 & 13.70 & 46.78  \\
     \hline
     First-std & 1.93 & 1.64 & 22.07 & 10.20 & 50.12 \\
     \hline

     Original-dev  & 49.14 & 38.35 & 79.63 & 60.48 & -\\
     Original-std  & 49.15 & 36.52 & 79.45 & 60.32 & - \\

     \hline
    \end{tabular}}
    \centering
    \captionsetup{justification=centering}
    \caption{HieCoAtt (Alt,VGG19) model evaluation results.}
\end{subtable}%
    \hfil
\begin{subtable}[t]{0.3\linewidth}

\centering
\scalebox{0.5}{
    \begin{tabular}{ c | c c c c | c} 
     Task Type &    & \multicolumn{3}{c}{Open-Ended (CIDEr)} &  \\ [0.5ex]
     \hline
     Method &    & \multicolumn{3}{c}{MLB with Attention} &  \\ [0.5ex]
     \hline
     Test Set&  \multicolumn{4}{c}{dev} & diff \\ [0.5ex]
     \hline
     Partition & Other & Num & Y/N & All & All \\ [0.5ex] 
     \hline
     First-dev & 1.28 & 2.21 & 24.63 & 10.96 & 54.83  \\ 
     
     Second-dev & 1.88 & 2.04 & 29.59 & 13.27 & 52.52 \\
     
     Third-dev & 1.53 & 2.22 & 30.38 & 13.45 & 52.34  \\
     
     Fourth-dev & 2.24 & 2.31 & 26.90 & 12.37 & 53.42  \\
     
     Fifth-dev & 1.70 & 1.80 & 22.54 & 10.27 & 55.52 \\
     
     Sixth-dev & 1.97 & 2.23 & 26.66 & 12.13 & 53.66  \\
     
     Seventh-dev & 1.95 & 2.20 & 34.14 & 15.19 & 50.60  \\
     \hline
     First-std & 1.41 & 2.24 & 24.34 & 10.94 & 54.74 \\
     \hline
     
     Original-dev  & 57.01 & 37.51 & 83.54 & 65.79 & -\\
     Original-std  & 56.60 & 36.63 & 83.68 & 65.68 & - \\

     \hline
    \end{tabular}}
    \centering
    \captionsetup{justification=centering}
    \caption{MLB with Attention model evaluation results.}
    
\scalebox{0.5}{
    \begin{tabular}{ c | c c c c | c} 
     Task Type &    & \multicolumn{3}{c}{Open-Ended (CIDEr)} &  \\ [0.5ex]
     \hline
     Method &    & \multicolumn{3}{c}{MUTAN with Attention} &  \\ [0.5ex]
     \hline
     Test Set&  \multicolumn{4}{c}{dev} & diff \\ [0.5ex]
     \hline
     Partition & Other & Num & Y/N & All & All \\ [0.5ex] 
     \hline
     First-dev & 0.89 & 1.67 & 24.52 & 10.67 & 55.31  \\ 
     
     Second-dev & 1.34 & 1.52 & 27.07 & 11.92 & 54.06 \\
     
     Third-dev & 1.18 & 1.77 & 28.88 & 12.61 & 53.37  \\
     
     Fourth-dev & 1.79 & 2.25 & 33.61 & 14.90 & 51.08  \\
     
     Fifth-dev & 1.14 & 1.09 & 22.81 & 10.03 & 55.95 \\
     
     Sixth-dev & 1.69 & 1.42 & 27.40 & 12.21 & 53.77  \\
     
     Seventh-dev & 1.46 & 1.63 & 40.24 & 17.39 & 48.59  \\
     \hline
     First-std & 0.95 & 1.94 & 24.22 & 10.64 & 55.13 \\
     \hline
     
     Original-dev  & 56.73 & 38.35 & 84.11 & 65.98 & -\\
     Original-std  & 56.29 & 37.47 & 84.04 & 65.77 & - \\

     \hline
    \end{tabular}}
    \centering
    \captionsetup{justification=centering}
    \caption{MUTAN with Attention model evaluation results.}
\end{subtable}%
    \hfil
\begin{subtable}[t]{0.3\linewidth}
        
\centering
\scalebox{0.5}{
    \begin{tabular}{ c | c c c c | c} 
     Task Type &    & \multicolumn{3}{c}{Open-Ended (CIDEr)} &  \\ [0.5ex]
     \hline
     Method &    & \multicolumn{3}{c}{HieCoAtt (Alt,Resnet200)} &  \\ [0.5ex]
     \hline
     Test Set&  \multicolumn{4}{c}{dev} & diff \\ [0.5ex]
     \hline
     Partition & Other & Num & Y/N & All & All \\ [0.5ex] 
     \hline
     First-dev & 2.02 & 3.46 & 25.31 & 11.74 & 50.07  \\ 
     
     Second-dev & 2.57 & 2.99 & 28.60 & 13.30 & 48.51 \\
     
     Third-dev & 3.04 & 3.48 & 33.16 & 15.45 & 46.36  \\
     
     Fourth-dev & 2.95 & 3.47 & 31.42 & 14.69 & 47.12  \\
     
     Fifth-dev & 2.46 & 3.09 & 25.13 & 11.83 & 49.98 \\
     
     Sixth-dev & 3.50 & 3.08 & 35.11 & 16.43 & 45.38  \\
     
     Seventh-dev & 2.83 & 2.90 & 31.99 & 14.80 & 47.01  \\
     \hline
     First-std & 2.13 & 3.55 & 24.88 & 11.65 & 50.41 \\
     \hline

     Original-dev  & 51.77 & 38.65 & 79.70 & 61.81 & -\\
     Original-std  & 51.95 & 38.22 & 79.95 & 62.06 & - \\

     \hline
    \end{tabular}}
    \centering
    \captionsetup{justification=centering}
    \caption{HieCoAtt (Alt,Resnet200) model evaluation results.}

\scalebox{0.5}{
    \begin{tabular}{ c | c c c c | c} 
     Task Type &    & \multicolumn{3}{c}{Open-Ended (CIDEr)} &  \\ [0.5ex]
     \hline
     Method &    & \multicolumn{3}{c}{LSTM Q+I} &  \\ [0.5ex]
     \hline
     Test Set&  \multicolumn{4}{c}{dev} & diff \\ [0.5ex]
     \hline
     Partition & Other & Num & Y/N & All & All \\ [0.5ex] 
     \hline
     First-dev & 1.16 & 3.75 & 24.60 & 11.06 & 46.96  \\ 
     
     Second-dev & 1.17 & 1.60 & 25.30 & 11.12 & 46.90 \\
     
     Third-dev & 1.18 & 2.06 & 30.91 & 13.48 & 44.54  \\
     
     Fourth-dev & 1.69 & 2.15 & 32.15 & 14.24 & 43.78  \\
     
     Fifth-dev & 1.09 & 2.51 & 26.54 & 11.69 & 46.33 \\
     
     Sixth-dev & 1.43 & 0.93 & 36.37 & 15.72 & 42.30  \\
     
     Seventh-dev & 1.47 & 2.06 & 36.58 & 15.94 & 42.08  \\
     \hline
     First-std & 1.17 & 3.67 & 24.15 & 10.90 & 47.28 \\
     \hline
     
     Original-dev  & 43.40 & 36.46 & 80.87 & 58.02 & -\\
     Original-std  & 43.90 & 36.67 & 80.38 & 58.18 & - \\

     \hline
    \end{tabular}}
    \centering
    \captionsetup{justification=centering}
    \caption{LSTM Q+I model evaluation results.}
\end{subtable}
\caption{The table shows the six state-of-the-art pretrained VQA models evaluation results on the YNBQD and VQA dataset. ``-'' indicates the results are not available, ``-std'' represents the accuracy of VQA model evaluated on the complete testing set of YNBQD and VQA dataset and ``-dev'' indicates the accuracy of VQA model evaluated on the partial testing set of YNBQD and VQA dataset. In addition, $diff = Original_{dev_{All}} - X_{dev_{All}}$, where $X$ is equal to the ``First'', ``Second'', etc.}
\label{table:table20}
\end{table*}

\begin{table*}
\renewcommand\arraystretch{1}
\setlength\tabcolsep{11pt}
    \centering
\begin{subtable}[t]{0.3\linewidth}
\centering
\scalebox{0.5}{
    \begin{tabular}{ c | c c c c | c} 
     Task Type &    & \multicolumn{3}{c}{Open-Ended (METEOR)} &  \\ [0.5ex]
     \hline
     Method &    & \multicolumn{3}{c}{MUTAN without Attention} &  \\ [0.5ex]
     \hline
     Test Set&  \multicolumn{4}{c}{dev} & diff \\ [0.5ex]
     \hline
     Partition & Other & Num & Y/N & All & All \\ [0.5ex] 
     \hline
     First-dev & 1.58 & 2.71 & 26.49 & 11.93 & 48.23  \\ 
     
     Second-dev & 1.53 & 2.66 & 26.84 & 12.04 & 48.12 \\
     
     Third-dev & 1.56 & 2.60 & 27.43 & 12.29 & 47.87  \\
     
     Fourth-dev & 1.46 & 2.56 & 27.65 & 12.33 & 47.83  \\
     
     Fifth-dev & 1.50 & 2.67 & 27.50 & 12.30 & 47.86 \\
     
     Sixth-dev & 1.51 & 2.70 & 27.33 & 12.24 & 47.92  \\
     
     Seventh-dev & 1.55 & 2.50 & 27.58 & 12.34 & 47.82  \\
     \hline
     First-std & 1.68 & 3.03 & 26.93 & 12.23 & 48.22 \\
     \hline
     
     Original-dev  & 47.16 & 37.32 & 81.45 & 60.16 & -\\
     Original-std  & 47.57 & 36.75 & 81.56 & 60.45 & - \\

     \hline
    \end{tabular}}
    \centering
    \captionsetup{justification=centering}
    \caption{MUTAN without Attention model evaluation results.}

\scalebox{0.5}{
    \begin{tabular}{ c | c c c c | c} 
     Task Type &    & \multicolumn{3}{c}{Open-Ended (METEOR)} &  \\ [0.5ex]
     \hline
     Method &    & \multicolumn{3}{c}{HieCoAtt (Alt,VGG19)} &  \\ [0.5ex]
     \hline
     Test Set&  \multicolumn{4}{c}{dev} & diff \\ [0.5ex]
     \hline
     Partition & Other & Num & Y/N & All & All \\ [0.5ex] 
     \hline
     First-dev & 2.24 & 2.88 & 27.39 & 12.63 & 47.85  \\ 
     
     Second-dev & 2.21 & 3.06 & 27.66 & 12.75 & 47.73 \\
     
     Third-dev & 2.22 & 3.30 & 27.80 & 12.83 & 47.65  \\
     
     Fourth-dev & 2.21 & 2.89 & 27.85 & 12.80 & 47.68  \\
     
     Fifth-dev & 2.29 & 2.89 & 27.93 & 12.88 & 47.60 \\
     
     Sixth-dev & 2.17 & 2.79 & 28.02 & 12.85 & 47.63  \\
     
     Seventh-dev & 2.29 & 2.97 & 28.21 & 13.00 & 47.48  \\
     \hline
     First-std & 2.17 & 2.77 & 27.54 & 12.69 & 47.63 \\
     \hline

     Original-dev  & 49.14 & 38.35 & 79.63 & 60.48 & -\\
     Original-std  & 49.15 & 36.52 & 79.45 & 60.32 & - \\

     \hline
    \end{tabular}}
    \centering
    \captionsetup{justification=centering}
    \caption{HieCoAtt (Alt,VGG19) model evaluation results.}
\end{subtable}%
    \hfil
\begin{subtable}[t]{0.3\linewidth}

\centering
\scalebox{0.5}{
    \begin{tabular}{ c | c c c c | c} 
     Task Type &    & \multicolumn{3}{c}{Open-Ended (METEOR)} &  \\ [0.5ex]
     \hline
     Method &    & \multicolumn{3}{c}{MLB with Attention} &  \\ [0.5ex]
     \hline
     Test Set&  \multicolumn{4}{c}{dev} & diff \\ [0.5ex]
     \hline
     Partition & Other & Num & Y/N & All & All \\ [0.5ex] 
     \hline
     First-dev & 1.92 & 2.06 & 26.27 & 11.93 & 53.86  \\ 
     
     Second-dev & 1.82 & 2.48 & 26.84 & 12.16 & 53.63 \\
     
     Third-dev & 1.81 & 2.24 & 27.33 & 12.33 & 53.46  \\
     
     Fourth-dev & 1.74 & 2.31 & 27.89 & 12.53 & 53.26  \\
     
     Fifth-dev & 1.84 & 2.34 & 27.57 & 12.45 & 53.34 \\
     
     Sixth-dev & 1.84 & 2.26 & 27.30 & 12.33 & 53.46  \\
     
     Seventh-dev & 1.78 & 2.26 & 27.68 & 12.46 & 53.33  \\
     \hline
     First-std & 1.91 & 2.20 & 26.76 & 12.18 & 53.50 \\
     \hline
     
     Original-dev  & 57.01 & 37.51 & 83.54 & 65.79 & -\\
     Original-std  & 56.60 & 36.63 & 83.68 & 65.68 & - \\

     \hline
    \end{tabular}}
    \centering
    \captionsetup{justification=centering}
    \caption{MLB with Attention model evaluation results.}
    
\scalebox{0.5}{
    \begin{tabular}{ c | c c c c | c} 
     Task Type &    & \multicolumn{3}{c}{Open-Ended (METEOR)} &  \\ [0.5ex]
     \hline
     Method &    & \multicolumn{3}{c}{MUTAN with Attention} &  \\ [0.5ex]
     \hline
     Test Set&  \multicolumn{4}{c}{dev} & diff \\ [0.5ex]
     \hline
     Partition & Other & Num & Y/N & All & All \\ [0.5ex] 
     \hline
     First-dev & 1.55 & 2.35 & 26.75 & 11.98 & 54.00  \\ 
     
     Second-dev & 1.48 & 2.46 & 27.23 & 12.16 & 53.82 \\
     
     Third-dev & 1.42 & 2.25 & 27.63 & 12.27 & 53.71  \\
     
     Fourth-dev & 1.38 & 2.49 & 28.28 & 12.54 & 53.44  \\
     
     Fifth-dev & 1.43 & 2.30 & 27.91 & 12.39 & 53.59 \\
     
     Sixth-dev & 1.44 & 2.25 & 27.97 & 12.41 & 53.57  \\
     
     Seventh-dev & 1.42 & 2.08 & 27.69 & 12.27 & 53.71  \\
     \hline
     First-std & 1.57 & 2.31 & 27.41 & 12.30 & 53.47 \\
     \hline
     
     Original-dev  & 56.73 & 38.35 & 84.11 & 65.98 & -\\
     Original-std  & 56.29 & 37.47 & 84.04 & 65.77 & - \\

     \hline
    \end{tabular}}
    \centering
    \captionsetup{justification=centering}
    \caption{MUTAN with Attention model evaluation results.}
\end{subtable}%
    \hfil
\begin{subtable}[t]{0.3\linewidth}
        
\centering
\scalebox{0.5}{
    \begin{tabular}{ c | c c c c | c} 
     Task Type &    & \multicolumn{3}{c}{Open-Ended (METEOR)} &  \\ [0.5ex]
     \hline
     Method &    & \multicolumn{3}{c}{HieCoAtt (Alt,Resnet200)} &  \\ [0.5ex]
     \hline
     Test Set&  \multicolumn{4}{c}{dev} & diff \\ [0.5ex]
     \hline
     Partition & Other & Num & Y/N & All & All \\ [0.5ex] 
     \hline
     First-dev & 2.71 & 3.26 & 26.99 & 12.73 & 49.08  \\ 
     
     Second-dev & 2.81 & 3.34 & 27.43 & 12.97 & 48.84 \\
     
     Third-dev & 2.83 & 3.41 & 27.46 & 13.00 & 48.81  \\
     
     Fourth-dev & 2.78 & 3.12 & 27.22 & 12.85 & 48.96  \\
     
     Fifth-dev & 2.70 & 3.12 & 27.30 & 12.84 & 48.97 \\
     
     Sixth-dev & 2.77 & 2.97 & 27.37 & 12.89 & 48.92  \\
     
     Seventh-dev & 2.76 & 3.03 & 27.78 & 13.06 & 48.75  \\
     \hline
     First-std & 2.73 & 3.03 & 27.26 & 12.87 & 49.19 \\
     \hline

     Original-dev  & 51.77 & 38.65 & 79.70 & 61.81 & -\\
     Original-std  & 51.95 & 38.22 & 79.95 & 62.06 & - \\

     \hline
    \end{tabular}}
    \centering
    \captionsetup{justification=centering}
    \caption{HieCoAtt (Alt,Resnet200) model evaluation results.}

\scalebox{0.5}{
    \begin{tabular}{ c | c c c c | c} 
     Task Type &    & \multicolumn{3}{c}{Open-Ended (METEOR)} &  \\ [0.5ex]
     \hline
     Method &    & \multicolumn{3}{c}{LSTM Q+I} &  \\ [0.5ex]
     \hline
     Test Set&  \multicolumn{4}{c}{dev} & diff \\ [0.5ex]
     \hline
     Partition & Other & Num & Y/N & All & All \\ [0.5ex] 
     \hline
     First-dev & 1.64 & 2.78 & 27.95 & 12.56 & 45.46  \\ 
     
     Second-dev & 1.48 & 2.82 & 28.42 & 12.68 & 45.34 \\
     
     Third-dev & 1.63 & 2.43 & 28.44 & 12.72 & 45.30  \\
     
     Fourth-dev & 1.47 & 2.58 & 28.65 & 12.74 & 45.28  \\
     
     Fifth-dev & 1.57 & 2.47 & 29.04 & 12.94 & 45.08 \\
     
     Sixth-dev & 1.59 & 2.57 & 28.46 & 12.72 & 45.30  \\
     
     Seventh-dev & 1.52 & 2.40 & 28.84 & 12.83 & 45.19  \\
     \hline
     First-std & 1.53 & 2.75 & 28.19 & 12.64 & 45.54 \\
     \hline
     
     Original-dev  & 43.40 & 36.46 & 80.87 & 58.02 & -\\
     Original-std  & 43.90 & 36.67 & 80.38 & 58.18 & - \\

     \hline
    \end{tabular}}
    \centering
    \captionsetup{justification=centering}
    \caption{LSTM Q+I model evaluation results.}
\end{subtable}
\caption{The table shows the six state-of-the-art pretrained VQA models evaluation results on the YNBQD and VQA dataset. ``-'' indicates the results are not available, ``-std'' represents the accuracy of VQA model evaluated on the complete testing set of YNBQD and VQA dataset and ``-dev'' indicates the accuracy of VQA model evaluated on the partial testing set of YNBQD and VQA dataset. In addition, $diff = Original_{dev_{All}} - X_{dev_{All}}$, where $X$ is equal to the ``First'', ``Second'', etc.}
\label{table:table21}
\end{table*}

\end{document}